\documentclass[pdflatex,sn-basic]{sn-jnl}% Basic Springer Nature Reference Style/Chemistry Reference Style
\usepackage{csquotes}
\usepackage{graphicx, rotating}%
\usepackage{multirow}%
\usepackage{amsmath,amssymb,amsfonts}%
\usepackage{amsthm}%
\usepackage{mathrsfs}%
\usepackage[title]{appendix}%
\usepackage{xcolor}%
\usepackage{textcomp}%
\usepackage{manyfoot}%
\usepackage{booktabs}%
\usepackage{algorithm}%
\usepackage{algorithmicx}%
\usepackage{algpseudocode}%
\usepackage{listings}%
\usepackage{tcolorbox}
\usepackage{ragged2e}

\usepackage{array}
\usepackage{makecell}
\usepackage{tabularx}
\usepackage{adjustbox}
\usepackage{caption}
\usepackage{booktabs}
\usepackage{tabularx}
\usepackage{threeparttable}
\usepackage{makecell}
\usepackage{kotex}
\usepackage{ml-arxiv-readable}

\theoremstyle{thmstyleone}%
\theoremstyle{thmstyletwo}%

\theoremstyle{thmstylethree}%

\begin{document}

\title[Article Title]{
% Cross-Layer Safety for Long-Horizon Robotic Manipulation: An Embodied AI Perspective
Safe Embodied AI for Long-horizon Tasks: A Cross-layer Analysis of Robotic Manipulation
% Long-Horizon Safety in Embodied AI: A Cross-Layer Survey on Robotic Manipulation
}


\author[1,2]{\fnm{Dabin} \sur{Kim}}\email{dabin404@snu.ac.kr}
% \author[1,2]{\fnm{Dabin} \sur{Kim}}\email{dabin404@snu.ac.kr}
% \author[1,2]{\fnm{Dabin} \sur{Kim}}\email{dabin404@snu.ac.kr}

\author[3]{\fnm{Daemin} \sur{Park}}\email{eoalsqkr12@snu.ac.kr}
% % \equalcont{These authors contributed equally to this work.}
\author[4]{\fnm{Sangyub} \sur{Lee}}\email{nickyub@snu.ac.kr}
\author[3]{\fnm{Jinsik} \sur{Kim}}\email{jinsik03@snu.ac.kr}
\author[3]{\fnm{Yeongtak} \sur{Oh}}\email{oyt9306@gmail.com}
\author[5]{\fnm{Jongho} \sur{Shin}}\email{jongho@cs.stanford.edu}

\author*[2,3,4]{\fnm{Sungroh} \sur{Yoon}}\email{sryoon@snu.ac.kr}

\affil[1]{\orgdiv{UNIST InnoCORE AI-Space Solar Initiative}, \orgname{Ulsan National Institute of Science and Technology (UNIST)}, \orgaddress{\street{50 UNIST-gil}, \city{Ulsan}, \postcode{44919}, \country{Republic of Korea}}}

\affil[2]{\orgdiv{Automation and Systems Research Institute}, \orgname{Seoul National University}, \orgaddress{\street{1 Gwanak-ro}, \city{Seoul}, \postcode{08826}, \country{Republic of Korea}}}

\affil[3]{\orgdiv{Department of Electrical and Computer Engineering}, \orgname{Seoul National University}, \orgaddress{\street{1 Gwanak-ro}, \city{Seoul}, \postcode{08826}, \country{Republic of Korea}}}

\affil[4]{\orgdiv{Interdisciplinary Program in Artificial Intelligence}, \orgname{Seoul National University}, \orgaddress{\street{1 Gwanak-ro}, \city{Seoul}, \postcode{08826}, \country{Republic of Korea}}}

\affil[5]{\orgname{LG Electronics}, \orgaddress{\street{128 Yeoui-daero}, \city{Seoul}, \postcode{07336}, \country{Republic of Korea}}}

\abstract{
Embodied AI systems are increasingly expected to reason and act over extended horizons in physical environments. This growing capability brings safety to the foreground, because failures in the physical world can harm people, damage objects, and disrupt workplaces. Although safe embodied AI has attracted substantial attention, the literature remains fragmented across planning, policy design, and runtime execution. Long-horizon robotic manipulation is a particularly revealing anchor domain for this problem because semantic misgrounding, subtask-level error propagation, execution drift, and contact-rich physical risk can accumulate within the same closed-loop system. This survey therefore provides a structured review of safety in long-horizon robotic manipulation from an embodied AI perspective. We organize the literature by intervention locus, covering planning-time, policy-time, and execution-time safety, and we analyze the strength of the evidence that each line of work provides, distinguishing formal guarantees, statistical support, and empirical safety heuristics. This framework clarifies the distinct roles of backbone capability papers, direct safety mechanisms, and benchmark or evaluation studies, while exposing where current safety claims are well supported and where they remain indirect. We identify persistent gaps, including limited evidence for policy-time safety, weak formal support for contact-rich long-horizon manipulation, immature uncertainty-triggered intervention, and a shortage of manipulation-specific safety benchmarks. We conclude by outlining research directions for cross-layer assurance, evaluation design, and safer deployment of long-horizon robotic agents in real-world settings.}




\maketitle

\section{Introduction} \label{sec:intro}

Embodied AI systems are increasingly deployed in open, physically consequential environments. As robotics becomes more deeply integrated with foundation models, vision-language-action (VLA) systems, and broad agent architectures, the central challenge has shifted: the question is no longer whether these systems can complete tasks, but whether they can do so safely in the real-world \citep{liu2025aligningcyberspace,firoozi2024foundationmodelsrobotics,hu2023towardgeneralpurpose,kawaharazuka2025visionlanguageaction}. This urgency stems from the fact that embodied systems do not merely generate predictions or content. Their outputs move hardware, manipulate objects, contact the physical world, and can directly affect humans and surrounding environments. Safety in embodied AI must therefore be treated as a core system property rather than as an optional post-hoc constraint.

\begin{figure}[t]
    \centering
    \includegraphics[width=1.0\linewidth]{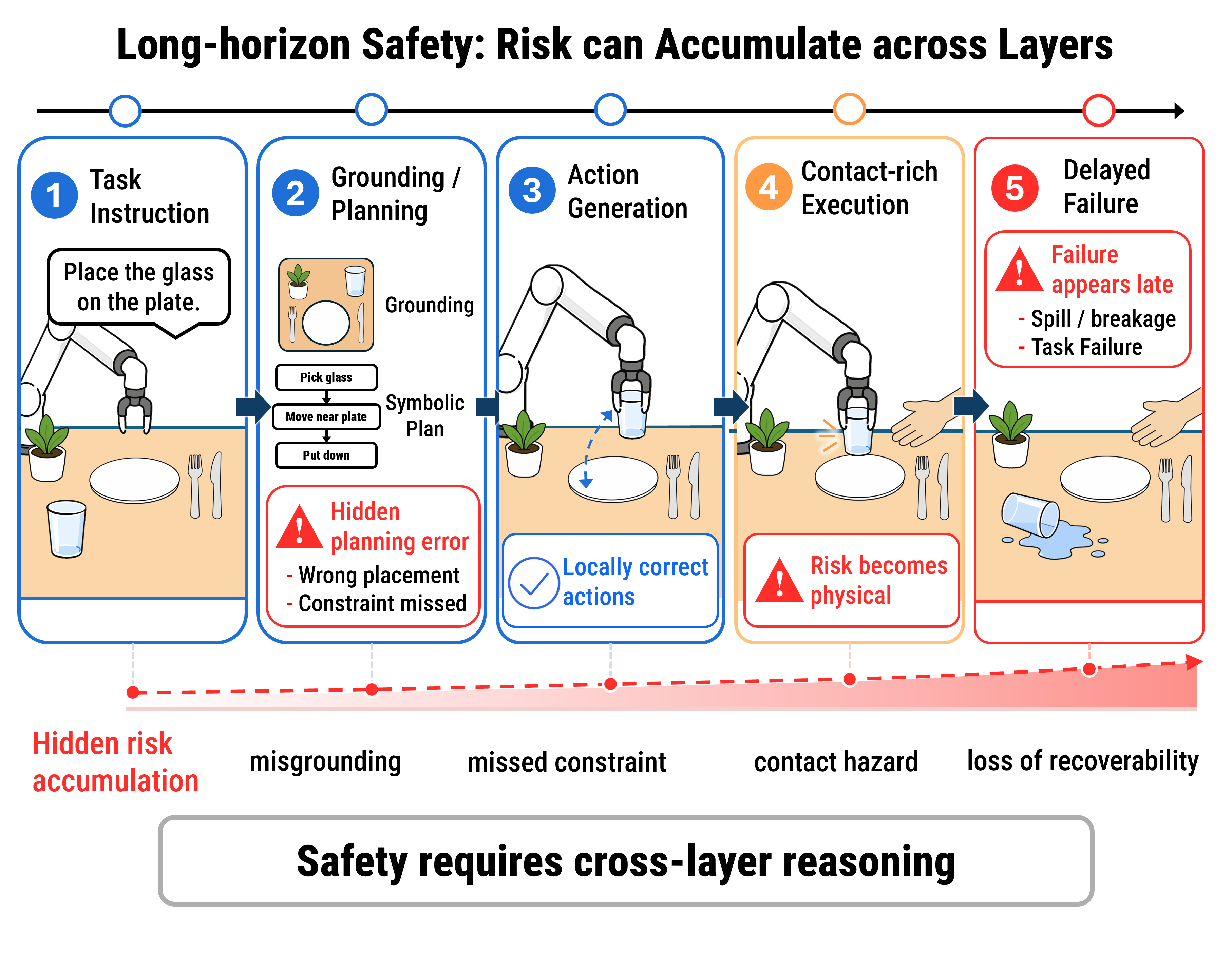}
    \caption{Motivating illustration of long-horizon manipulation safety as a cross-layer problem. A multi-stage tabletop manipulation task can appear locally competent at each step while hidden risk accumulates across grounding, action generation, and contact-rich execution. Small early errors may therefore compound and surface only later in the rollout as a larger physical safety failure.}
    \label{fig:front_figure}
\end{figure}

The core difficulty is not only that robots can fail, but that long-horizon embodied systems can fail in delayed, compounding, and partially observable ways~\citep{ross2011reduction}. Errors may begin as semantic misgrounding or visual hallucination \citep{zhou2023analyzing,zhai2023hallecontrolcontrolling}, propagate through underspecified planning objectives or policy-level misalignment, and surface only when the robot reaches a later contact-rich interaction under uncertainty, disturbance, or environmental change \citep{ravichandran2026contextualsafetyreasoning,feng2025fromwordssafety,ouyang2025longhorizonlocomotion,song2025towards}. A system may therefore appear locally competent while still accumulating hidden risk. Fig.~\ref{fig:front_figure} illustrates this motivation: safety in long-horizon manipulation is not confined to one module, but emerges from the interaction between grounding, action generation, execution, and physical feedback.
In this survey, we use safety in a broad embodied sense. Safety is not limited to immediate collision avoidance or force-threshold violation; it also includes safety-relevant failures that can produce physical, operational, or human-facing risk, or cause irreversible degradation of task recoverability during a long-horizon rollout. Section~\ref{sec:section2} provides details about this scope used throughout the survey.

While existing literature across robot learning, control theory, and reliable AI has illuminated specific facets of safety in embodied AI, most efforts remain fragmented and lack a holistic framework. Safe control and runtime shielding provide rigorous tools for low-level constraint satisfaction under modeled dynamics \citep{ames2019control,bansal2017hamilton,wabersich2023data,hwang2024safe}, whereas safe reinforcement learning (RL) focuses on constraint-aware exploration and policy deployment \citep{liu2022robot,liu2025safereinforcementlearning,ji2023omnisafeinfrastructureaccelerating,gu2024reviewsafereinforcement}. Simultaneously, alignment and guardrail research addresses multimodal safety, including harmful instructions and behavioral constraints \citep{bai2022training_rlhf,bian2023beavertailstowardsimproved,ji2024alignanythingtraining,inan2023llamaguardllm,chi2024llamaguardvision}. Furthermore, physical safety is governed by industrial protocols such as ISO 10218 and ANSI/RIA R15.06 \citep{iso10218_1_2025,ansiria_r15_06_2012_alt}. Despite these advancements, these domains are typically studied in isolation, even though many embodied failures emerge precisely from the unaddressed interactions between these layers. This survey aims to map this scattered landscape, analyzing how diverse methodologies conceptualize safety and identifying the integrated components needed to analyze and support end-to-end safety in embodied AI.

Rather than attempting an exhaustive review of all embodied AI safety, this survey focuses on long-horizon robotic manipulation as an analytically dense anchor domain. 
We do not treat manipulation as a naive proxy for the broader landscape; rather, we use it because it concentrates several safety pressures that are often decoupled in other domains: semantic task specification~\citep{saycan,huang2023voxposercomposable3d}, delayed error propagation across subtasks~\citep{yang2026lilo_vla,huang2026wmrobotictask}, contact-rich physical interaction~\citep{he2025foarforceaware,yu2025forcevla}, and latent unsafe behaviors that may persist despite nominal task success~\citep{zhang2025responsiblerobotbenchbenchmarkingresponsible,lu2026isbenchevaluating}. 
For instance, a grounding or planning error early in a sequence may remain dormant until a subsequent object interaction triggers a failure~\citep{guo2024doremigroundinglanguage}. 
% Similarly, a task may achieve its objective while exhibiting unsafe force levels, brittle recovery, or near-failure dynamics~\citep{lin2025failsafereasoningand,dai2025racerrichlanguage}. 
Manipulation thus serves as a rigorous stress test for cross-layer safety claims.

Evaluating safety across the entire embodied AI spectrum also requires distinguishing task domains from robot platforms. 
Embodied AI encompasses several robotics domains, including navigation~\citep{shah2023gnm}, legged locomotion~\citep{lee2020learning}, and manipulation~\citep{saycan}. 
These domains are distinct from robot platforms or application categories: humanoids~\citep{cao2025humanoid}, mobile manipulators~\citep{ghodsian2023mobile}, service robots~\citep{nanavati2023physically}, and autonomous vehicles~\citep{bathla2022autonomous} may instantiate or combine several domains within a single deployed system. 
% We therefore compare domains rather than robot form factors or deployment sectors.
In our survey, we distinguish our main focus from robot forms or deployment sectors to clearly convey safety-oriented analyses.

% Embodied AI encompasses navigation~\citep{shah2023gnm}, locomotion~\citep{lee2020learning}, autonomous driving~\citep{wang2025alpamayo}, and manipulation, whereas platforms such as humanoids or mobile manipulators may combine several of these domains within a single embodiment. 
% We therefore compare settings at the domain level rather than the platform level. 
Table~\ref{tab:why_manipulation} provides a qualitative positioning along five safety-relevant factors: long-horizon dependence, semantic specification, contact-rich interaction, latent unsafe behavior under nominal task success, and coupling across planning, policy generation, and execution~\citep{saycan,shirai2024visionlanguageinterpreter}. 
This comparison underscores why manipulation serves as a uniquely compact paradigm compare to navigation and locomotion, where all five challenges intersect simultaneously, rendering it an ideal testbed for studying long-horizon embodied AI safety.

% Evaluating safety across the entire embodied AI spectrum requires distinguishing fundamental task domains from robot platforms. While embodied AI encompasses navigation \citep{shah2023gnm}, locomotion \citep{lee2020learning}, and autonomous driving \citep{wang2025alpamayo}, these domains are analytically distinct from hardware platforms like humanoids or mobile manipulators, which combine multiple domains within a single form factor. To accurately assess safety pressures, our analysis must remain at the domain level. Table~\ref{tab:why_manipulation} compares these embodied settings accordingly, demonstrating why manipulation stands out: it is the setting where long-horizon dependence, semantic specification pressure, and cross-layer coupling converge most intensely. 

Because manipulation concentrates these cross-layer safety pressures, a dedicated survey on this domain addresses a critical blind spot in the current literature. Existing broad surveys on embodied AI, foundation models, and VLA systems successfully map the modern ecosystem and acknowledge the need for safe real-world deployment \citep{liu2025aligningcyberspace,firoozi2024foundationmodelsrobotics,kawaharazuka2025visionlanguageaction}. Conversely, surveys focused on embodied manipulation clarify learning formulations and force-aware execution \citep{zheng2025surveyembodiedlearning,tsuji2026surveyimitationlearning,zhang2025safe}, while safety-specific surveys predominantly cover low-level safe control and safe RL~\citep{ames2019control,bansal2017hamilton,wabersich2023data,gu2024reviewsafereinforcement}. The closest overarching frameworks are provided by \cite{tan2025towards} and \cite{kojima2025comprehensive}, which address broad safety risks in embodied AI and physical risk control in foundation-model-enabled robotics, respectively. However, while these works are valuable, they treat safety either as an abstract concept or as a modular technique. They neither systematically organize long-horizon safety by where safety interventions enter the lifecycle, nor critically assess what level of evidence supports the resulting claims. We summarize the main emphasis of each related survey and the gaps addressed in this work in Table~\ref{tab:survey_positioning}.

\begin{table}[t]
  \centering
  \small
  \setlength{\tabcolsep}{4.5pt}
  \caption{Qualitative comparison of representative embodied robotics domains for safety analysis.}
  \label{tab:why_manipulation}
  \begin{tabularx}{\columnwidth}{@{}>{\raggedright\arraybackslash}Xccccc@{}}
      \toprule
      Domain & Horizon & Semantic & Contact & Hidden & Coupling \\
      \midrule
      Navigation
        & $\checkmark$ & $\triangle$ & $\times$ & $\triangle$ & $\triangle$ \\
      Locomotion
        & $\triangle$ & $\times$ & $\checkmark$ & $\triangle$ & $\triangle$ \\
      \textbf{Manipulation}
        & $\checkmark$ & $\checkmark$ & $\checkmark$ & $\checkmark$ & $\checkmark$ \\
      \bottomrule
  \end{tabularx}
  \vspace{2pt}
  \begin{minipage}{\columnwidth}
  \footnotesize
  \setlength{\parindent}{0pt}
  \emph{Coding.} 
  $\checkmark$: central and recurring; 
  $\triangle$: variant-dependent or secondary; 
  $\times$: typically absent.\\
  % \emph{Scope.} 
  % Semantic denotes scene/task specification pressure, including language grounding when applicable. 
  % Contact denotes task-relevant contact dynamics or contact switching; navigation collisions are treated as failures, not task mechanisms.\\
  \emph{Evidence.} 
  The coding is based on representative work on navigation~\citep{shah2023gnm,gu2022vision}, 
  legged locomotion~\citep{lee2020learning}, 
  and long-horizon, language-conditioned, contact-rich, and safety-evaluated manipulation~\citep{saycan,huang2023voxposercomposable3d,yang2026lilo_vla,huang2026wmrobotictask,he2025foarforceaware,yu2025forcevla,lu2026isbenchevaluating,zhang2025responsiblerobotbenchbenchmarkingresponsible}.\\
  % \emph{Caveat.} 
  % Symbols are conservative domain-level codes, not exhaustive claims about all possible instantiations.
  \end{minipage}
\end{table}

% \begin{table}[t]
%   \centering
%   \small
%   \setlength{\tabcolsep}{4.5pt}
%   \caption{Qualitative positioning of representative embodied settings for safety analysis.}
%   \label{tab:why_manipulation}

%   \begin{tabularx}{\columnwidth}{@{}>{\raggedright\arraybackslash}Xccccc@{}}
%       \toprule
%       Setting & Horizon & Semantic & Contact & Hidden & Coupling \\
%       \midrule
%       Goal-conditioned navigation        & $\checkmark$ & $\triangle$  & $\times$     & $\triangle$  & $\triangle$ \\
%       Legged locomotion                  & $\triangle$  & $\times$     & $\checkmark$ & $\triangle$  & $\triangle$ \\
%       Mobile service tasks  & $\checkmark$ & $\checkmark$ & $\times$     & $\triangle$  & $\checkmark$ \\
%       Autonomous driving              & $\checkmark$ & $\checkmark$ & $\times$     & $\checkmark$ & $\checkmark$ \\
%       \textbf{Long-horizon manipulation}
%                                          & $\checkmark$ & $\checkmark$ & $\checkmark$ & $\checkmark$ & $\checkmark$ \\
%       \bottomrule
%   \end{tabularx}

%   \vspace{2pt}
%   \begin{minipage}{\columnwidth}
%   \footnotesize
%   \emph{Notation.}
%   \textcolor{purple}{$\checkmark$: central and recurring factor; 
%   $\triangle$: present in some variants or secondary; 
%   $\times$: typically absent under the task-family definition.
%   Contact denotes intentional, safety-relevant physical contact or contact switching; 
%   collisions in navigation/driving are treated as failures rather than task-relevant contact.}
%   \end{minipage}
% \end{table}

\newcolumntype{L}[1]{>{\RaggedRight\arraybackslash}p{#1}}
\newcolumntype{Y}{>{\RaggedRight\arraybackslash}X}

\begin{table}[t]
\centering
{
% \footnotesize
\setlength{\tabcolsep}{3pt}
\renewcommand{\arraystretch}{1.08}
\caption{Positioning of this survey relative to adjacent surveys.}
\label{tab:survey_positioning}
\begin{tabularx}{\linewidth}{@{}L{0.19\linewidth}L{0.22\linewidth}L{0.28\linewidth}Y@{}}
\toprule
\textbf{References} &
\textbf{Primary scope} &
\textbf{Main emphasis} &
\textbf{Gap addressed here} \\
\midrule

\citet{liu2025aligningcyberspace}
& Broad embodied AI
& Perception; interaction; sim-to-real adaptation
& Manipulation-specific safety evidence \\
\addlinespace[2pt]

\citet{firoozi2024foundationmodelsrobotics}
& Foundation models for robotics
& Applications; uncertainty; safety evaluation
& Safety as lifecycle-level organizing lens \\
\addlinespace[2pt]

\citet{kawaharazuka2025visionlanguageaction}
& VLA models for robotics
& Architectures; datasets; benchmarks; evaluation
& Safety beyond capability metrics \\
\addlinespace[2pt]

\citet{zheng2025surveyembodiedlearning}
& Object-centric manipulation
& Object perception; task modeling; policy learning
& Safety evidence as a primary analytical axis \\
\addlinespace[2pt]

\citet{tsuji2026surveyimitationlearning}
& Contact-rich imitation learning
& Demonstrations; sensing; representation; learning
& Long-horizon safety evidence for manipulation \\
\addlinespace[2pt]

\citet{zhang2025safe}
& Safe contact-rich learning
& Safe exploration; shielding; execution safety
& Integration of planning-, policy-, and evaluation-level evidence \\
\addlinespace[2pt]

\citet{gu2024reviewsafereinforcement}
& Safe RL
& Constraints; safe RL theory; benchmarks
& Safety beyond policy-learning formulations \\
\addlinespace[2pt]

\citet{tan2025towards}
& Trustworthy embodied AI
& Maturity levels; trustworthiness principles
& Domain-grounded in-depth analysis with manipulation \\
\addlinespace[2pt]

\citet{kojima2025comprehensive}
& Physical risk control
& Pre-deployment; pre-incident; post-incident risk control
& Boundaries among task-, policy-, and execution-level evidence \\
\bottomrule
\end{tabularx}
}
\end{table}

Consequently, the central question of this survey is not just how to build more capable embodied systems, but how to evaluate and compare the safety claims made about them. We organize the literature along two primary axes. The first is \emph{intervention locus}: where safety enters the system, including planning-time, policy-time, and execution-time mechanisms. The second is \emph{evidence boundary}: what a reported result actually supports, and what it does not prove. We distinguish formal guarantees, statistical safety evidence, and empirical safety evidence, while separating these from capability improvements or generic robustness results that only indirectly support safety. This two-axis perspective clarifies the distinct roles of backbone capability papers, direct safety mechanisms, and benchmark or evaluation studies, while exposing where current safety claims are strongly supported and where they remain partial or indirect.

The main contributions of this survey are as follows.
\begin{itemize}
    \item We position long-horizon robotic manipulation as an dense anchor domain for embodied AI safety, showing how it brings together semantic task specification, delayed subtask-level error propagation, spatial and motion-feasibility constraints, contact-rich physical interaction, and hidden unsafe behavior under nominal task success.

    \item We introduce a cross-layer organization of the literature by \emph{intervention locus}. Rather than treating safety as a monolithic property, we map where safety mechanisms enter the lifecycle: planning-time task formation and validation, policy-time action-generation shaping, and execution-time monitoring, restoration, and contact regulation.

    \item We propose an evidence boundary framework for interpreting safety claims. By distinguishing formal guarantees, statistical safety evidence, and empirical safety evidence, we clarify what each line of work actually supports and avoid overreading task success, capability improvement, or generic robustness as direct evidence of safety.

    \item We analyze current evaluation and benchmark practices for long-horizon manipulation safety. We show that capability-oriented benchmarks remain insufficient for procedural safety, and that safety-aware evaluation is fragmented across plan-level screening, rollout-level safe success, runtime detection, recovery, and contact-quality metrics.

    \item We synthesize cross-layer research directions for safer embodied AI, including abstraction-preserving safety representations, sim-to-real safety evidence, embodiment-aware revalidation, calibrated intervention selection, procedural safety observability, and safety-oriented data and evaluation infrastructure.

    \item We diagnose the critical pitfalls of existing models through a dedicated analysis, thereby establishing more fine-grained safety boundaries for long-horizon tasks. Based on this systematic analysis, we structure the essential safety constraints that future manipulation frameworks must adhere to and highlight the critical safety challenges that next-generation frameworks must overcome.
\end{itemize}

The remainder of this paper is organized as follows. Section~\ref{sec:section2} introduces the cross-layer framework used throughout the survey. Sections~\ref{sec:section3}--\ref{sec:section5} review planning-time, policy-time, and execution-time safety mechanisms, respectively. Section~\ref{sec:section6} examines evaluation and benchmark practices for safety analysis, and Section~\ref{sec:section7} discusses future directions for cross-layer safety assurance. Section~\ref{sec:section8} concludes the survey.

% The remainder of this paper is organized as follows. Section~\ref{sec:section2} introduces the main taxonomy and clarifies the scope of the survey, including our treatment of lifecycle states, failure modes, and evidence. Section~\ref{sec:section3} reviews planning-level safety, and Section~\ref{sec:section4} examines policy-time safety. Section~\ref{sec:section5} discusses execution-time safety assurance, while Section~\ref{sec:section6} reviews how safety in robotic manipulation is evaluated through datasets, benchmarks, and assurance-oriented evidence. Section~\ref{sec:section7} outlines open challenges and promising future research directions. Section~\ref{sec:section8} concludes the survey.

\paragraph{Literature Scope and Selection Criteria}
This survey follows a framework-guided review strategy. The review focuses on safety in long-horizon robotic manipulation, rather than robot manipulation, embodied AI, or robot safety in general. We therefore selected papers according to their relevance to three analytical questions: where safety enters the manipulation lifecycle, which failure mode or risk pressure is addressed, and what form of evidence supports the corresponding safety claim.

The corpus was constructed in stages. We first considered work directly related to long-horizon robotic manipulation, language-conditioned task planning, vision-language-action policies, runtime monitoring, intervention, recovery, contact-rich execution, and safety-oriented evaluation. We then expanded this set through targeted searches in IEEE Xplore, ACM Digital Library, arXiv, Google Scholar, Semantic Scholar, and publisher pages, followed by reference tracing from representative papers and neighboring surveys in safe control, safe RL, embodied-agent evaluation, robot foundation models, and broader embodied AI safety. Search terms combined phrases such as ``long-horizon robotic manipulation'', ``robot safety'', ``vision-language-action safety'', ``runtime monitoring'', ``failure recovery'', ``safe control'', ``safe reinforcement learning'', ``contact-rich manipulation'', and ``robot manipulation benchmark''. The corpus was updated through April 2026.

A paper was included when it met at least one of four criteria. First, it addresses a safety-relevant failure source in long-horizon manipulation, such as specification error, unsafe action generation, execution drift, or contact-rich physical risk. Second, it proposes a mechanism for constraining, verifying, monitoring, correcting, recovering, or evaluating robot behavior under safety-relevant conditions. Third, it provides formal, statistical, or empirical evidence that helps determine the strength of a safety claim. Fourth, it supplies the necessary contextual foundation for interpreting manipulation-specific safety evidence within the broader landscape of these adjacent domains.
Papers whose main contribution is general capability improvement, foundation-model scaling, or broad embodied AI benchmarking are not treated as direct safety evidence. 
% They are included only when they materially clarify a manipulation-specific safety claim, and are coded as contextual or boundary papers rather than as core safety mechanisms.

% For analysis, each paper was assigned a primary intervention locus: planning-time, policy-time, execution-time, or evaluation and assurance. We also coded the evidence type supported by the paper's validation, distinguishing formal guarantees, statistical support, empirical safety evidence, and robustness-oriented results. For papers spanning multiple layers, primary placement follows the dominant point of safety intervention, while secondary relevance is discussed when it changes the interpretation of the claim.

Note that because the field is moving quickly, especially around VLA models and embodied-agent safety benchmarks, some included papers are recent preprints whose archival status and empirical maturity may change. 

\section{Cross-layer Safety Framework for Long-horizon Robotic Manipulation} \label{sec:section2}

This section establishes the conceptual framework that guides the remainder of the survey. 
Central to our framework is the realization that safe long-horizon robotic manipulation cannot be treated as a property of an isolated module or metric; instead, it must be understood as an emergent characteristic of the entire system.
% Safe long-horizon robotic manipulation is not a property of a single module or metric, but an emergent characteristic of the entire system. 
A task may become unsafe at various stages: through ambiguous instructions, incorrect goal grounding, violations of hidden preconditions in subtask sequences, or misaligned policy objectives. Even during execution, safety can be compromised by uncertainty-induced drift or hazardous physical interactions such as excessive force, jamming, or damage. Consequently, a method that improves safety at one layer does not inherently guarantee the safety of the complete embodied system.

We therefore treat safety as a \textit{cross-layer safety-assessment problem}. The central inquiry extends beyond whether a robot completes a task to examine where safety assumptions are introduced, how they transform across the system, and what evidence supports the resulting safety claims. This perspective is vital for long-horizon manipulation, where risks are often delayed and compositional: a locally plausible goal interpretation can lead to an unsafe plan, and a valid plan may still induce unsafe action proposals or unrecoverable contact errors.
% We therefore treat safety as a cross-layer assurance problem. The central question is not only whether a robot completes a task, but where safety-relevant assumptions are introduced, how they are transformed across the system, when risk becomes observable, and what kind of evidence supports the resulting safety claim. This perspective is especially important for long-horizon manipulation because risks are often delayed and compositional. A locally plausible goal interpretation can create an unsafe plan; a valid plan can still induce an unsafe action proposal; an apparently safe policy rollout can drift under changing observations; and a minor contact error can eventually destroy the possibility of safe recovery.

% In this survey, \textit{safety} refers to the prevention, detection and mitigation of failures that result in physical harm, object damage, or irreversible loss of recoverability. We distinguish safety-relevant failures from nominal task errors; a failure becomes a safety issue when it increases operational risk, violates constraints, or allows hidden hazards to propagate across the rollout. This distinction is critical because nominal task success and safe execution are not equivalent. A robot may complete a task through brittle recovery or near misses, while another may \enquote{fail safely} by detecting risk early and preserving a recoverable state.

In this survey, \textit{safety} refers to the prevention, detection, and mitigation of failures that can lead to physical harm, object damage, specified constraint violation, or unrecoverable task states during long-horizon tasks. 
This includes (i) \textit{physical safety}, which concerns harm to humans, the robot, manipulated objects, or the environment; (ii) \textit{procedural safety}, which concerns maintaining task order, preconditions, constraints, and recoverable states across a rollout; (iii) \textit{operational safety}, which concerns avoiding states in which continued autonomous execution becomes unsafe or no longer justifiable; and (iv) \textit{semantic safety}, which concerns hazardous instructions, misgrounded goals, hallucinated affordances, or omitted constraints that can make an otherwise plausible execution unsafe.
These categories represent diverse and orthogonal components of system safety for long-horizon embodied AI. Accordingly, this survey delves into dedicated, fine-grained safety measures, addressing risks that emerge from long-horizon challenges, which are particularly evident in robotic manipulation. 
% We therefore distinguish safety-relevant failures from nominal task errors: nominal task success and safe execution are not equivalent, and a robot may fail safely if it detects risk early and preserves a recoverable state.
% These categories of safety are diverse and orthogonal components of the system safety for long-horizon embodied AI, and this survey focuses on dedicated, fine-grained safety especially focused on safety risks emerging from long-horizon challenges which can be observed in case of robotic manipulation. 
% We therefore distinguish safety-relevant failures from nominal task errors: nominal task success and safe execution are not equivalent, and a robot may fail safely if it detects risk early and preserves a recoverable state.

This boundary also governs our treatment of adjacent concepts. Reliability, robustness, and alignment do not inherently constitute safety evidence; they qualify as such only when explicitly tied to hazard reduction, unsafe-state avoidance, or the mitigation of failure propagation. 
Likewise, topics such as adversarial attacks, data poisoning, prompt injection, jailbreaks, model exfiltration, and general robot cybersecurity fall outside the independent scope of this survey. While critical, these issues primarily concern security boundaries rather than functional safety.

The literature is organized along two primary axes:
\begin{itemize}
    \item \textbf{Intervention locus}: Identifies where a safety mechanism enters the pipeline—before rollout, during action generation, or during physical execution.
    \item \textbf{Evidence boundary}: Clarifies the scope and limitations of reported safety results.
\end{itemize}
To support these axes, we use \textit{risk pressures} as a descriptive vocabulary to connect specific mechanisms to the safety concerns they address. Figure~\ref{fig:survey_framework} summarizes the intervention locus view, while the following subsections detail the application of this framework across Sections~\ref{sec:section3}--\ref{sec:section7}.

\subsection{Intervention Locus: Where Safety Enters the System}

The first organizing axis of this survey is the intervention locus. We distinguish planning-time, policy-time, and execution-time safety based on the point at which a method shapes, restricts, or repairs behavior. While analytically distinct, these loci are operationally coupled within a single closed-loop embodied system.

(i)~\textbf{Planning-time safety} concerns the formation and validation of the task representation before physical rollout. At this stage, the system determines the task's grounding, constraints, and decomposition. Planning-time safety focuses on formal validation, spatial grounding, and task-and-motion feasibility to reduce the chance that unsafe or ambiguous task specifications are passed downstream. Section~\ref{sec:section3} reviews this layer, moving from grounded specifications to validated long-horizon structures and model-based planning support.

(ii)~\textbf{Policy-time safety} involves shaping action proposals before they are committed to the environment. The focus shifts from plan validity to making the action-generation mechanism more constrained and better aligned. This layer includes constraint injection, preference alignment, and long-horizon structures that preserve consistency across subtask transitions. Its primary role is to reduce the probability of the policy proposing misaligned or procedurally brittle actions. Section~\ref{sec:section4} reviews this layer by treating policy generation as a constrained and objective-shaped decision process.

(iii)~\textbf{Execution-time safety} addresses the system during live interaction with the physical world, where abstract assumptions confront observation noise, distribution shift, and contact uncertainty. This stage encompasses runtime monitoring, anomaly detection, and corrective interventions such as shielding, human handoff, or autonomous recovery. Its role is to mitigate risks left unresolved upstream and to attempt to restore recoverable or safer task progress when the rollout diverges from the intended regime. Section~\ref{sec:section5} reviews this layer, covering runtime risk assessment, task restoration, and physical interaction safety.

\begin{figure}[t]
    \centering
    \includegraphics[width=0.85\linewidth]{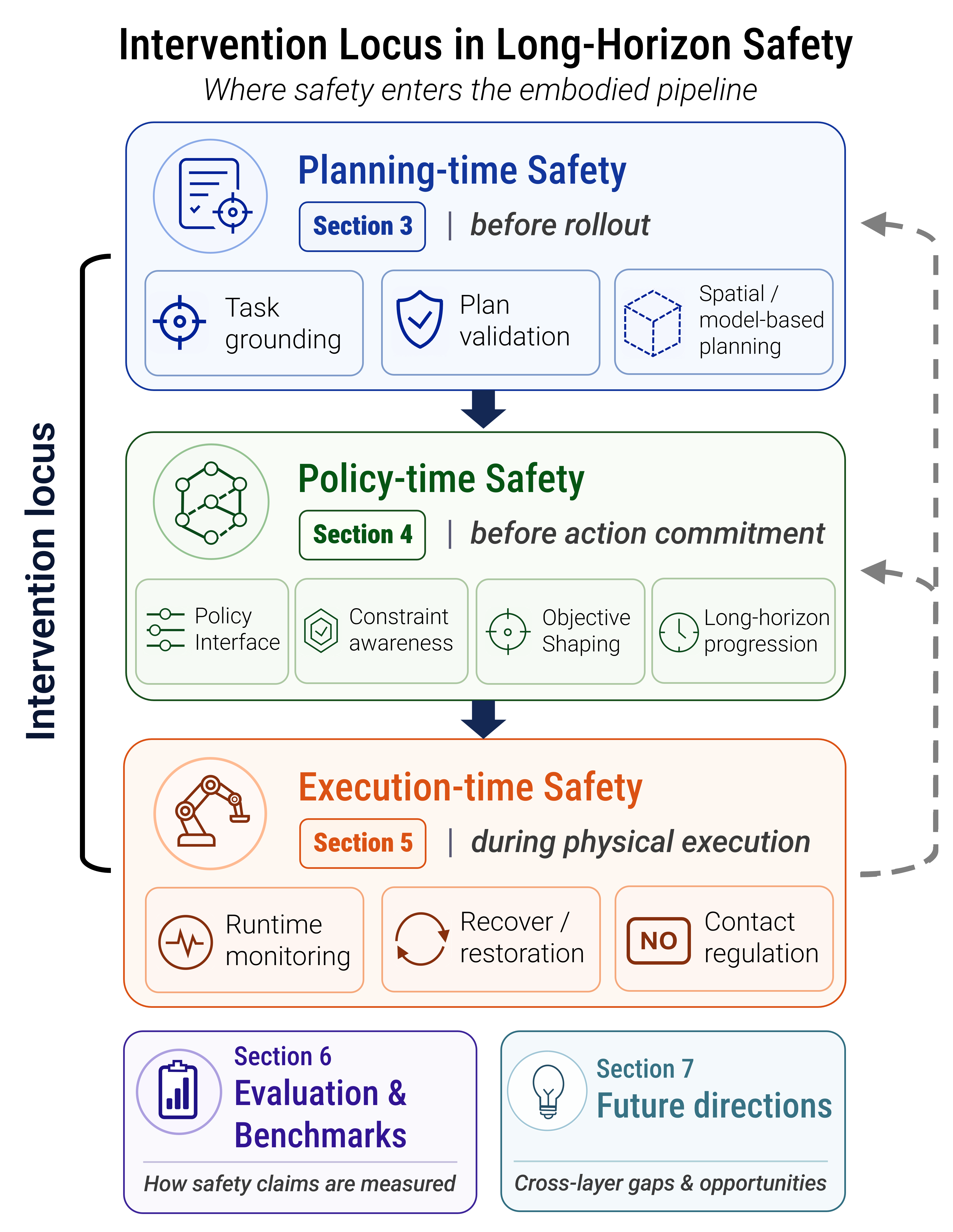}
    \caption{
    Intervention-locus organization of the survey. Planning-time methods shape task representations before rollout, policy-time methods constrain or align action generation before commitment, and execution-time methods monitor, gate, recover, or regulate behavior during physical interaction.
    }
    \label{fig:survey_framework}
\end{figure}

These three loci are not independent modules; they are deeply interconnected. A planning-time constraint must be translatable into a policy interface, and policy-time uncertainty signals must be interpretable by execution-time monitors. For this reason, Fig.~\ref{fig:survey_framework} serves as a survey map rather than a rigid architecture, emphasizing how risk and feedback propagate across the stages of the manipulation lifecycle.

\subsection{Evidence Boundaries: What a Safety Claim Supports}
\label{subsec:evidence_boundaries}

The second primary axis of the survey is the \textit{evidence boundary}. Safety claims in the literature differ not only in their mechanisms but also in the rigor of the evidence supporting them. To avoid overclaiming, this survey distinguishes the \textit{rigor} of the evidence from the \textit{object} to which it applies, using the taxonomy summarized in Table~\ref{tab:evidence_reading_guide}.

In terms of rigor, we identify three primary categories: \textit{formal}, \textit{statistical}, and \textit{empirical} safety evidence. 
As detailed in Table~\ref{tab:evidence_reading_guide}, these categories represent a spectrum of assurance---from the proven correctness of formal methods to the observed outcomes of benchmark-driven empirical studies. 
For consistency, we assign each paper to an evidence category according to the strongest safety-relevant claim that is directly supported by its analysis or validation. 
A work is treated as providing \textit{formal} evidence when it derives a deterministic safety conclusion from explicit assumptions, such as a theorem, certificate, or logical verification of constraint satisfaction. 
It is treated as providing \textit{statistical} evidence when it supports a probabilistic safety statement, such as a calibrated risk estimate, confidence bound, or chance-constraint guarantee tied to a safety-relevant event. 
It is treated as providing \textit{empirical} evidence when safety is supported primarily through observed behavior in simulations, datasets, benchmarks, or case studies without a formal or probabilistic guarantee. 
While formal guarantees offer the strongest local claims, they are inherently bounded by abstraction gaps; conversely, empirical evidence provides practical behavioral insights but lacks inherent generalization to long-tail hazards.

Crucially, we distinguish these safety categories from generic \textit{robustness}. Robustness to noise or distribution shift may support a capability or generalization claim, but it becomes safety evidence only when the perturbations or metrics are explicitly connected to hazards, such as collision, force overload, or failed recovery. Without such a link, robustness should be interpreted as indirect support rather than a verified safety claim.

\begin{table}[t]
\centering
\small
% 열 너비를 비율로 자동 조절 (총합이 딱 \textwidth가 되도록 설정)
\newcolumntype{Y}{>{\hsize=0.6\hsize\linewidth=\hsize\raggedright\arraybackslash}X} % Evidence 열
\newcolumntype{Z}{>{\hsize=1.3\hsize\linewidth=\hsize\raggedright\arraybackslash}X} % What it supports 열
\newcolumntype{W}{>{\hsize=1.1\hsize\linewidth=\hsize\raggedright\arraybackslash}X} % Boundary 열

\begin{tabularx}{\textwidth}{Y Z W}
\toprule
\textbf{Evidence Category} & \textbf{What it Supports} & \textbf{Boundary of the Claim} \\
\midrule

\textbf{Formal guarantees} & 
Proved safety or correctness within an explicit model and constraint set. \newline
\textit{Examples:} CBF/reachability filters, Temporal logic-based planning. 
& 
Invalid outside stated assumptions; inevitable \emph{abstraction gap} from the real world. 
\\

\addlinespace[8pt] % 행 사이 여백으로 가독성 확보

\textbf{Statistical safety} & 
Bounded failure probability or risk-sensitive support under stated assumptions. \newline
\textit{Examples:} Uncertainty bounds, confidence-based interventions. 
& 
Invalid outside stated assumptions or modeled data regimes. \\

\addlinespace[8pt]

\textbf{Empirical safety} & 
Observed safety metrics within specific benchmarks or evaluated scenarios. \newline
\textit{Examples:} Observed reductions in safety cost and contact failures.
& No guarantee of generalization or coverage of long-tail edge cases. \\
\bottomrule
\end{tabularx}
\caption{Taxonomy of evidence categories and their claim boundaries.}
\label{tab:evidence_reading_guide}
\end{table}

The \textit{object of evidence} further defines the scope of these claims across the manipulation lifecycle. \textit{Plan-level evidence} (planning-time) may verify that a task specification is valid or executable in a simulator, but it does not guarantee safe physical execution under perception or policy drift. \textit{Trajectory-level evidence} (policy-time) typically provides post-hoc summaries of rollouts—such as safe-success rates or cumulative costs—which reflect embodied behavior without necessarily measuring the temporal onset of hazards or the precision of intervention timing. Finally, \textit{runtime- or contact-level evidence} (execution-time) focuses on active risk mitigation, such as failure detection and force regulation. While physically closest to safety, this evidence is often highly specific to a particular embodiment and sensor configuration.

Consequently, safety evidence is not interchangeable across layers. A formally verified plan does not imply safe contact, nor does a constrained policy ensure calibrated runtime intervention. Throughout this survey, we evaluate each work by asking: \textit{what safety-relevant mechanism is introduced, and what claim does the evidence actually justify?} This distinction is central to Section~\ref{sec:section6}, which examines how fragmented metrics across these layers can be integrated into a unified evaluation of safe long-horizon manipulation.

\subsection{Representative Safety Concerns Across Layers}
\label{subsec:risk_pressures}

In addition to the intervention locus and evidence boundary, we identify a set of representative safety concerns that define the problems different mechanisms aim to address. It is crucial to distinguish these \textit{safety concerns}—the underlying conditions that can lead to failure—from \textit{safety mechanisms}, such as monitoring, shielding, or recovery, which are the system's responses to those conditions. A mechanism's failure matters only because it allows an underlying hazard to remain unmitigated or unrecoverable.

Across Sections~\ref{sec:section3}--\ref{sec:section5}, we categorize the recurring safety concerns as follows:
\begin{itemize}
    \item \textit{Task specification and grounding risk}: Errors in goal interpretation, scene grounding, or the omission of safety-relevant constraints.
    \item \textit{Sequence and transition risk}: Incorrect decomposition or unsafe ordering of subtasks, causing the system to proceed under unmet preconditions.
    \item \textit{Spatial and motion-feasibility risk}: Abstract plans that lack geometric commitments, such as collision-free trajectories, reachable poses, or stable grasp sites.
    \item \textit{Policy generation and objective risk}: Unsafe action proposals stemming from misaligned objectives, unconstrained action interfaces, or loss of long-horizon context.
    \item \textit{Runtime drift and uncertainty risk}: Departures from planning assumptions due to perception errors, distribution shifts, or miscalibrated confidence during rollout.
    \item \textit{Contact and physical-interaction risk}: Hazards arising from force, friction, or jamming that can damage objects, harm humans, or cause irreversible task degradation.
\end{itemize}

These concerns are coupled across the long-horizon rollout through \textit{cascading failure} and the \textit{loss of recoverability}. A minor specification error or contact disturbance can propagate across layers, eventually leading to a state from which may no longer reliably prevent harm or preserve recoverability. Recovery and restoration mechanisms specifically target this loss of recoverability, aiming to preserve the possibility of safe continuation or graceful termination.

This framing clarifies the roles of the safety mechanisms. For instance, in execution-time safety mechanisms reviewed in Section~\ref{sec:section5}, while monitoring and diagnosis make runtime risk observable, shielding and contact regulation directly mitigate physical harm. By mapping mechanisms to these specific concerns, the survey evaluates not only where a method intervenes but also the precise nature of the hazard it seeks to resolve.

\subsection{Framework-Guided Organization of the Survey}
\label{subsec:framework_guided_organization}

The survey follows the cross-layer framework introduced above, organizing the literature by the locus at which safety assumptions are formed, constrained, or revised. Rather than treating sections as independent blocks, we categorize them by their \textit{safety object}, representative \textit{concerns}, and typical \textit{evidence} (Table~\ref{tab:lifecycle_mechanisms}).

Section~\ref{sec:section3} focuses on \textit{planning-time safety}, where safety assumptions or constraints are introduced before rollout. The safety objects are task representations—such as goal grounding and temporal specifications—while concerns involve specification errors and feasibility gaps. Evidence here is primarily \textit{plan-level} (e.g., formal validation, subtask specification), providing pre-execution assurance but lacking robustness against real-world perception or policy drift.

Section~\ref{sec:section4} examines \textit{policy-time safety}, targeting the action generation mechanism, including constrained policy generation and preference alignment. Key risks include objective misalignment and procedural drift. Evidence is typically \textit{trajectory-level} (e.g., safe-success rates, violation costs), which reflects embodied behavior but remains post-hoc and often fails to capture the precise onset or severity of risks.

Section~\ref{sec:section5} addresses \textit{execution-time safety} during real-time interaction. Objects include uncertainty estimates, shielding actions, and recovery policies. Risks center on runtime drift and contact-related hazards. Evidence involves \textit{interaction-level} metrics like intervention efficiency and force regulation. While closest to physical safety, this layer is highly dependent on embodiment-specific sensing and calibration.

Section~\ref{sec:section6} discusses \textit{evaluation and benchmarks}, focusing on how safety claims are measured. It distinguishes nominal capability from safe execution, emphasizing the need for \textit{procedural evidence}—not just whether a task succeeded, but which layer detected risks and how the system intervened.

% Section~\ref{sec:section6} examines \textbf{evaluation and benchmarks}. This section does not introduce a fourth intervention locus. Instead, it asks how safety claims from planning-time, policy-time, and execution-time mechanisms are actually measured. Existing manipulation benchmarks have largely standardized capability evaluation, but safety-aware evaluation remains fragmented across plan-level risk screening, rollout-level safe success, runtime failure detection, recovery outcome, and contact-quality metrics. Section~\ref{sec:section6} therefore clarifies what current benchmarks can and cannot prove. Its central role is to distinguish nominal task success from safe execution, and to show why long-horizon manipulation safety requires procedural evidence: what risk appeared, when it appeared, which layer detected it, how the system intervened, and whether safe continuation or restoration was achieved.

Section~\ref{sec:section7} identifies \textit{future directions} by examining the interfaces between layers. It addresses challenges such as abstraction transfer, sim-to-real safety, and multimodal assurance. This organization highlights that many works span multiple stages; such overlap is not a classification error but a fundamental property of long-horizon manipulation safety, where information must be preserved across planning, policy, and execution.

% Section~\ref{sec:section7} identifies \textbf{future directions and opportunities} that arise from the limitations of layer-local safety evidence. The relevant object is no longer one mechanism, but the interface between mechanisms. Planning-time constraints may be weakened when translated into policy actions; policy-time alignment may fail under contact-rich execution; runtime monitors may operate outside their calibration regime; and evaluation protocols may hide the procedural history of near misses, interventions, and unsafe successes. Section~\ref{sec:section7} therefore focuses on abstraction transfer, sim-to-real safety evidence, embodiment transfer, calibrated intervention selection, procedural safety observability, semantic and multimodal safety, deployment assurance, and safety-oriented data and simulation infrastructure.
% Table~\ref{tab:lifecycle_mechanisms} summarizes this framework-guided organization.

\newcolumntype{L}[1]{>{\raggedright\arraybackslash\hspace{0pt}}p{#1}}
\newcolumntype{Y}{>{\raggedright\arraybackslash\hspace{0pt}}X}

\begin{sidewaystable}[p]
\centering
\scriptsize
\setlength{\tabcolsep}{5pt}
\renewcommand{\arraystretch}{1.18}

\begin{adjustbox}{max totalsize={\textheight}{0.94\textwidth},center}
\begin{tabularx}{\linewidth}{L{1.5cm} L{3.95cm} Y Y L{2.05cm} L{3.55cm}}
\toprule
\textbf{Stage} & \textbf{Mechanism Family} & \textbf{Typical Role} & \textbf{Primary Risk / Failure Mode} & \textbf{Evidence Level} & \textbf{Rep. Citations} \\
\midrule

\addlinespace[2pt]
\multirow{3}{*}{\makecell[l]{\textbf{Planning}\\\textbf{(Sec. 3)}}}
& \makecell[tl]{\textbf{3.1 Grounding Goals}\\ \textbf{and Task}\\ \textbf{Specifications}}
& Grounding of language, goals, constraints, and scene states.
% Grounds goals, initial states, constraints, and executable task intent before rollout.
& Semantic/specification risk, grounding failure, omitted constraints.
% Semantic/specification risk, grounding failure, omitted constraints, interface mismatch.
& Empirical
& \citep{saycan,singh2023progpromptgeneratingsituated,shirai2024visionlanguageinterpreter,liang2023codeaspolicies} \\ 
% \citep{saycan,innermonologue,huang2023voxposercomposable3d,shirai2024visionlanguageinterpreter}  \\
\addlinespace[5pt]
\cmidrule(lr){2-6}

\addlinespace[1pt]
& \makecell[tl]{\textbf{3.2 Structuring and}\\ \textbf{Validating Long-}\\ \textbf{horizon Plans}}
& 
% Decomposes tasks, encodes formal constraints, and validates executable plans against symbolic specifications.
Task decomposition, logical constraint encoding, and pre-execution plan validation.
& Task-planning failure, specification error, invalid ordering, transition mismatch.
& Formal / statistical / empirical
& \citep{yang2024plugsafetychip,wang2025conformaltemporallogic,lee2025veriplanintegratingformal,wu2025selpgeneratingsafe} \\
\addlinespace[5pt]
\cmidrule(lr){2-6}

\addlinespace[1pt]
& \makecell[tl]{\textbf{3.3 Spatial and}\\ \textbf{Model-based}\\ \textbf{Planning Support}}
& 
Spatial grounding, object-centric constraints, and model-based planning support.
% Bridges high-level instructions to spatial, object-centric, and trajectory-level geometries before execution.
& Spatial infeasibility, collision-prone plans.
& Empirical
& \citep{huang2023voxposercomposable3d,huang2024rekepspatiotemporal,garrett2020pddlstreamintegratingsymbolic,huang2024copageneralrobotic}  \\
\addlinespace[5pt]
\midrule

\addlinespace[2pt]
\multirow{2}{*}{\makecell[l]{\textbf{Policy}\\\textbf{(Sec. 4)}}}
& \makecell[tl]{\textbf{4.1 Policy Class,}\\ \textbf{Interfaces, and}\\ \textbf{Long-horizon Context}}
& Policy class and intervention interfaces definition rather than a safety mechanism.
& N/A (Establishes policy framework rather than mitigation)
& Context
& \citep{brohan2023rtroboticstransformer,chi2023diffusionpolicyvisuomotor,liang2023codeaspolicies} \\
\addlinespace[5pt]
\cmidrule(lr){2-6}

\addlinespace[1pt]
& \makecell[tl]{\textbf{4.2 Constraint-aware}\\ \textbf{Policy Generation}}
& Policy constraint injection, constrained decoding, or safe optimization.
& Unsafe action selection, incomplete constraints, over- or under-filtering.
& Formal / statistical / empirical
& \citep{achiam2017cpo,brunke2022safetyfilters,yang2024plugsafetychip, wang2025conformaltemporallogic, zhang2025safevla} \\
\addlinespace[5pt]
\bottomrule
\end{tabularx}
\end{adjustbox}

\caption{Lifecycle intervention map for long-horizon robotic manipulation, covering the main safety mechanisms reviewed in Sections~3--5. Rows follow the manuscript's current subsection structure and summarize each mechanism family's intervention role, primary risk emphasis, and dominant evidence profile. 
% Evaluation and assurance are treated separately in Section~\ref{sec:section6} because they assess the evidential strength of safety claims rather than constituting a direct intervention locus.
}
\label{tab:lifecycle_mechanisms}
\end{sidewaystable}

\begin{sidewaystable}[p]
\ContinuedFloat
\centering
\scriptsize
\setlength{\tabcolsep}{5pt}
\renewcommand{\arraystretch}{1.18}

\begin{adjustbox}{max totalsize={\textheight}{0.94\textwidth},center}
\begin{tabularx}{\linewidth}{L{1.5cm} L{3.95cm} Y Y L{2.05cm} L{3.55cm}}
\toprule
\textbf{Stage} & \textbf{Mechanism Family} & \textbf{Typical Role} & \textbf{Primary Risk / Failure Mode} & \textbf{Evidence Level} & \textbf{Rep. Citations} \\
\midrule

\addlinespace[2pt]
\multirow{2}{*}{\makecell[l]{\textbf{Policy}\\\textbf{(Sec. 4)}}}
& \makecell[tl]{\textbf{4.3 Alignment and}\\ \textbf{Objective Shaping}}
& Objective shaping from preferences, rewards, and language feedback.
& Semantic misalignment, reward hacking, insufficient feedback information.
& Empirical
& \citep{moletta2026preferencealignedvisuomotor,hu2025flareachievingmasterful,niu2025deepreinforcementlearning} \\
% & \citep{niu2025deepreinforcementlearning,hu2025flareachievingmasterful,moletta2026preferencealignedvisuomotor} \\
\addlinespace[5pt]
\cmidrule(lr){2-6}

\addlinespace[1pt]
& \makecell[tl]{\textbf{4.4 Long-horizon}\\ \textbf{Extensions of}\\ \textbf{Policy-time Safety}}
& Progress-aware structure, stage-aware rewards, and delayed-risk mitigation.
& Premature transitions, skipped prerequisites, procedural drift.
& Empirical
& \citep{saycan,liang2023codeaspolicies,hu2025flareachievingmasterful,lu2026isbenchevaluating} \\
\addlinespace[5pt]
\midrule

\addlinespace[2pt]
\multirow{3}{*}{\makecell[l]{\textbf{Execution}\\\textbf{(Sec. 5)}}}
& \makecell[tl]{\textbf{5.1 Runtime Risk}\\ \textbf{Assessment}}
& Anomaly detection, failure diagnosis, and steering policies during rollout.
& Hidden failure risk, miscalibrated confidence, unsafe continuation.
& Formal / statistical / empirical
& \citep{sinha2024realtimeanomaly,xu2025canwedetect,wajid2025successivecontrolbarrier,wu2025fromforesightforethought} \\
\addlinespace[5pt]
\cmidrule(lr){2-6}

\addlinespace[1pt]
& \makecell[tl]{\textbf{5.2 Runtime Adaptation}\\ \textbf{and Task Restoration}}
& Task restoration via handoff, correction, replanning, recovery.
& Imminent collision, stalled progress, and subtask-transition failure.
& Empirical 
& \citep{hung2021introspectivevisuomotorcontrol,raman2024capecorrectiveactions,dai2025racerrichlanguage} \\
\addlinespace[5pt]
\cmidrule(lr){2-6}

\addlinespace[1pt]
& \makecell[tl]{\textbf{5.3 Physical Interaction}\\ \textbf{Safety Under Contact}}
& Compliance, force-limit regulation, and multi-timescale contact correction.
& Force overload, unstable contact, irreversible damage.
& Formal / empirical
&  \citep{tang2023formalverificationrobotic,brunke2025semanticallysaferobot,wang2025guarding,ankile2025fromimitationrefinement} \\
\addlinespace[5pt]
\bottomrule
\end{tabularx}
\end{adjustbox}

\caption[]{(continued)}
\end{sidewaystable}

The following section begins with planning-time safety, where safety first enters the lifecycle through task grounding, constraint interpretation, plan structuring, formal validation, and spatial or model-based support.

\section{Planning-time Safety} \label{sec:section3}

Planning-time safety concerns failures that begin before behavior is ever executed, when semantic or specification mistakes are first transformed into downstream physical risk. At this stage, the system is deciding what task is actually being asked for, which constraints define that task, and whether the resulting plan structure is coherent enough to deserve execution. Errors at this layer can propagate in two distinct but connected ways. Mistakes in goal interpretation or constraint formation can yield the wrong task specification, while mistakes in plan structure can produce an invalid or unsafe task sequence. 
Planning-time safety therefore concerns where safety must intervene before execution begins, and which planning objects must be made correct, inspectable, and executable early enough to prevent later physical failure.

\begin{figure}[t!]
    \centering
    \includegraphics[width=\textwidth]{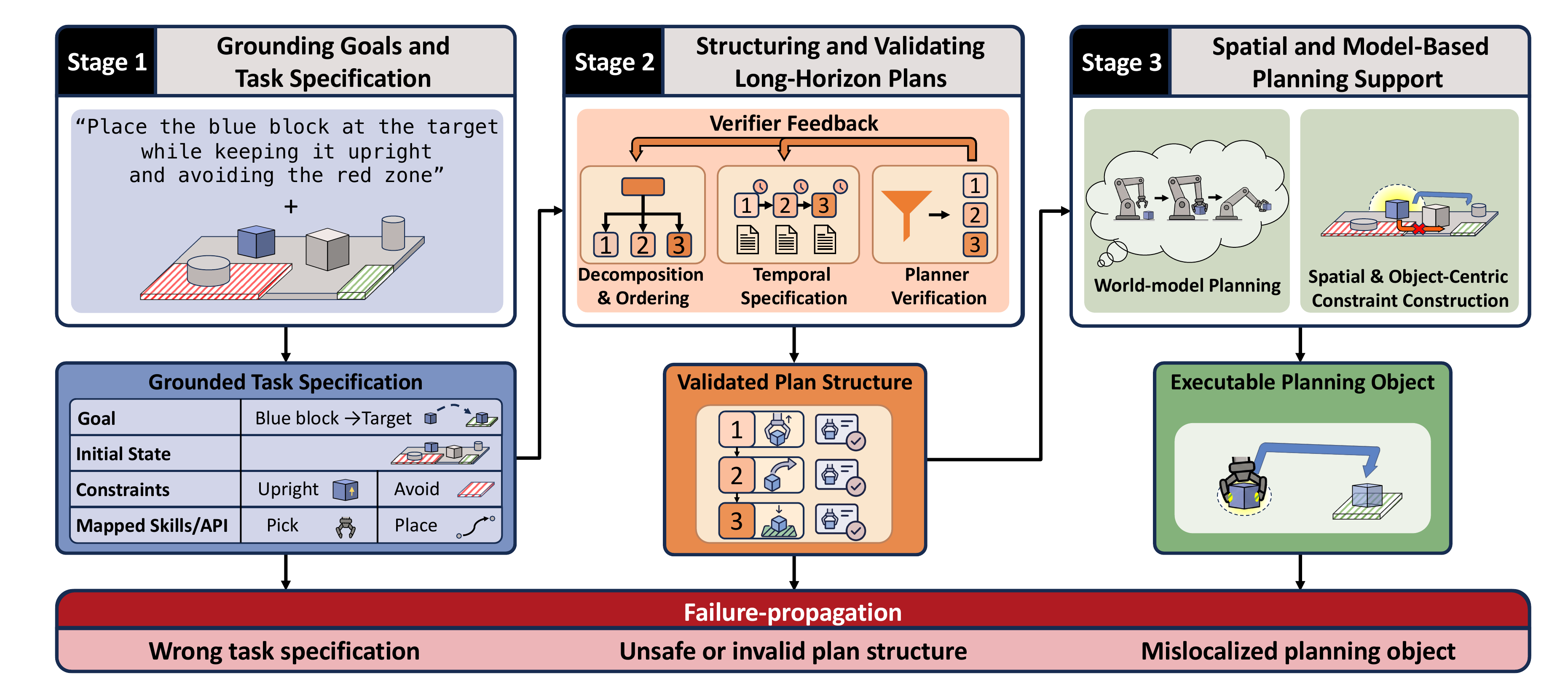}
    \caption{Overview of planning-time safety. Section~\ref{sec3:grounding} grounds goals, initial states, constraints, and executable task intent into inspectable planning objects. Section~\ref{sec3:plan_structuring} then organizes those objects into a long-horizon plan structure through decomposition, verification, and executability checks. Section~\ref{sec3:spatial_model_support} finally refines the validated plan into a more scene-grounded and motion-feasible planning object.}
    \label{fig:section3_overview}
\end{figure}

We organize the rest of this section as follows. Section~\ref{sec3:grounding} examines the grounding assumptions required for safe rollout, including how goals, initial states, constraints, and executable task intent are formed. Section~\ref{sec3:plan_structuring} then studies how a grounded task is turned into a long-horizon plan structure through decomposition, specification, verification, and executability checks. Section~\ref{sec3:spatial_model_support} finally considers how that validated plan is brought closer to the physical world through future-state prediction, scene-grounded spatial commitments, and motion-feasibility reasoning before execution. Figure~\ref{fig:section3_overview} summarizes this progression by illustrating the main planning objects and failure-sensitive interfaces across grounding, plan structuring, and spatial/model-based support.

In the terminology defined in Section~\ref{sec:section2}, this section primarily addresses semantic and procedural safety: unsafe rollout may begin when goals, constraints, preconditions, or plan transitions are represented incorrectly. Physical safety appears here mainly as a downstream boundary, because plan-level validity does not by itself certify safe motion, contact, or force interaction.

%%%%%%%%%%%%%%%%%%%%%%%%%%%%%%%%%%%%%%%%%%%%%%%%%%%%%
\subsection{Grounding Goals and Task Specifications}\label{sec3:grounding}
%%%%%%%%%%%%%%%%%%%%%%%%%%%%%%%%%%%%%%%%%%%%%%%%%%%%%

For rollout to be safe, the system must first ground the goal, initial world state, and relevant constraints into a task specification that downstream planning modules can actually use. If goals, states, or constraints are grounded incorrectly, then later correctness checks and runtime safety mechanisms are already operating over the wrong task object from the beginning. This subsection therefore moves from goal and initial-state grounding, to deciding which constraints belong in the task specification, and finally to mapping grounded task intent into executable interfaces.

\subsubsection{Goal and Initial-state Grounding}\label{sec3:goal_state_grounding}

Goal and initial-state grounding is the stage at which the planner aligns what must be achieved with the world state from which rollout will begin. Planning-time safety here depends on correctly identifying both the task intent and the starting context. However, this alignment must be strongly feasibility-aware, as a linguistically plausible interpretation is not necessarily executable. Affordance-aware grounding must therefore evaluate not just whether an inferred goal makes semantic sense, but whether it is physically achievable given the current world state and available skill set. In this sense, goal grounding serves a dual purpose: it acts as both intent parsing and an early filtering step to determine the most viable interpretation.

To address this challenge, contemporary frameworks deploy two representative grounding mechanisms: implicit affordance scoring, which balances semantic skill usefulness against learned, state-conditioned feasibility probabilities~\citep{saycan}, and explicit symbolic structuring, which translates joint language-scene observations into formal, inspectable problem definitions specifying object sets and initial conditions~\citep{shirai2024visionlanguageinterpreter}. These approaches can be dynamically extended by incorporating closed-loop textual feedback from the environment to iteratively update and re-align the active interpretation during rollout~\citep{innermonologue}. The evidence boundary of this entire cluster is strictly confined to establishing semantic compatibility and symbolic consistency at discrete decision boundaries; even with textual tracking, these methods only provide evidence that a plan remains logically consistent with high-level perception information.

The definitive open gap stems from the underlying assumption that discrete semantic abstractions can sufficiently capture the continuous topological feasibility of the environment. These frameworks inherently assume that high-level perception tokens (e.g., object predicates and scene graphs) provide a loss-free representation for task-level reasoning. When this assumption is violated—such as when complex geometric blocking or object clutter is simplified into a binary \enquote{reachable} predicate—the planning-time grounding framework becomes decoupled from physical reality. Because the task planner operates purely on this logical information, it may generate a semantically flawless high-level sequence that the downstream motion planner repeatedly rejects at the geometric level. Consequently, the system remains vulnerable to severe planning-time failures, becoming trapped in infinite symbolic re-planning loops or total task deadlocks before any physical rollout can even be initiated.

\subsubsection{Constraint Interpretation and Specification Formation}\label{sec3:constraint_spec}
%%%%%%%%%%%%%%%%%%%%%%%%%%%%%%%%%%%%%%%%%%%%%%%%%%%%%

Once the goal and initial state have been grounded, the system must determine which constraints govern the task execution. While the previous step focused on \textit{what} must be achieved, this stage specifies which restrictions should shape \textit{how} it is achieved. The core challenge is not generic instruction following, but translation: converting a natural-language restriction into a planner-usable representation. If a constraint remains a free-form instruction, downstream planning modules cannot reliably reason about what must be prohibited or preserved. To enforce safety, interpreted constraints should become explicit, inspectable task objects that directly modify the planning stack.

To bridge this linguistic-symbolic gap, contemporary frameworks employ distinct representational regimes that transform abstract instructions into dense geometric value maps~\citep{huang2023voxposercomposable3d}, structured symbolic problem descriptions~\citep{shirai2024visionlanguageinterpreter}, or explicit prohibitive criteria for high-level plan filtering~\citep{yang2024plugsafetychip}. The evidence boundary of these mechanisms is strictly limited to representational compliance; 
they aim to map natural-language restrictions into structured specifications that downstream planners can interpret.
% they ensure that natural-language restrictions can be faithfully mapped into mathematically structured specifications that downstream planners can interpret.

The critical open gap, however, lies in the inherent trade-off between logical expressivity, solver tractability, and semantic alignment. If this formalization pipeline induces an over-constraint scenario—where multiple qualitative safety rules are translated into rigid, mutually exclusive logical formulas—the downstream optimization or symbolic solver will fail to find a valid solution space. Conversely, in under-constraint scenarios, if the translation pipeline misses implicit safety invariants or contextual nuances, the downstream planner can mathematically satisfy the narrow constraint specification while generating an abstract task sequence that fundamentally violates the user's true safety intent.

% Recent methods achieve this translation through distinct representational regimes. VoxPoser~\citep{huang2023voxposercomposable3d} represents a spatial approach, inferring constraints from language and rendering them as explicit 3D cost or value maps, which are then directly consumed by a model-based planner. 
% Alternatively, a symbolic approach embeds these restrictions directly into a structured planning problem description~\citep{shirai2024visionlanguageinterpreter}.
% % Building on the alignment discussed in Section~\ref{sec3:goal_state_grounding}, ViLaIn~\citep{shirai2024visionlanguageinterpreter} takes a more symbolic approach, embedding these restrictions directly into its structured planning problem description.
% Finally, Plug in the Safety Chip~\citep{yang2024plugsafetychip} illustrates why natural-language \texttt{don't} constraints must be formalized into explicit task restrictions before any downstream action filtering or formal blocking can occur.

Together, these approaches show that planning-time safety depends on making constraints explicit, inspectable, and planner-usable. However, correctly translating a restriction into a specification object does not by itself guarantee that the planner will find a tractable solution. This representational handoff bridges directly into the formal specification, verification, and enforcement machinery covered next in Section~\ref{sec3:plan_structuring}.

% Planner-facing specification formation matters because interpreted constraints must eventually become task objects that actually modify the planning stack. Even when a constraint has been understood and made inspectable, downstream planning still cannot act on it unless that result is packaged as a specification object the planner can directly use. ViLaIn~\citep{shirai2024visionlanguageinterpreter} is the most direct example on this point because it organizes instructions and scene observations into a structured planning problem description that the planner can consume as input. VoxPoser~\citep{huang2023voxposercomposable3d} shows that the same function can be achieved in a different representational regime, where inferred affordance and constraint maps still serve as directly planning-relevant objects. Plug in the Safety Chip~\citep{yang2024plugsafetychip} closes this cluster by making clear why natural-language ``don't'' constraints must first be turned into explicit task restrictions before downstream formal blocking or action filtering can occur. The result is a direct handoff from constraint interpretation in Section~\ref{sec3:grounding} to later specification enforcement in Section~\ref{sec3:plan_structuring}. The supported claim therefore remains representational: these methods help make constraints explicit, inspectable, and planner-usable, but they do not by themselves guarantee that the resulting specification will be enforced correctly or remain safe through later planning and execution.

%%%%%%%%%%%%%%%%%%%%%%%%%%%%%%%%%%%%%%%%%%%%%%%%%%%%%
\subsubsection{Grounding Task Intent to Executable Interfaces}\label{sec3:exec_interfaces}
%%%%%%%%%%%%%%%%%%%%%%%%%%%%%%%%%%%%%%%%%%%%%%%%%%%%%

Even after goals and constraints have been grounded, the system must still map the resulting task intent into interfaces the robot can actually invoke. While the previous subsections focused on determining \textit{what} belongs in the task specification, the focus here shifts to \textit{how} that specification is expressed. At this stage, executability is defined strictly at the semantic and action vocabulary level. The objective is to ensure that an abstract plan step corresponds to an available skill, API call, or control primitive, rather than verifying whether the resulting motion is geometrically feasible or physically stable. This translation layer is crucial because a semantically plausible plan is useless if it cannot be lowered into the robot's specific action space.

To bridge this gap between high-level reasoning and physical capability, contemporary frameworks employ two primary translation mechanisms that transition from discrete vocabulary mapping to structured programmatic encapsulation. The most straightforward approach establishes this alignment through textual similarity mapping, translating free-form language proposals into a restricted set of admissible actions, increasing the chance that generated steps match predefined primitive commands~\citep{huang2022languagemodelsas}. 
Moving beyond post-hoc vocabulary matching, a programmatic paradigm exposes the robot's available action APIs, object lists, and assertion-like state checks directly inside the generation prompt, allowing the model to express task plans as executable code-like action sequences~\citep{singh2023progpromptgeneratingsituated}.

The evidence boundary of this interface-mapping layer is strictly confined to syntactic compliance and API compatibility. The critical open gap lies in the underlying assumption of complete alignment between the abstract planner and the deterministic API signatures. Because these interfaces operate purely on semantic or programmatic abstractions, they cannot dynamically evaluate the dense, context-dependent preconditions hidden behind a static function definition. When this assumption is violated, the interface layer suffers from severe logical failures. Specifically, the system remains highly vulnerable to triggering uncaught symbolic exceptions, state-tracking desynchronizations, or fatal programmatic deadlocks at the task-planning level, halting the execution pipeline before any continuous motion can even be computed.

\subsection{Structuring and Validating Long-horizon Plans}\label{sec3:plan_structuring}
%%%%%%%%%%%%%%%%%%%%%%%%%%%%%%%%%%%%%%%%%%%%%%%%%%%%%

While Section~\ref{sec3:grounding} established \textit{what} must be grounded as the task specification, this section addresses \textit{how} that specification is organized into a verifiable long-horizon plan. Even a perfectly grounded task can result in structural failure or logical deadlock if its subgoals are not coherently sequenced, if temporal requirements remain underspecified, or if validation checks do not meaningfully constrain the plan. The discussion therefore moves from subtask decomposition, to formal spatiotemporal specification, and finally to pre-execution verification.

% If Section~\ref{sec3:grounding} concerns what must be grounded as the task object, Section~\ref{sec3:plan_structuring} concerns how that grounded task becomes a long-horizon plan structure that can be checked, specified, and verified before execution. Even when the grounded task description is correct, physical failure can still result if the task is not organized into a checkable plan structure, if temporal or geometric requirements remain underspecified, or if validation does not meaningfully constrain the plan. This subsection therefore moves from subgoal structuring, to temporal or spatio-temporal specification, and then to verification and formal feedback before execution.

%%%%%%%%%%%%%%%%%%%%%%%%%%%%%%%%%%%%%%%%%%%%%%%%%%%%%
\subsubsection{Task Decomposition and Subtask Ordering}\label{sec3:decomposition}
%%%%%%%%%%%%%%%%%%%%%%%%%%%%%%%%%%%%%%%%%%%%%%%%%%%%%

Task decomposition divides a long-horizon objective into a structured sequence of subgoals, while ordering defines the critical dependencies and transition conditions between them. If a task is decomposed incorrectly or linked out of order, the downstream control stack inherits an inherently flawed plan. Safety at this stage, therefore, depends on organizing subgoals coherently enough to support rigorous pre-execution validation before any low-level trajectory generation begins.

To achieve this logical coherence, contemporary frameworks deploy three representative structuring and screening mechanisms that bridge linguistic reasoning with formal logic. The first mechanism utilizes prompt-level logical verification, employing chain-of-thought reasoners to evaluate fine-grained invariants, preconditions, and postconditions directly during the text-generation phase to filter out flawed plan structures~\citep{obi2025safeplanleveragingformal}. Expanding this textual reasoning into mathematical boundaries, a second mechanism leverages automata-driven temporal pruning, translating natural-language constraints into formal Linear Temporal Logic (LTL) specifications to actively monitor state transitions and prune inadmissible actions from the candidate action space~\citep{yang2024plugsafetychip}. 
A third layer grounds decomposition in a symbolic planning substrate: scene-graph-based environment representations and task descriptions can be converted into PDDL domain and problem files, after which long-horizon goals are decomposed into subgoals that are solved autoregressively by an automated planner~\citep{liu2025deltadecomposedefficient}. 
% Rather than relying solely on language-to-logic conversion, a third mechanism enforces neuro-symbolic decomposition, anchoring subgoal sequences within structured environmental topologies by mapping textual scene graphs into formal PDDL problem files solved via classical task planners~\citep{liu2025deltadecomposedefficient}. 
The evidence boundary of this decomposition cluster is strictly confined to causal consistency within the assumed symbolic domain model; these methods check that the macro-sequence of subgoals is logically sound and adheres to temporal rules, provided the environment's true states are perfectly mirrored by the planner's predicates.

The definitive open gap, however, is the underlying assumption of predicate completeness and static causal invariance. Because these verifiers operate purely on high-level symbolic and semantic abstractions, they cannot reason about hidden, unmodeled state dependencies or non-obvious physical parameters. When this assumption fails—such as when a cabinet door appears accessible in the scene graph but is internally jammed, or when an object's weight exceeds the implicit capability threshold of a selected skill primitive—the task planning framework suffers from a severe semantic abstraction mismatch.

% Recent frameworks tackle this by framing decomposition as a problem of logical validity and unsafe-sequence interception. One method for strengthening the logical coherence of a generated sequence involves enforcing invariants, preconditions, and postconditions alongside prompt-level sanity checks prior to refinement or execution \citep{obi2025safeplanleveragingformal}.
% Furthermore, the introduction of explicit safety constraints—such as those derived from translating natural-language instructions into Linear Temporal Logic (LTL)—allows for formal reasoning over potential violations and the pruning of unsafe actions from the admissible subtask space \citep{yang2024plugsafetychip}. Another approach to improving planning-stage feasibility and efficiency involves anchoring subgoal decomposition in structured world representations, such as scene graphs combined with Planning Domain Definition Language (PDDL) descriptions \citep{liu2025deltadecomposedefficient}. Collectively, these methodologies emphasize the importance of admissible ordering and early sequence screening; however, a logically validated sequence does not inherently guarantee executability under subsequent geometric, contact, or deployment-time disturbances.

%%%%%%%%%%%%%%%%%%%%%%%%%%%%%%%%%%%%%%%%%%%%%%%%%%%%%
\subsubsection{Temporal and Spatio-temporal Specification}\label{sec3:temporal_spec}
%%%%%%%%%%%%%%%%%%%%%%%%%%%%%%%%%%%%%%%%%%%%%%%%%%%%%

While the previous subsection organized a task into sequential subgoals, this stage focuses on making the constraints governing those subgoals formally checkable. Verifying a long-horizon plan requires translating natural-language requirements into explicit temporal, logical, or geometric conditions. Temporal logics, such as LTL or Signal Temporal Logic (STL), provide the standard mathematical language for this transformation, offering rigorous operators for sequencing, prohibition, and obligation that inherit a rich lineage from classical correct-by-construction hybrid control~\citep{kressgazit2009temporallogicbased}. In modern language-driven pipelines, however, this translation step introduces a critical front-end bottleneck: if the system maps an instruction into an incorrect formal specification, the downstream planner will faithfully optimize for the wrong task.

% While the previous subsection organized a task into subgoals, this stage focuses on making the constraints governing those subgoals formally checkable. To verify a plan, natural-language requirements must be translated into explicit temporal, logical, or geometric conditions. \textbf{Temporal logic}, such as LTL or Signal Temporal Logic (STL), provides the standard formal framework for this, offering explicit operators for temporal ordering, prohibition, and obligation. 
% Classical approaches in this domain demonstrate how such reactive task specifications can be successfully compiled into correct-by-construction hybrid controllers \citep{kressgazit2009temporallogicbased}.

To reduce this front-end vulnerability, recent language-driven systems increasingly separate the specification problem into three complementary safeguards: syntactic validity, calibrated semantic confidence, and geometric structure. 
A first safeguard is to constrain the language-to-logic interface so that the generated formula is at least syntactically admissible by the downstream verifier or planner. 
Earlier approaches learn this mapping from weak supervision, such as sentence--trajectory pairs, while more recent LLM-based pipelines use code-generation prompts to improve the syntactic correctness of the resulting LTL formula~\citep{patel2020groundinglanguagenon,rabiei2025ltlcodegencodegeneration}. 
However, syntactic validity alone does not imply that the formula captures the intended task. 
A second safeguard therefore treats the translation itself as an uncertain decision process: natural-language requirements can be decomposed into intermediate question-answering steps, and conformal prediction can be used to proceed only when the generated logical decision is sufficiently reliable, or otherwise request assistance~\citep{conformalnl2ltl2025}. 
% Related temporal-logic planning pipelines use conformal calibration at the planner--LLM interface to manage uncertainty when natural-language subtasks are embedded inside formal task structures~\citep{wang2025conformaltemporallogic}. 
A third safeguard becomes necessary in manipulation, where the relevant constraints are not only temporal but also spatial, geometric, and relational. In this setting, natural-language instructions can be translated into hierarchical spatio-temporal logic structures that encode object-centric geometric relations before being rendered into executable formal specifications~\citep{luo2025nl2spatialgeneratinggeometric}.

These safeguards improve the reliability of the language-to-specification interface, but several failure modes remain. One such failure mode is vacuous satisfaction: an implication-style requirement may be satisfied because its triggering condition never occurs, rather than because the intended response is achieved~\citep{kupferman2003vacuity}. 
For a robot, this means that a formally valid plan can still miss the user's intended safety semantics when environmental assumptions, object availability, or trigger conditions are mismatched. 
Another failure mode arises after the specification has been constructed. 
Temporal logic planning is computationally demanding even in classical settings, with LTL reasoning already exhibiting PSPACE-complete worst-case complexity for common formulations~\citep{sistla1985complexity}. 
In embodied domains, this difficulty is further amplified by geometric abstraction and long-horizon composition.

\subsubsection{Planner Verification and Formal Feedback}\label{sec3:verification}
%%%%%%%%%%%%%%%%%%%%%%%%%%%%%%%%%%%%%%%%%%%%%%%%%%%%%
% \textcolor{red}{앞 내용과의 연결성을 위해, goal spec grounding, generated constraints를 만족시키는 planning 결과가 만들어졌는지를 verify하는 로직 혹은 모듈의 필요성을 강조하는 문장. 그리고 그 뒤에 verification을 넘어 실제로 planning ,과정의 feedback으로서도 기능할 수 있음을 언급?}

Once goals, constraints, and action interfaces have been formalized into inspectable specifications, the planning system must verify whether a candidate action sequence satisfies these established parameters before it reaches the physical execution stack. This verification stage serves a critical dual purpose: it acts as a pre-execution gatekeeper to screen plan compliance and provides structured feedback to guide iterative refinement. Crucially, contemporary frameworks operationalize this machinery by transitioning from post-hoc screening to active decoding constraints, and ultimately to closed-loop corrective feedback. In its foundational form, verification can serve as a pre-execution screening layer, using formal-logic-inspired invariant, precondition, and postcondition checks to examine, refine, or reject structurally invalid or specification-violating plans before rollout~\citep{obi2025safeplanleveragingformal}. Instead of filtering completed outputs, more tightly integrated strategies leverage formal specifications to restrict the generative decoding space directly, using LTL-derived constraints to prune inadmissible plan continuations during autoregressive generation~\citep{wu2025selpgeneratingsafe}. When violations do occur, verification failures can be transformed from binary rejections into corrective signals; by translating model-checking diagnostics, formal-methods feedback, or counterexample-style information back into the planning prompt, these frameworks support iterative plan repair, automated prompt refinement, and targeted correction channels~\citep{lee2025veriplanintegratingformal,yang2025advfllm,meng2025inprovfleveraginglarge}.

To mitigate the high computational overhead of repeatedly invoking full formal verification stacks, a separate optimization paradigm distills verification outputs into learned surrogate verifiers~\citep{yang2025repvsafetyseparable}. 
Rather than replacing formal verification with an equivalent guarantee, this approach transfers part of the formal-checking signal into a lightweight neural screening module, improving scalability while making the assurance dependent on training coverage, approximation quality, and failure-rate evaluation.
This verification cluster is further complemented by a broader category of lower-assurance critique loops and empirical executability checks. These heuristic strategies either employ a secondary language model as a qualitative \enquote{safety judge} to critique candidate plans~\citep{khan2025safetyawaretask}, or rely on simulation environments, human-in-the-loop interventions, and syntax error logs to iteratively repair malformed behavior trees and invalid programs~\citep{izzo2024btgenbotbehaviortree, llmasbt, skreta2023errorsareuseful}. Accordingly, this collective hierarchy of verification supports planning-time screening, constrained generation, and verifier-guided refinement, supporting checks for mathematical or syntactic consistency before physical rollout.

Accordingly, the evidence boundary of this verification cluster is specification-relative: it can increase confidence that a plan is consistent with encoded rules and constraints, but only with respect to what the verifier is able to observe and formalize. 
The central open gap is that specification-relative correctness is not the same as embodied safety. 
A plan can pass formal or learned verification while relying on incomplete predicates, stale perception, missing contact assumptions, or infeasible geometric transitions. 
At the same time, exhaustive formal checking is often computationally costly, motivating learned or heuristic verifiers whose scalability comes with weaker assurance.

\definecolor{jinsik}{RGB}{0,112,121}

\subsection{Spatial and Model-based Planning Support} \label{sec3:spatial_model_support}
%%%%%%%%%%%%%%%%%%%%%%%%%%%%%%%%%%%%%%%%%%%%%%%%%%%%%

While Section~\ref{sec3:grounding} defines the semantic intent of a task and Section~\ref{sec3:plan_structuring} organizes it into a logically verified sequence, the remaining planning-time challenge is fundamentally physical: transforming this abstract sequence into a scene-grounded, motion-feasible specification before execution begins. Section~\ref{sec3:spatial_model_support} addresses this transition through three complementary mechanisms. First, world models and foresight methods introduce future-state prediction into the planning loop. Second, spatial and object-centric representations specify exactly \textit{where} in the scene the plan applies. Finally, integrated Task and Motion Planning (TAMP) combines these commitments with explicit grasp, kinematics, and collision reasoning.

% If Section~\ref{sec3:grounding} establishes what the task means and Section~\ref{sec3:plan_structuring} organizes that task into a validated plan structure, the remaining planning-time question is more physical: how that validated plan becomes a scene-grounded and motion-feasible planning object before execution. Section~\ref{sec3:spatial_model_support} addresses this through three complementary mechanisms: world models and foresight methods bring future-state prediction into planning, spatial and object-centric representations specify where in the scene the plan applies, and integrated task-and-motion planning combines those commitments with grasp, inverse-kinematics, motion, and collision reasoning.

%%%%%%%%%%%%%%%%%%%%%%%%%%%%%%%%%%%%%%%%%%%%%%%%%%%%%
\subsubsection{World model and Foresight-guided Planning}
%%%%%%%%%%%%%%%%%%%%%%%%%%%%%%%%%%%%%%%%%%%%%%%%%%%%%

Long-horizon manipulation failures often begin as small state drifts or subtask-transition mismatches, rather than as immediately visible collisions or infeasible motions. A robust pre-execution assessment therefore cannot merely evaluate whether the immediate next step is plausible; it must also estimate the intermediate states that step is likely to produce. World models and foresight methods address this requirement by embedding future-state anticipation directly into the planning-time risk analysis.

Predictive rollouts offer a means to evaluate state continuity by tracking the logical and visual consequences of actions across stages \citep{huang2026wmrobotictask}. Action sequences can be iteratively revised prior to execution by utilizing imagined future states as corrective signals to improve plan stability \citep{feng2025reflectiveplanningvision}. To further mitigate error accumulation over extended horizons, multi-view world models can be combined with stage-aware structures to improve transition consistency \citep{chen2025robohorizonllmassisted}.
% Collectively, these approaches treat long-horizon planning as a state-evolution problem where compounding errors and transition consistency are primary concerns. Future-aware plans can also be generated by converting high-level instructions into stable trajectories guided by stage-conditioned foresight imagery \citep{zhang2026foreactsteeringyour}. 
Collectively, these approaches treat long-horizon planning as a state-evolution problem where compounding errors and transition consistency are primary concerns. Future-aware plans can also be guided by stage-conditioned foresight imagery, allowing policies to condition on imagined future observations and subtask descriptions rather than only high-level language instructions~\citep{zhang2026foreactsteeringyour}.

However, the evidence provided by these methods remains primarily empirical and prediction-relative. Foresight images and learned world model transitions can improve long-horizon consistency, but they do not guarantee that predicted futures are accurate, geometrically feasible, or physically safe. Predicted future states become increasingly vulnerable to hallucination, distribution shift, and error accumulation as the horizon grows. Furthermore, visual foresight often fails to capture underlying physical feasibility under occlusion, unexpected object dynamics, or contact-rich interaction. Finally, predicting \textit{what} the future will look like does not by itself geometrically specify \textit{where} the robot must move to achieve it. The next planning-time requirement is therefore converting a plan into a spatially grounded one.

%%%%%%%%%%%%%%%%%%%%%%%%%%%%%%%%%%%%%%%%%%%%%%%%%%%%%
\subsubsection{Spatial and Object-centric Constraint Construction}
%%%%%%%%%%%%%%%%%%%%%%%%%%%%%%%%%%%%%%%%%%%%%%%%%%%%%

While a logically valid symbolic plan dictates the sequence of actions, it does not determine the physical manipulation sites or spatial relationships required to execute them. In long-horizon tasks, identical subtask labels can yield drastically different reachability, contact stability, and placement outcomes depending on which part of an object is grasped and which relative pose is targeted. A complete planning specification must therefore dictate not just \textit{what} to do, but exactly \textit{where} to act.

Transitioning from symbolic structures to physically actionable plans first requires scene-aware problem descriptions, but these symbolic objects must then be enriched with explicit geometric constraints to specify where the robot should act~\citep{shirai2024visionlanguageinterpreter}.
Recent methodologies achieve this by mapping task intent into 3D regions, keypoints, and object parts. For instance, the formulation of manipulation tasks as relational keypoint constraints allows for the precise specification of interaction targets and spatial boundaries \citep{huang2024rekepspatiotemporal}. This can be further refined through part-level spatial grounding, where sub-object geometric structures are treated as active planning variables \citep{huang2024copageneralrobotic}. Additionally, fine-grained visual semantics can be converted into refinable 3D spatial constraints to bridge the gap between abstract descriptions and manipulable geometry \citep{su2025resem3drefinable3d}. Collectively, these techniques translate high-level instructions into explicit and optimizable geometric restrictions necessary for reasoning over actual scene context.

The practical necessity of such spatial reasoning is evident in systems that connect language-generated plans to continuous constraint satisfaction, rejecting or revising plans that lead to collisions, kinematic infeasibilities, or unstable configurations~\citep{curtis2024trustproc3ssolving}. 
Furthermore, recent benchmark evidence indicates that omitting explicit spatial grounding substantially limits the executability of long-horizon plans~\citep{jung2026spatiallygroundedlong}. 
The evidence boundary of this cluster is therefore empirical spatial grounding; these methods do not guarantee that a collision-free, kinematically feasible, dynamically valid, or contact-stable trajectory exists. 
The remaining open gap is the handoff from spatially grounded task constraints to verified motion feasibility, since a plan may correctly identify the relevant object part or target pose while still failing under robot-specific kinematics, cluttered geometry, or contact dynamics. 
This motivates the next planning-time layer, where spatial commitments must be coupled with TAMP, trajectory optimization, or downstream motion verification before they can support stronger claims about safe execution.

% While these scene-grounded spatial commitments are vital for empirical success, establishing a spatial target is distinct from generating the physical trajectory required to reach it. Achieving full motion feasibility and formal safety guarantees ultimately requires handing these spatially grounded plans off to integrated TAMP frameworks.

% The necessity of this spatial reasoning is made explicit in recent applied systems. \textit{Trust the PRoC3S}~\citep{curtis2024trustproc3ssolving} connects language intent directly to object-aware geometry, preemptively filtering out collisions, kinematic infeasibilities, and unstable grasps during the planning phase. Similarly, \cite{jung2026spatiallygroundedlong} demonstrates  that omitting precise spatial coordinates rapidly degrades long-horizon executability.

% The evidence in this cluster confirms that scene-grounded spatial commitments are empirically vital for long-horizon planning. However, defining a spatial target is not the same as generating the physical trajectory to reach it. To achieve full downstream motion feasibility and formal guarantees, these spatially grounded plans must finally be handed off to integrated TAMP frameworks.

%%%%%%%%%%%%%%%%%%%%%%%%%%%%%%%%%%%%%%%%%%%%%%%%%%%%%
\subsubsection{Integrated Task-and-Motion Planning Support}
%%%%%%%%%%%%%%%%%%%%%%%%%%%%%%%%%%%%%%%%%%%%%%%%%%%%%

To reach physical execution, the system must also resolve the specific grasps, poses, and collision-avoidance trajectories required to realize them. The central operation is synthesizing grounded task intent into a unified representation that binds symbolic structure with continuous geometric constraints.

Early efforts to bridge this gap approximated this transition through state-conditioned feasibility scoring or by translating linguistic constraints into 3D value maps for model-based planning~\citep{saycan,huang2023voxposercomposable3d}. However, these approaches often stop short of a fully integrated TAMP framework capable of joint symbolic and continuous reasoning. A more formal baseline for such integration involves connecting continuous variables---such as inverse kinematics, object poses, and collision-free trajectories---to a symbolic planner through black-box samplers, making physical feasibility conditions available during the symbolic search process~\citep{garrett2020pddlstreamintegratingsymbolic}.

Building on these foundations, modern systems directly merge discrete logic with continuous kinematics to ground language into physical reality. This can be achieved by transforming instructions and observations into domain-specific specifications that support geometric constraint reasoning, resulting in executable TAMP plans under modeled assumptions~\citep{siburian2025grounded}. Other frameworks convert open-world language grounding into discrete and continuous language-parameterized constraints that can be integrated into standardized TAMP solvers~\citep{kumar2026openworldtask}. Together, these methodologies demonstrate how high-level intent can be lowered into representations that expose scene geometry and feasibility constraints to the planner.

Even within rigorous TAMP formulations, however, planning-time refinement remains necessary to address unsolvable states. Feeding motion planning failures---such as collisions or unreachable grasp solutions---back into the high-level reasoning process allows iterative realignment of symbolic choices and continuous action parameters with physical constraints~\citep{wang2024llmsupsup}. This feedback loop serves as a final pre-rollout refinement mechanism, increasing the likelihood that the synthesized plan is physically feasible before the robot initiates movement.

While classical frameworks provide formal structure under modeled assumptions, recent language-grounded approaches offer the empirical machinery needed to make high-level reasoning outputs concretely executable. Collectively, these advancements can help translate abstract instructions into more validated, scene-grounded, and motion-feasibility-aware plans under modeled assumptions. However, this planning-time rigor serves as a prerequisite for safety rather than a final guarantee; once the robot begins to move, policy-level action generation and runtime disturbances become the primary concerns, as examined in Sections \ref{sec:section4} and \ref{sec:section5}.

\section{Policy-time Safety}
\label{sec:section4}

% Policy-time safety concerns how a manipulation policy is shaped before its next action is committed to the physical environment (\textcolor{red}{e.g. during training of RL / IL policy, trajectory optimization}).
Policy-time safety concerns how a manipulation policy is shaped before its next action is committed to the physical environment, for example, during RL or IL training. If Section~\ref{sec:section3} evaluated safety-relevant risks in the abstract task formulation, this section focuses on the policy layer itself: how candidate actions, subtask transitions, and learned preferences are formed at decision time. Under the standard two-level architecture of high-level planning and low-level control, this encompasses the action generation stage of the policy, alongside the wrappers that modify what that policy is allowed to propose~\citep{liu2025aligningcyberspace}.

\begin{figure}[t]
    \centering
    \includegraphics[width=0.99\linewidth]{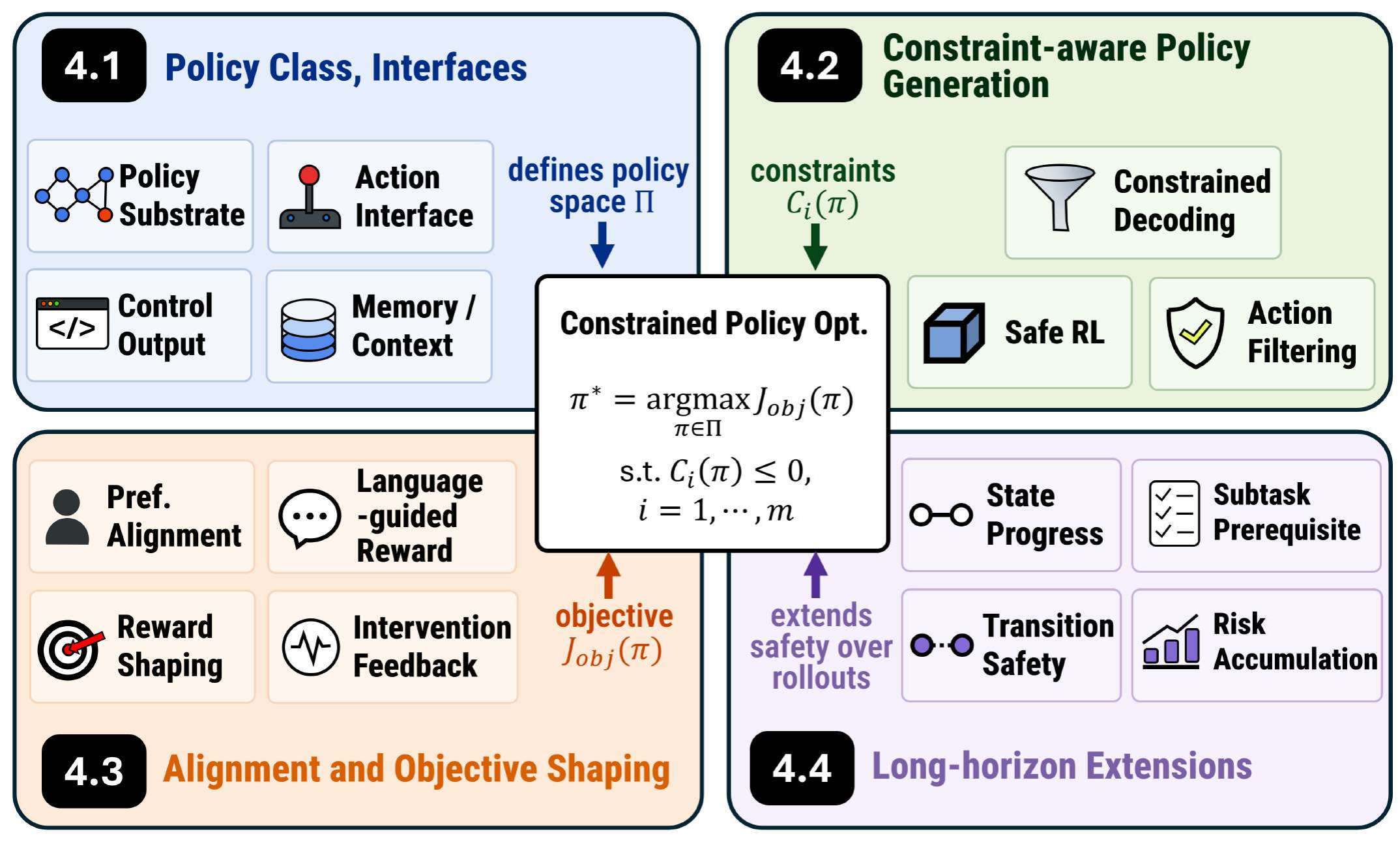}
    \caption{
    Graphical overview of policy-time safety. Policy substrates determine the policy class $\Pi$ and the interfaces available before environmental commitment (Section~\ref{sec4:substrates}). Safety is then integrated either through explicit constraints that restrict admissible behavior during policy generation (Section~\ref{sec4:constraint_aware}) or through objective shaping (Section~\ref{sec4:alignment}). Furthermore, long-horizon tasks introduce additional procedural risks arising from stage transitions and delayed failure accumulation (Section~\ref{sec4:procedural_safety}).
    }
    \label{fig:section4}
\end{figure}

To organize the methods operating at this layer, we frame policy generation as a generic constrained optimization problem:
\begin{align}
\pi^\star &= \arg\max_{\pi \in \Pi} J_{\mathrm{obj}}(\pi) \\
\text{s.t.}&\quad C_i(\pi) \le 0, \qquad i = 1, \dots, m .
\end{align}
While not a literal formulation for every method reviewed, this mathematical view serves as a unifying abstraction for the remainder of this section (summarized in Figure~\ref{fig:section4}). 
% Section~\ref{sec4:substrates} defines the available policy class ($\Pi$), examining the backbone substrates and action spaces that determine what the policy outputs. Section~\ref{sec4:constraint_aware} then focuses on explicit constraints ($C_i(\pi)$), exploring how decoding, learning, and sequence generation are restricted to block unsafe proposals\textcolor{red}{, more emphasis on safety-critical components}. Finally, Section~\ref{sec4:alignment} examines objective shaping ($J_{\mathrm{obj}}$), where safety is embedded directly into what the policy is trained to prefer\textcolor{red}{, especially focusing on human preference or guidance}.
Section~\ref{sec4:substrates} defines the available policy class ($\Pi$), examining the backbone substrates, action spaces, executable interfaces, and context structures that determine what the policy outputs. Section~\ref{sec4:constraint_aware} then focuses on explicit constraints ($C_i(\pi)$), exploring how decoding, learning, and sequence generation are restricted to block unsafe proposals. Section~\ref{sec4:alignment} next examines objective shaping ($J_{\mathrm{obj}}$), where safety is embedded directly into what the policy is trained to prefer. Section~\ref{sec4:procedural_safety} then examines how long-horizon manipulation changes the policy-time failure structure, with a focus on procedural progression, stage transitions, and delayed risk accumulation.

% Throughout this progression, the defining distinction from Section~\ref{sec:section5} is the \textit{locus of intervention}: this section concerns shaping unsafe proposals \textit{before} commitment, whereas Section~\ref{sec:section5} concerns responding to unsafe behavior \textit{during} execution.

Note that policy-time safety mainly concerns procedural and semantic risks before action commitment, but at a lower level compared to the abstract task-level considerations in planning-time safety mechanisms. 
Alignment, robustness, and reward shaping are treated as safety evidence when they are explicitly tied to hazard reduction, constraint satisfaction, unsafe proposal suppression, or the mitigation of failure propagation.

\subsection{Policy Class, Interfaces, and Long-horizon Context}\label{sec4:substrates}
% ------------------------------------------------------------

At the policy-substrate level, defining the policy class $\Pi$ requires looking beyond task performance to consider the pre-commitment structure the backbone provides for safety shaping. A policy may expose different action representations, such as continuous controls or discrete action tokens, or higher-level executable interfaces like skill calls or program lines. These architectural choices are critical because they determine which candidate actions can be represented, which action formats can be constrained, and which forms of policy-time shaping are available before commitment.

To systematically exploit these architectural boundaries, policy-time safety interventions should be understood as interface-dependent rather than uniformly applicable across all policy substrates. 
In tokenized VLA regimes, where robot actions are represented through discrete outputs, the action vocabulary itself becomes a potential intervention surface: unsafe or inadmissible action tokens can in principle be blocked, reweighted, or constrained before commitment~\citep{brohan2023rtvisionlanguage,kim2024openvla}. 
By contrast, continuous or chunked action representations shift the intervention surface away from tokens toward trajectory-level distributions and action sequences, making projection, resampling, or continuous safety filtering more natural forms of pre-commitment shaping~\citep{chi2023diffusionpolicyvisuomotor,team2024octo}. 
Beyond low-level motor streams, programmatic and skill-based interfaces expose higher-level intervention points: generated code lines and API invocations can be inspected before execution, while state-conditioned skill spaces restrict or adapt the set of executable primitives available to the policy~\citep{liang2023codeaspolicies,rana2023residual}. 
% Multimodal reasoning substrates further insert language or visual reasoning as intermediate context for downstream robot control, allowing high-level interpretations to modulate lower-level action generation rather than directly outputting motor commands~\citep{driess2023palmembodiedmultimodal}.

The safety relevance of these diverse policy classes also depends on how they handle longer-horizon context, since stale observations, progress confusion, and subtask drift can accumulate during extended rollouts. 
Hierarchical language-conditioned policies address this issue by decomposing long-horizon control into higher-level latent plans and a lower-level visuomotor policy~\citep{mees2022matters_hulc}. 
More explicitly memory-augmented substrates retain information from past observations or actions, supporting temporally dependent manipulation decisions~\citep{fang2025sam2act,shi2025memoryvla}. 
A related line of long-horizon VLA models exposes intermediate progression variables through interleaved language planning, visual forecasting, or visual chain-of-thought reasoning~\citep{hu2026bagelvla,zhao2025cotvlavisual}. 
Finally, latent-action alignment changes the intermediate action representation itself, providing action-centric latent spaces for scalable VLA pretraining rather than directly enforcing safety constraints~\citep{luo2026joint_aligned_latent_action}.

Together, these structural variations demonstrate that policy-time safety cannot be applied uniformly; rather, it must be tailored to the specific code, skill, token, or continuous control interfaces exposed by the backbone. 

\subsection{Constraint-aware Policy Generation} \label{sec4:constraint_aware}
% ------------------------------------------------------------

Constraint-aware policy generation focuses on shaping the policy so that unsafe behavior is restricted, filtered, or made less likely within the candidate policy space. While Section~\ref{sec4:substrates} defined the available action representations and interfaces, this section examines how those are actively restricted. It proceeds in two steps: first by reviewing direct constraint injection at policy-generation stage, and then by examining constrained learning and safe optimization that internalize these safety structures during training. Together, these mechanisms offer varying strengths of assurance, ranging from specification-relative restriction under explicit formal models to empirical reductions in unsafe candidate actions.

% ------------------------------------------------------------
\subsubsection{Constraint Injection During Policy Generation}\label{sec4:constraint_injection}
% ------------------------------------------------------------

Constraint injection is the most direct method to enforce specified constraints or block specified unsafe proposals before execution. Rather than depending on the training distribution alone to avoid unsafe behavior, these methods actively restrict the action space. 
Depending on the policy interface, this may take the form of masking or down-weighting tokenized continuations under explicit temporal constraints~\citep{kapoor2025constraineddecodingrobotics}, checking generated program lines against available APIs or assertion-like symbolic state checks~\citep{singh2023progpromptgeneratingsituated}, or modifying continuous actions through safety layers, projections, or differentiable optimization modules~\citep{dalal2018safe,amos2018differentiable}.
% Depending on the policy interface, this may take the form of masking or down-weighting tokenized continuations~\citep{kapoor2025constraineddecodingrobotics}, rejecting candidate program lines against explicit APIs or assertions~\citep{singh2023progpromptgeneratingsituated}, or projecting continuous actions back into a feasible set~\citep{dalal2018safe, amos2018differentiable}. 
However, its safety coverage remains strictly bounded by the accuracy of the modeled rules, the system interface, and the quality of the specification. 

% The safety claim here is narrow but robust: when constraints are explicit, injection prevents unsafe proposals by construction. 
% \textcolor{red}{As Section~\ref{sec4:substrates} suggested, the concrete mechanism depends on what the policy emits: tokenized policies admit masking and reweighting, programmatic or skill-level interfaces admit line- or call-level rejection, and continuous controllers admit projection or filtering.}

% \paragraph{Action-space restriction before commitment}
% The most direct form of policy-time safety acts on the action proposal itself. Rather than hoping that training alone will make unsafe behavior unlikely, these methods restrict what the policy may emit at decision time by masking, pruning, reweighting, or projecting candidate actions before they are committed to the environment. This is still distinct from execution-time intervention in Section~\ref{sec:section5}: the key operation here is not to halt or recover after rollout has begun, but to ensure that the current action proposal already respects an explicit safety rule or feasible-set approximation. The safety claim is correspondingly narrow but important. When the constraint representation is explicit, policy-time injection can prevent some unsafe proposals by construction, even though coverage remains limited to the modeled rules, system interface, and specification quality.

\paragraph{Formal Constraint Injection}

The most direct way to shape policy-time safety is to attach explicit formal constraints to the action proposal before commitment. 
This idea builds on a broader temporal logic control lineage, where LTL or STL specifications define admissible robot behavior under modeled dynamics~\citep{verginis2024planning,sewlia2022cooperative}. 
In learning-based settings, shielding provides the closest classical analogue: a dedicated safety layer monitors the learner's proposed action and either restricts the available choices or corrects unsafe actions with respect to an explicit temporal specification, without retraining the underlying policy~\citep{alshiekh2018shielding}. 
Recent foundation-model-oriented work moves this intervention into autoregressive action generation itself, using STL constraints to mask or down-weight inadmissible candidate continuations during decoding~\citep{kapoor2025constraineddecodingrobotics}. 
These methods therefore support a narrow but strong policy-time claim: when the specification and system model are explicit, unsafe action proposals can be blocked or modified before execution, although the guarantee remains relative to the encoded constraint and interface.

\paragraph{Specification Construction and Interface Grounding}

Constraint injection is not solely about the enforcement mechanism itself; it also depends on constructing explicit specification objects that can later shape generation, skill selection, or planning. Natural-language-to-LTL pipelines provide one route by translating informal user instructions into formal temporal specifications, with recent methods improving either translation reliability through conformal uncertainty control or data efficiency through synthetic training and constrained decoding~\citep{conformalnl2ltl2025,pan2023dataefficient}. At a higher semantic abstraction, robot constitutions provide normative rule sets for identifying semantically unsafe or counter-normative robot behavior, although these rules should be understood as semantic guardrails rather than formal specifications~\citep{liu2025constitutions}.  Once such specifications are available, they can be grounded in executable interfaces, for example by composing or transferring learned skills to satisfy LTL task specifications~\citep{liu2022ltltransfer}. Taken together, these methods show that policy-time restriction depends on the fidelity, expressivity, and enforceability of the specification object.

\subsubsection{Constrained Learning and Safe Optimization}\label{sec4:constrained_learning}
% ------------------------------------------------------------
% While constraint injection filters unsafe actions at decision time, constrained learning seeks to embed safety directly into the policy optimization process. This paradigm fundamentally alters how the policy is trained, ensuring that structural safety is internalized within the learned parameters from the ground up. Constrained learning embeds safety directly into the policy optimization process. In the manipulation literature, this is typically achieved through two main approaches: incorporating explicit safety constraints into the reward model, or leveraging formal safety certificatesâ€”such as Safety Filtersâ€”to actively guide the policy's learning and exploration.

While constraint injection filters unsafe actions at decision time, constrained learning seeks to embed safety directly into the policy optimization process. This paradigm fundamentally alters how the policy is trained, aiming to internalize safety-related constraints within the learned parameters from the ground up. Constrained learning is typically achieved through two main approaches: incorporating explicit safety constraints into the reward model, or leveraging formal safety certificates to actively guide the policy's learning and exploration.

% Constrained learning can embed safety directly into the policy optimization process. Some papers didthis by using safety constraints and put constraints into reward model, or safety certificate can be used as guidance for the learning of policy with safety.

% Constrained learning moves the same safety logic into training: the policy is optimized so that unsafe behavior becomes less likely to be proposed even before any decoding-time filter is applied.

\paragraph{Safe Reinforcement Learning}

The theoretical foundation for integrating safety into policy optimization comes from safe RL, broadly defined as learning policies that maximize return while respecting safety constraints during learning or deployment~\citep{garcia2015saferl}. In the constrained-RL lineage, Constrained Markov Decision Process (CMDP) formulations optimize expected reward subject to expected safety-cost limits, and foundational algorithms operationalize this view through trust-region updates or adaptive Lagrangian multipliers~\citep{achiam2017cpo,stooke2020responsivesafetyreinforcement}. These methods treat safety as part of the policy optimization objective rather than as a purely post-hoc veto, but their guarantees are typically expectation-based and may still permit transient or state-wise violations during learning. To address stricter feasibility requirements, hard-constrained RL methods restrict the policy output or action parameterization for particular classes of constraints, such as equality constraints handled through reduced policy optimization or affine state constraints enforced by POLICEd RL~\citep{ding2023reduced,bouvier2024policed}. These methods provide stronger guarantees under their respective structural assumptions, but they do not directly solve the broader problem of long-horizon, language-guided manipulation. Recent VLA safety-alignment work moves constrained learning into foundation-model settings by eliciting unsafe scenarios and fine-tuning VLA policies to reduce high-risk behaviors~\citep{zhang2025safevla}. Because such alignment still depends on statistical optimization and scenario coverage, its evidence remains empirical and distribution-dependent. Establishing deterministic or formal safety guarantees for long-horizon VLA manipulation therefore remains open.

\paragraph{Certificate-guided Safe Learning}
% Control-theoretic safe learning offers a more formal extension of constrained optimization by replacing safety rewards with explicit certificates or safe sets that shape the learning process itself. This includes Hamilton--Jacobi reachability-based safe sets~\citep{shao2021reachability,wabersich2023data} and barrier certificates~\citep{taylor2020cbf,dawson2022certificates,cbfsurvey2024}, which act to supervise policy improvement in continuous control and explicitly safeguard the trained policy. Similarly, constraint-manifold methods~\citep{liu2022atacom} restrict exploration exclusively to actions compatible with an explicit safety manifold.
Control-theoretic safe learning offers a more formal extension of constrained optimization by replacing generic safety rewards with explicit safe sets, certificates, or filters that shape learning and exploration. This includes Hamilton--Jacobi reachability-based safe sets and predictive safety filters~\citep{shao2021reachability,wabersich2023data}, as well as CBFs and learned certificate methods that supervise or verify safe policy improvement in continuous control~\citep{taylor2020cbf,dawson2022certificates,cbfsurvey2024}. Constraint-manifold methods provide another route by transforming the action space so that exploration occurs along directions compatible with an explicit safety manifold~\citep{liu2022robot}.

However, applying these methods to high-dimensional manipulation faces critical synthesis bottlenecks. Analytically deriving certificates for complex kinematics and contact dynamics is difficult, and exact reachability methods suffer from the curse of dimensionality~\citep{bansal2017hamilton}. Neural certificate methods attempt to learn scalable Lyapunov or barrier functions from data~\citep{ma2022joint,zhang2023exact}, but certifying the required inequalities over continuous state spaces remains challenging and often requires additional verification machinery. More broadly, learned or latent safety filters remain vulnerable to distribution shift and out-of-distribution risks when their training coverage is incomplete~\citep{seo2025uncertaintyawarelatent}. Consequently, lower training-time violation rates or locally verified certificates do not by themselves guarantee robust deployment-time safety across the full procedural risk of long-horizon manipulation.

% However, applying these methods to high-dimensional manipulation faces critical synthesis bottlenecks. Analytically deriving certificates for complex kinematics is mathematically intractable, and exact reachability methods suffer severely from the curse of dimensionality~\citep{bansal2017hamilton}. While recent approaches attempt to learn neural certificates from data~\citep{ma2022joint,zhang2023exact}, enforcing strict differential inequalities across continuous state spaces is difficult, frequently leading to violations in out-of-distribution regions~\citep{seo2025uncertaintyawarelatent}. Furthermore, in long-horizon tasks, even successfully approximated local certificates rarely cover the full procedural hazard surface. Consequently, reduced training-time violation rates do not inherently guarantee robust deployment-time safety under distribution shift.

\subsection{Alignment and Objective Shaping} \label{sec4:alignment}

Explicit constraints can block some unsafe candidate actions before commitment, but they do not by themselves determine which behaviors the objective function rewards. This matters for safety because policy-time risk is not limited to immediate physical collision: it also includes semantic misalignment, unsafe human-facing behavior, and objective misspecification that makes harmful or catastrophically misaligned behavior more likely over time. This subsection therefore turns from explicit action-space restriction to objective shaping, asking how reward, preference, and feedback design can make the policy less likely to favor unsafe, semantically misaligned, or human-dispreferred behavior. In long-horizon manipulation, this is especially important because many failures originate in objective misspecification rather than in a single obviously invalid action.

% ------------------------------------------------------------
\subsubsection{Preference and Reward Model Alignment}\label{sec4:preference_alignment}
% ------------------------------------------------------------
Many long-horizon safety failures are linked to specification mismatches: the policy may optimize for proxy metrics that are easy to score, while underweighting important procedural behaviors such as preserving safe state progression and satisfying prerequisites. Preference and reward model alignment aim to address this by calibrating the training objective directly against human judgments of safety, efficiency, and success, thereby mitigating the limitations of handcrafted reward functions.

Aligning robot policies with trajectory-level human preferences, for instance, has demonstrated significant empirical reductions in collision rates by balancing task efficiency with safety constraints \citep{zhang2024grape}.
Furthermore, the application of preference-aligned diffusion models to complex domains, such as deformable object manipulation, has been shown to improve both model personalization and adherence to human guidance \citep{moletta2026preferencealignedvisuomotor}.

% To manage the substantial cost of collecting extensive human preference data, recent frameworks~\citep{liu2023pearlzeroshot,tian2024maximizing, mattson2024representationalignment} demonstrate that preference-shaped rewards can remain effective even with sparse feedback, across diverse task embodiments, and when learning from mixed-quality datasets.
To manage the substantial cost of collecting extensive human preference data, recent frameworks~\citep{liu2023pearlzeroshot,tian2024maximizing,mattson2024representationalignment} demonstrate that preference-shaped rewards can remain effective with fewer labels, transfer across tasks, or support learning from mixed-quality datasets. 
While this line of research traces back to classical preference-based reinforcement learning~\citep{christiano2017deep}, the safety claim for modern robot foundation policies remains limited: preference alignment can bias the learned policy away from dispreferred or unsafe behaviors, but it does not exclude unsafe candidates by construction or provide formal assurance.
% the safety claim for modern foundation models remains limited: they successfully steer behavior away from unsafe policy before commitment, but they do not provide formal assurance.

% ------------------------------------------------------------
\subsubsection{Language-guided and Intervention-derived Reward Shaping}\label{sec4:reward_shaping}
% ------------------------------------------------------------

Beyond preferences, another line of alignment research expands the feedback channel used to shape the policy objective. 
The key motivation is that long-horizon failures are often difficult to diagnose from sparse success labels alone: a task may fail because the policy chooses the wrong subgoal or violates an implicit precondition.
Richer feedback modalities therefore aim to make the reward signal more diagnostic of \textit{why} a behavior should be encouraged, penalized, or corrected. 
This direction is consistent with broader alignment research, where language feedback, separate reward--cost objectives, and multimodal safety feedback have been explored as alternatives to scalar preference labels~\citep{zhou2025sequencesequencereward,dai2023saferlhf,ji2025saferlhfsafe}.

In embodied manipulation, language-conditioned reward models use descriptive data or demonstrations to support sample-efficient adaptation to unseen task variations, reducing reliance on per-task reward engineering or additional demonstrations~\citep{zhang2025rewind}. 
Language can also serve as an editable reward-design interface, where LLMs generate dense programmatic reward routines and refine them through human textual feedback~\citep{xie2023text2reward}. 
Visual feedback extends this idea by grounding reward evaluation in observable behavior: video-language critics learn reward functions from cross-embodiment video data, while failure prompts help distinguish successful and failed executions across new environments and instructions~\citep{alakuijala2024videolanguagecritic,yang2024adapt2reward}. 
Finally, physical intervention traces provide a direct corrective signal, since human takeovers identify regions of the behavior distribution that require assistance and can be converted into residual rewards~\citep{chen2024mereqmaxent}.

Together, these methods broaden policy alignment from \textit{learning which trajectory is preferred} to \textit{learning why a behavior should be rewarded or corrected}. 
Their evidence boundary is primarily empirical and objective-relative: richer supervision can improve the diagnostic quality of the learned reward, but it does not certify that the resulting objective captures all safety-critical behavior. 
Because these feedback channels mainly shape training-time or pre-deployment objectives rather than serving as certified runtime monitors, their safety claims remain bounded by feedback coverage, reward model generalization, and distributional match between training and deployment.
% Together, these methods broaden the scope of policy-time alignment from \textit{learning from preferences} to \textit{learning from richer evaluative feedback}. However, within these frameworks, language, vision, and physical interventions serve strictly as offline optimization objectives to structure the static reward function prior to deployment, rather than acting as dynamic, closed-loop monitors during live execution. Consequently, the safety claims supported by these methodologies remain inherently empirical: while richer supervision significantly enhances nominal behavior shaping before deployment, it provides no formal or deterministic guarantees once the policy encounters out-of-distribution hazards during rollout.

\subsection{Long-horizon Extensions of Policy-time Safety} \label{sec4:procedural_safety}

Long-horizon manipulation changes the failure structure that policy-time safety must address: safety violations may emerge through premature stage transitions, skipped prerequisites, or latent risk accumulation across extended rollouts.

This subsection therefore examines how explicit restriction and objective shaping must be extended when the central problem is procedural progression rather than only on single-step action admissibility. We do so through two complementary lenses: structural extensions that make the policy space more progress-aware, and objective-side extensions that make the training signal more stage-aware.

% ------------------------------------------------------------
\subsubsection{Progress-aware Structural Extensions}\label{sec4:progress_structure}
% ------------------------------------------------------------

Long-horizon policy-time safety extends beyond blocking immediate, obviously unsafe actions; it requires preserving enough structure for task progress to remain inspectable before actions are committed. Procedural failures often arise from entering a task phase prematurely, omitting prerequisite mitigation steps, or accumulating latent risk while local behavior still appears nominal. A first structural route is to expose higher-level policy interfaces, such as reusable skills or programmatic action routines, rather than relying only on unconstrained low-level controls. Such abstractions make subtask progression more inspectable and provide natural points for checking skill choice, API calls, or program structure~\citep{saycan,liang2023codeaspolicies}. To reduce stale-state errors over extended rollouts, these interfaces can be augmented with context-retention mechanisms that compress past observations or maintain multi-scale memory for long-horizon decision making~\citep{li2024contextvla,lu2025mem}. Beyond architectural memory, inference-time steering methods add active validation before commitment, either by using foresight models to evaluate candidate future outcomes or by checking alignment between the policy's stated reasoning and its proposed action~\citep{wu2025fromforesightforethought,dowhatyousay}. 
% These developments also motivate process-oriented evaluation: long-horizon safety should not be measured only by final task success, but also by whether intermediate risks are perceived and mitigation steps occur in the correct procedural order~\citep{lu2026isbenchevaluating}.
The evidence boundary of these structural extensions is therefore architectural and inference-time rather than formal. Skill abstractions, programmatic interfaces, memory tokens, and reasoning-action checks can make progress more inspectable and reduce certain context-loss or misordered-action failures, but they do not prove that the task is procedurally safe over the full rollout.
% However, even a well-organized hierarchy remains susceptible to failure if the underlying training objective optimization inadvertently rewards local progress at the expense of systemic, long-horizon safety. Once these progress-aware mechanisms fail and the system must detect active drift or repair a fractured trajectory online, the problem transitions from policy-time proposal shaping to closed-loop execution-time intervention, which forms the focus of Section~\ref{sec:section5}.

% ------------------------------------------------------------
\subsubsection{Stage-aware Objective Shaping for Long-horizon Manipulation}\label{sec4:stage_aware_shaping}
% ------------------------------------------------------------
Building on the progress-aware architectures from Section~\ref{sec4:progress_structure}, recent methods extend this temporal structure directly into the training objective. In long-horizon manipulation, monolithic scalar rewards often conflate prerequisite completion with final success, allowing policies to silently accumulate procedural errors. 
To mitigate this, stage-aware reward modeling decouples high-level task stages from fine-grained within-stage progress, allowing the optimization process to penalize procedural drift more directly~\citep{chen2025sarmstageaware}.
% To address this, SARM~\citep{chen2025sarmstageaware} introduces stage-aware reward modeling, explicitly decoupling the learning signals for subtask completion, safe handling, and recovery to penalize procedural drift during optimization.

Beyond the use of explicitly defined stages, long-horizon behavioral coherence can be further enhanced through reward-guided latent planning and large-scale post-training \citep{huang2025thinkact, hu2025flareachievingmasterful}.
% Beyond explicitly defined stages, related frameworks like \textit{ThinkAct}~\citep{huang2025thinkact} and FLaRe~\citep{hu2025flareachievingmasterful} demonstrate that reward-guided latent planning and large-scale post-training can further enhance long-horizon behavioral coherence. 
However, these alignment channels remain highly vulnerable to noisy or corrupted human feedback~\citep{vatsa2026rodifrobustdirect}. 
While stage-aware and post-training objectives can reduce some procedural errors and steer the policy toward more coherent long-horizon behavior, their safety claims remain empirical and distribution-dependent; they do not provide calibrated mathematical bounds on policy-level procedural safety.

The remaining open gap is how to ensure that the policy's representation of progress, memory, and stage-wise reward is faithful to the task's true procedural dependencies, rather than merely improving local coherence or final success on the training distribution.
Otherwise, a policy may appear well organized while still skipping hidden prerequisites, accumulating unrecoverable risk, or exploiting stage rewards in ways that remain unsafe at the long-horizon level.

Taken together, Section~\ref{sec:section4} shows how policy-time safety can be strengthened by restricting candidate actions, shaping objectives, and making long-horizon progress more inspectable before commitment. Its evidence boundary remains policy-level: these methods can improve the structure, objective, and screening of action proposals, but they do not establish that the policy has captured all procedural dependencies required for safe long-horizon behavior. Section~\ref{sec:section5} then turns to a different layer of evidence, where already selected actions must be monitored, gated, or repaired during physical execution.

\section{Execution-time Safety} \label{sec:section5}
\begin{figure}[t]
    \centering
    \includegraphics[width=0.99\linewidth]{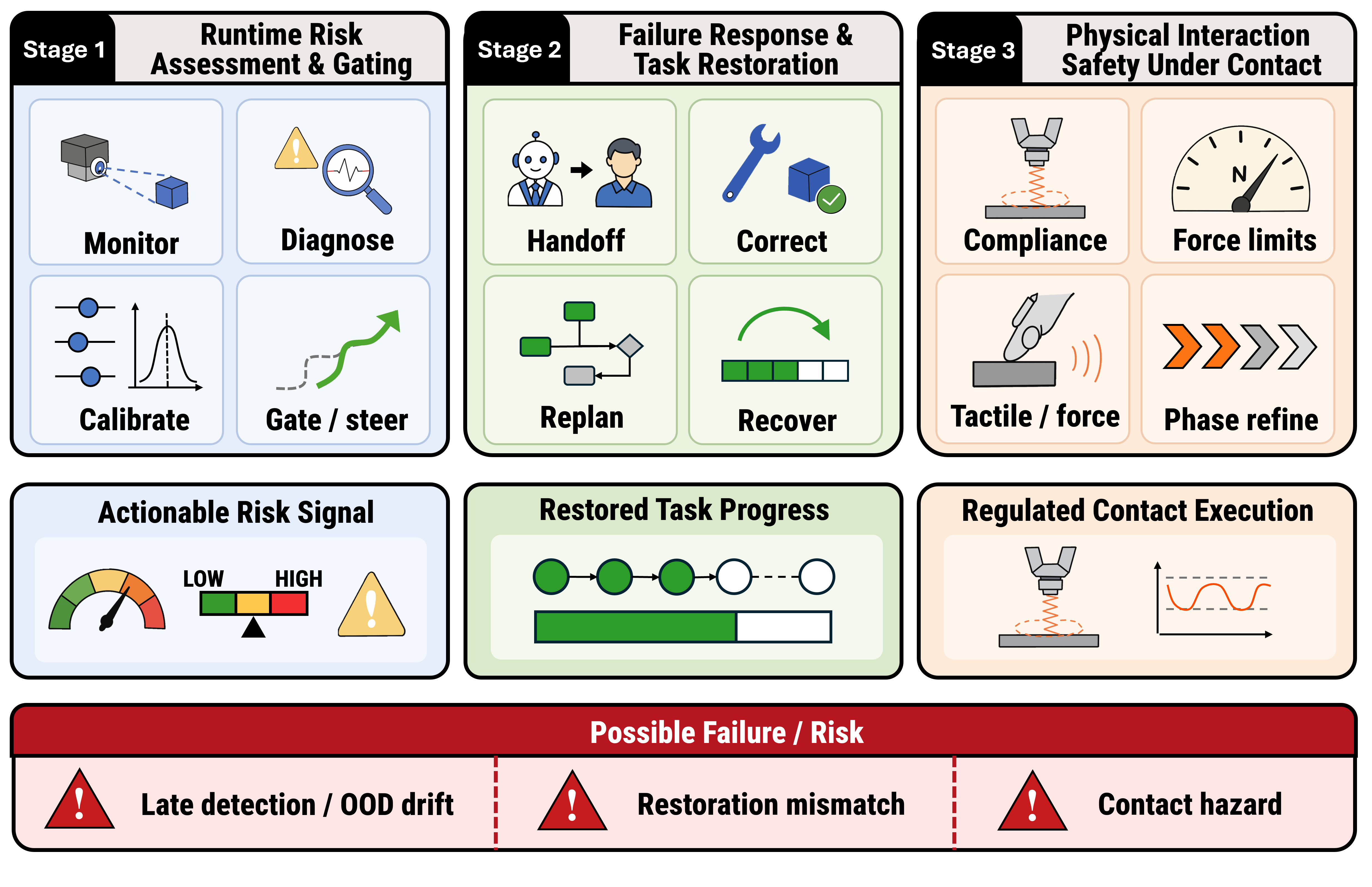}
    \caption{Graphical overview of execution-time safety. Once behavior is already being executed in the world, safety depends on recognizing emerging risk during rollout, deciding whether to gate or redirect behavior, restoring task progress after disruption, and regulating physical interaction once contact occurs. The section is organized around three phases: runtime risk assessment and gating (Section~\ref{sec5:runtime_gate}), failure response and task restoration (Section~\ref{sec5:runtime_restore}), and contact-rich execution safety (Section~\ref{sec5:physical}).}
    \label{fig:section5}
\end{figure}

If Section~\ref{sec:section4} focused on preventing unsafe actions from being proposed, this section addresses what happens once those actions are committed to the physical world. At execution-time, the system is no longer reasoning over idealized plans, static constraints, or shaped objectives; it must cope with observation noise, dynamic scene shifts, contact uncertainty, and failures that occur only during live rollout. Execution-time safety is therefore the layer where abstract safety claims become operational reality.

Execution-time safety intervention concerns how to monitor, diagnose, and repair behavior after rollout has already begun. We organize this section around the three core phases (summarized in Figure~\ref{fig:section5}). Section~\ref{sec5:runtime_gate} examines runtime risk assessment, exploring how systems monitor execution, reason about possible failure, and safeguard against imminent hazards. Once a failure is detected, Section~\ref{sec5:runtime_restore} studies the response, detailing how task progress is restored through human handoff, interactive correction, or autonomous replanning. Finally, because execution is grounded in the physical world, Section~\ref{sec5:physical} addresses the safety demands of physical contact. In this context, semantic correctness must be paired with dynamic, force-aware adaptation to regulate low-level interactions during execution.

Execution-time safety is where operational and physical safety become direct objects of intervention. Semantic or procedural failures may originate upstream, but at execution time they matter because they create unsafe continuation, loss of recoverability, delayed intervention, or physical harm during live interaction.

%%%%%%%%%%%%%%%%%%%%%%%%%%%%%%%%%%%%%%%%%%%%%%%%%%%%%%%%%%%
\subsection{Runtime Risk Assessment} \label{sec5:runtime_gate}
Execution-time safety hinges on a critical question: can the system detect an unsafe rollout early enough to prevent failure escalation? In long-horizon manipulation, risk is rarely a single catastrophic event; rather, it accumulates through distribution shifts, missed preconditions, semantic misalignment, or gradual environmental changes. Such risks often remain locally ambiguous until the system has drifted perilously far from its intended trajectory.

Consequently, runtime risk assessment serves as the primary line of defense. This subsection outlines the progression from \emph{recognizing} risk to \emph{acting} upon it. We first discuss monitoring and anomaly detection to identify execution deviations, followed by diagnosis to explain their root causes. Finally, we examine calibration, shielding, and predictive steering as mechanisms to block or redirect unsafe behavior before a failure fully materializes.

% Execution-time safety begins with a simpler but more urgent question than recovery: can the system recognize that the current rollout is becoming unsafe early enough to prevent failure escalation? In long-horizon manipulation, risk rarely appears as a single catastrophic event. It more often accumulates through distribution shift, missed preconditions, semantic misalignment, or gradually unstable contact, any of which may remain locally ambiguous until the rollout has already drifted far from the intended trajectory. Runtime risk assessment and gating therefore form the first operational defense line of execution-time safety. The subsection follows a progression from \emph{recognizing} risk to \emph{acting} on that recognition: monitoring and anomaly detection identify deviations from nominal execution, diagnosis explains what kind of deviation is occurring, calibration and shielding determine when a risk signal is actionable, and predictive verification or policy steering blocks or redirects unsafe behavior before the underlying failure fully materializes.

%%%%%%%%%%%%%%%%%%%%%%%%%%%%%%%%%%%%%%%%%%%%%%%%%%%%%%%%%%%
\subsubsection{Runtime Monitoring and Anomaly Detection}\label{sec5:monitoring}

The first layer of execution-time safety is to detect that the current rollout is departing from a safe, task-consistent regime. Because failures in long-horizon manipulation emerge gradually, runtime monitoring is better understood as \emph{early deviation detection} rather than post-hoc success classification.

\paragraph{State-level Anomaly Scoring and Multi-modal Monitoring}
A primary strategy for detecting unsafe execution focuses on physical and state-level deviations by modeling nominal behavior. For instance, preemptive anomaly detection can be achieved using robot-conditioned normalizing flows with task-aware state conditioning~\citep{zhou2026rcnfrobot}, while latent dynamics models can be employed to anticipate OOD states before they manifest~\citep{liu2024modelbasedruntime}. 
% Furthermore, observation-space anomaly signals can be extracted by comparing current rollouts against learned nominal distributions. For instance, FAIL-Detect~\citep{xu2025canwedetect}  introduces \texttt{logpZO}, which measures deviations in the latent noise space of a continuous normalizing flow, providing a raw input for safety-critical decisions. To improve robustness beyond single-modality signals, FIPER~\citep{romer2025fiper} simultaneously monitors consecutive OOD observations and high action-intent uncertainty. These frameworks generate continuous risk signals that capture various facets of execution deviation.
For generative and foundation-model-based policies, monitoring often involves extracting anomaly signals from the latent or action space. This includes measuring deviations in continuous noise spaces~\citep{xu2025canwedetect}, monitoring action-intent uncertainty through entropy-based scores~\citep{romer2025fiper}, or leveraging the internal representations of VLA models for multitask failure detection~\citep{gu2025safemultitaskfailure}. To render these raw scores actionable, these frameworks employ conformal prediction~\citep{angelopoulos2023conformal} can provide statistically calibrated thresholds for failure alarms. 
Such calibration helps convert monitoring outputs into more actionable decision signals under calibration assumptions 
% ensures that monitoring outputs are transformed into high-fidelity decision signals, providing the necessary foundation for subsequent diagnosis and shielding.

% Furthermore, observation-space anomaly signals can be introduced alongside threshold calibration to compute online anomaly scores~\citep{xu2025canwedetect, romer2025fiper}. If an anomaly signal—whether computed from action consistency or estimated by a separately trained failure detector—exceeds a calibrated threshold, the system classifies the current rollout as a failure. 
% At the foundation-model level, latent representations of VLA models have been shown to natively support lightweight, multitask failure detection~\citep{gu2025safemultitaskfailure}.

Because relying on a single scalar anomaly score can be brittle, robustness is often improved by fusing diverse sensory inputs. Fusing RGB, depth, and audio, for example, helps recognize ambiguous manipulation anomalies~\citep{inceoglu2021finonetdeep}, while multi-view camera setups and explicit 3D spatial features have been shown to improve failure recognition in occluded scenes~\citep{pacaud2025guardian, hu2025mfovmultidimensional}. Despite these advances, the shared challenge across these physical monitors remains predicting anomalies fast enough to consistently avert catastrophic failures while maintaining broad generalizability and a low false-positive rate.

\paragraph{Spatio-Temporal Reasoning and Semantic Misalignment}
Beyond instantaneous anomaly detection, robust monitoring requires tracking the temporal evolution and semantic intent of a long-horizon task. Pushing toward explicit runtime semantics, monitoring can be framed as continuous spatio-temporal constraint checking via VLM-generated code~\citep{zhou2025codeasmonitor}. This naturally extends to detecting broader progress failures, where an agent might remain physically stable but lose semantic alignment with the task intent. For example, explicitly decoupling temporal inconsistency from task-progress failure in generative policies illustrates that monitoring must transcend mere action-level consistency checks to incorporate task-progress-aware verification~\citep{agia2024unpacking}. This paradigm can be pushed even further by treating the semantic misalignment itself as a direct detection target~\citep{grislain2025failsensetowardsgeneral}.

The objective of this preceding stage remains purely \emph{recognition}: detecting that progress has stalled, temporal consistency has degraded, or behavior has diverged from the task intent. Yet, simply raising an alarm does not close the safety loop. Moving beyond detection to explain \emph{why} a failure occurred shifts the problem from monitoring to diagnosis.

%%%%%%%%%%%%%%%%%%%%%%%%%%%%%%%%%%%%%%%%%%%%%%%%%%%%%%%%%%%
\subsubsection{Failure Diagnosis and Reasoning}\label{sec5:diagnosis}

While runtime monitoring determines \emph{whether} a rollout deviates from a safe regime, failure diagnosis identifies the specific failure type and its root cause. This distinction is crucial for long-horizon manipulation, where failures—such as semantic grounding errors, execution drifts, or contact-induced disturbances—are rarely interchangeable. Although these errors may all trigger a generic ``failure'' signal, they demand entirely different downstream responses. Diagnosis, therefore, serves as the critical semantic bridge that moves beyond simple detection toward informed, targeted intervention.

\paragraph{Structured Diagnosis: Taxonomies and Symbolic Reasoning}
One prominent approach focuses on localizing failures within predefined categories or structured representations. By mapping anomalies to handcrafted failure classes~\citep{inceoglu2023multimodal} or symbolic predicates~\citep{hegemann2022learning}, these methods provide a consistent semantic framework for reasoning. This structural grounding is further enhanced through relational models, such as causal networks~\citep{diehl2023causal} or semantic scene graphs~\citep{das2021semanticbasedexplainable}, which help identify the specific root causes of a failure. 

Moving beyond per-episode analysis and static, pre-defined structures, a parallel line of research investigates the automated discovery of failure taxonomies from large-scale execution data. To facilitate operator-facing explainability, recent methods leverage unsupervised learning to organize raw execution logs into discoverable, semantically distinct failure modes~\citep{gupta2025unsupervised}. Complementing this, active search strategies can be used to identify specific environmental conditions most likely to induce recurring failures~\citep{sagar2024from_mystery}. By extracting these recurring modes for subsequent debugging, these failure-reasoning frameworks reinforce human-robot trust through semantically grounded explanations~\citep{ khanna2023userstudyexploring,das2021explainableairobot}.

% \citep{inceoglu2023multimodal} failure class handcrafted, identify based on this class. 
% \citep{hegemann2022learning} symbolic action predicates and use this symbolic graph structure to reason about failure. 
% Failure diagnosis can utilize the distinct failure modes or taxonomies to reason about frequently happended semantically distinct modes, instead of every per-episde open-ended reasoning. 
% Beyond per-episode analysis, diagnosis can organize recurring failure structures across multiple executions to support operator-facing explainability. For instance, \cite{gupta2025unsupervised} leverage unsupervised taxonomy discovery from deployment logs, while \cite{sagar2024from_mystery} employ active failure search using RL policy which changing environmental conditions, identify failure modes which most likely induce to failure. These methods extract recurring failure modes for subsequent debugging and robustness improvement, proving valuable even outside the live execution loop. A broader lineage of causal and explainable AI studies applied to the robotic task~\citep{diehl2023causal,das2021explainableairobot,das2021semanticbasedexplainable,khanna2023userstudyexploring} reinforces that semantically grounded explanations significantly improve human-robot trust calibration and analyzing failure.

\paragraph{Generative Diagnosis: Open-ended Reasoning via Foundation Models}
A more recent shift moves away from closed-set classifiers toward the open-ended reasoning capabilities of large-scale foundation models. Instead of mapping inputs to a fixed taxonomy, these approaches utilize LLMs to reason over summarized multimodal sensory data~\citep{liu2023reflect}.  This evolution is further exemplified by the transition toward vision-language reasoning, where manipulation-specific failure detection is integrated with rich, natural-language explanations of the underlying causes~\citep{duan2024aha, qi2026armor_self}. Furthermore, it has been observed that diagnostic precision depends significantly on the inclusion of failure-specific supervision and explicit spatio-temporal structures~\citep{pacaud2025guardian, hu2025mfovmultidimensional}. Collectively, these developments indicate that execution-time diagnosis has matured into a dedicated multimodal reasoning problem, where advancements in foundation models enable increasingly autonomous, real-time root cause analysis.

% Beyond per-episode explanation, diagnosis can also organize reusable failure structure across many executions and support operator-facing explainability. Unsupervised taxonomy discovery from deployment logs~\citep{gupta2025unsupervised} and active failure search~\citep{sagar2024from_mystery} show that recurring failure modes can be extracted and reused for later debugging, mitigation, or robustness improvement, even when the diagnostic routine is not itself embedded in a live execution loop. Causal and explainable-failure studies~\citep{diehl2022whydidfail,das2021explainableairobot,das2021semanticbasedexplainable,khanna2023userstudyexploring} provide adjacent evidence that semantically grounded explanations can improve trust calibration and accelerate human resolution. The claim should nevertheless remain modest: diagnosis makes failure more interpretable and more actionable, but it does not by itself guarantee that the resulting intervention will be correct.

Consequently, while diagnosis renders failures highly interpretable and actionable, knowing \emph{why} a rollout failed does not inherently guarantee safety. Ensuring system integrity requires translating these insights into protective intervention. Accordingly, the challenges of deciding \textit{how} to halt execution to prevent catastrophic failure are addressed in the subsequent discussion.

%%%%%%%%%%%%%%%%%%%%%%%%%%%%%%%%%%%%%%%%%%%%%%%%%%%%%%%%%%%
\subsubsection{Runtime Shielding}\label{sec5:shielding}

While failure monitoring and diagnosis identify emerging risks, shielding transforms these signals into action-level interventions. This layer of execution-time safety moves beyond passive detection to define the physical constraints or redirection required to safeguard the system. Formal shielding methods are inspired by control-theoretic works, which can provide safety guarantees for modeled robotic systems under explicit assumptions, and have been extended to learned shielding methods to enable more complex scenarios with high-dimensional constraints and robot platforms.

\paragraph{Certifiable and Control-theoretic Shielding}

Once risk estimates are deemed actionable, shielding converts them into direct constraints on physical execution. The most rigorous approach to this intervention emerges from control-theoretic methods that enforce safety-by-construction under explicit modeling assumptions, thereby yielding formal safety guarantees. Within this framework, set-theoretic methods~\citep{bansal2017hamilton, ames2019control} establish certification of forward invariance of the modeled closed-loop system within a specified safe set. Operationally, these safety filters enforce constraints through an additional optimization layer. This modular architecture is particularly advantageous as it functions as a plug-and-play safeguard that remains policy-agnostic~\citep{brunke2022safetyfilters}.

Consequently, these formal structures have been widely adapted for various manipulation challenges, demonstrating successful integration in tasks ranging from secure robotic grasping~\citep{cortez2019control} to occlusion avoidance in vision-based control pipelines~\citep{wei2024diffocclusion}. Additionally, for long-horizon manipulation, maintaining task-consistency is critical; safety interventions must not permanently destroy task feasibility~\citep{morton2025safetaskconsistent}. Despite their theoretical strength, these formal methods are fundamentally bottlenecked by their reliance on strong system assumptions, including the need for explicit physical models and analytically tractable safety certificates~\citep{dawson2022certificates}. Consequently, to enable real-world deployment across more diverse scenarios, recent research has increasingly shifted toward learned, latent, and semantic shielding.

% Once risk estimates are deemed actionable, shielding converts them into direct constraints on physical execution. The strongest evidence here emerges from control-theoretic methods that enforce safety-by-construction under explicit modeling assumptions, therefore preserving formal guarantee of safety. Set-theoretic methods establish forward-invariance guarantees~\citep{bansal2017hamilton,ames2019control}, which means the system stays always inside the defined safe set. The safety filter is based on constraints which will be enforced through additional optimization layer placed as plug-and-play to the policy, especially useful since it can be used agnostic to the policy type~\citep{brunke2022safetyfilters}. 
% For long-horizon manipulation, task-consistent CBF design~\citep{morton2025safetaskconsistent} is particularly critical, as a safety maneuver that permanently destroys task feasibility is often operationally unacceptable. Related applications on manipulation tasks, \citep{cortez2019control} grasping. \citep{wei2024diffocclusion} avoiding occlusion on vision-based control.  
% While these works form the formal core of runtime shielding, their guarantees remain strictly conditional on model fidelity, constraint correctness, and the physical portion of the system actually under certified control.

\paragraph{Learned, Latent, and Semantic Shielding}
In generative and open-vocabulary settings where explicit dynamics models are unavailable, shielding focuses on transforming latent, statistical, and semantic data into actionable physical constraints. This functional evolution begins with verifying or adapting the proposals generated within abstract representation spaces. For instance, diffusion-policy proposals can be verified via set-based reachability to ensure safety without collapsing the nominal policy distribution~\citep{romer2025fromdemonstrationssafe}, while latent safety filters are adapted online directly from visual constraint specifications~\citep{agrawal2025anysafeadaptinglatent}. Beyond direct policy filtering, shielding can treat the system’s internal confidence as a dynamic state for intervention. By augmenting the latent state space with calibrated epistemic uncertainty, reachability-based safety filters can proactively prevent the agent from entering both known hazards and OOD states~\citep{seo2025uncertaintyawarelatent}. 

A further functional layer involves translating high-level semantic understanding or natural-language into execution-time filters. Protective safeguards can be instantiated directly from open-vocabulary scene understanding~\citep{brunke2025semanticallysaferobot}, or integrated as plug-and-play safety layers that use VLMs to formulate CBFs for pre-trained VLAs~\citep{hu2025vlsa}. For LLM-driven systems, the agent's control outputs can be checked or filtered using data-driven reachability analysis to keep physical trajectories remain within specified bounds~\citep{hafez2025safellmcontrolled}. These methodologies bridge the gap between policy-time specifications and real-time execution filtering, transforming diverse diagnostic signals into active physical shields. While these approaches are conceptually grounded in rigorous formal control theory, they often operate within a predominantly empirical regime where absolute guarantees remain challenging to establish in high-dimensional, unmodeled environments.

\subsubsection{Runtime Policy Steering}\label{sec5:steering} 

Beyond imposing hard physical constraints, execution-time safety can be proactively managed through runtime policy steering. Unlike shielding mechanisms that focus on preventing worst-case safety violations, steering intervenes \emph{before} a failure fully materializes by using predictive verification or learned signals to select  a lower-risk or verifier-preferred continuation among feasible actions. This proactive adjustment is particularly critical in long-horizon manipulation, where early deviations are often entirely salvageable if the system reacts at the first warning sign.

A prominent approach to steering involves verifying candidate futures prior to physical commitment. For instance, impending failures can be treated as a breakdown in reasoning-action alignment. By forward-simulating action proposals and employing a VLM verifier, execution can be proactively steered toward the trajectory most consistent with the intended semantic plan, thereby improving robustness against OOD visual perturbations~\citep{dowhatyousay}. This verification architecture can be further refined by decoupling outcome prediction from evaluation. By utilizing a latent world model to project future states, a VLM can be aligned to reason over these latent representations and guide policy selection~\citep{wu2025fromforesightforethought}. Furthermore, to support lightweight deployment, steering can be executed online without altering the base parameters of pretrained generative models. This is achieved by employing auxiliary verifier functions—trained on policy rollouts—to dynamically bias execution toward successful outcomes~\citep{attarian2026updatefree, nakamoto2024steering}.

However, evaluating multiple rollouts inherently introduces computational latency. To overcome this, recent advances shift toward verifier-free steering by intervening directly within the policy's generation process or internal representation. For diffusion models, collision-avoidance gradients can be injected directly into the denoising loop to optimize trajectories without post-hoc evaluation~\citep{deng2025safebimanual}. Similarly, for VLA architectures, representation engineering steers execution toward cautious behaviors by manipulating sparse feature directions within the latent space, entirely bypassing external verifiers~\citep{khan2025controlling}.

Yet, action-level steering and shielding cannot cover all possible scenarios. When a hazard cannot be cleanly circumvented via minor trajectory adjustments, or when the current subtask itself becomes structurally infeasible, simply halting the robot is insufficient. The system must transition from simple blocking or steering unsafe kinematic actions to actively modifying the high-level task plan and restoring execution.

\subsection{Runtime Adaptation and Task Restoration} \label{sec5:runtime_restore} 

Execution-time safety, therefore, extends beyond modifying immediate trajectories; it requires structural resilience. In long-horizon manipulation, critical safety phases often unfold when the active plan breaks down—whether anticipated proactively before a physical collision, or detected reactively after an error. The robot may need to request human assistance, accept a lightweight correction, structurally revise its active subgoal sequence (replanning), or rewind to a recoverable state without discarding the entire task. Consequently, this subsection addresses the operational question of \emph{how} to safely adapt, preserve, and restore task progress once straightforward sequence continuation is no longer viable.

% Execution-time safety, therefore, extends beyond detecting risk, and blocking an unsafe action. In long-horizon manipulation, critical phases often unfold \emph{after} a warning has triggered or a shield has engaged. The robot may need to request human assistance, accept a lightweight correction, revise its active subgoal sequence, or rewind to a recoverable state without discarding the entire task. Consequently, this subsection addresses operational question of \emph{how} to safely preserve and restore task progress once straightforward autonomous continuation is no longer viable.

We organize this recovery logic along a spectrum of autonomy and human reliance, moving from human-driven interventions to fully autonomous task restoration. We begin with risk-aware handoff, where uncertainty or precision demands necessitate full human involvement. We then transition to interactive correction, where sparse language or physical feedback repairs the current execution without requiring a full takeover. Moving toward greater system autonomy, we consider proactive and reactive replanning, which modify the active plan in response to detected mismatches. Finally, we examine autonomous recovery, where the system must independently diagnose the failure, generate a corrective trajectory, and resume the long-horizon task from a viable intermediate state. 

%%%%%%%%%%%%%%%%%%%%%%%%%%%%%%%%%%%%%%%%%%%%%%%%%%%%%%%%%%%
\subsubsection{Human Intervention and Control Handoff}\label{sec5:handoff}

Human intervention and control handoff represent the anchor of this spectrum—the least autonomy-preserving responses, yet the most reliable safety mechanisms once unsupervised execution is no longer justifiable. In this paradigm, the central challenge shifts from executing the task to identifying exactly \emph{when} the system must halt and relinquish control to a human operator. For long-horizon manipulation, this timing is critical: unsafe continuations often appear locally plausible until accumulated drift, missed precision constraints, or cascading errors completely collapse the window for safe recovery.

\paragraph{Intervention Triggers: Uncertainty and Task Constraints}
For long-horizon manipulation, deciding when to hand off control to a human operator is critical and must be tied to explicit runtime signals rather than ad hoc judgment. Intervention can be triggered by internal uncertainty; for instance, token-level entropy from VLA models can prompt immediate requests for one-step human correction~\citep{karli2025ask_before_you_act}. To minimize unnecessary human intervention, this internal uncertainty can be physically grounded by evaluating scene affordances, enabling intelligent planners to more accurately calibrate their confidence~\citep{mullen2025lbap}. Similarly, in imitation learning, novelty detection—identifying states with no reference behavior to imitate—and estimated task risk can dynamically allocate human intervention~\citep{hoque2021thriftydaggerbudgetaware}. Beyond internal uncertainty, handoff criteria must also respond to external structural demands, where takeover triggers are derived directly from learned precision constraints in safety-critical tasks, such as object insertion in clearance-limited environments~\citep{oh2024leveragingdemonstratorperceived}.
% Comment: LAP - tried to minimize human intervention by confidence calibration using scene affordances of LLM to avoid hallucination. But in here, we set as example of calibration method (which can be possibly used deciding whether it needs intervention or not) 
% Comment: ThrisftyDagger also wants to minimize human intervention, so it tries to decide 'when' we should need human intervention, and in other cases, we do not. 
% 

\paragraph{Intervention Outcomes: Safe Boundary Mapping and Trust Repair}
The intended outcome of an intervention is a risk-reducing fallback from impending failures—whether driven by OOD scenarios, low policy confidence~\citep{hoque2021thriftydaggerbudgetaware,karli2025ask_before_you_act}, or physical constraints in demanding tasks~\citep{oh2024leveragingdemonstratorperceived}. Consequently, the physical traces generated during these handoffs provide more than just immediate failure mitigation; they act as dense, informative signals that map the boundaries of safe execution for subsequent policy learning~\citep{korkmaz2025milemodelbased}. By connecting intervention decisions to structural task demands, intervention becomes a predictable mechanism for mapping physical limits. Furthermore, ensuring this transition is seamless requires robust human-robot communication. By communicating the robot's perceptual uncertainty through haptic and 3D visual feedback, dynamic transitions from semi-autonomous execution to full teleoperation can be achieved reliably~\citep{lee2026spirit}.
% reinforcing that transparent communication surrounding mistakes is vital for post-failure trust repair~\citep{ye2019humantrustafter}. 
% MILE: design differentiable intervention model estimate when and how the intervention occurs, use the intervention data as additional policy training. 
% SPIRIT: aerial manipulation - haptic device. Semi-autonomous / teleop transition based on perception uncertainty (from learning-based module for perception)

%%%%%%%%%%%%%%%%%%%%%%%%%%%%%%%%%%%%%%%%%%%%%%%%%%%%%%%%%%%
\subsubsection{Interactive Correction and Repair}\label{sec5:correction}

Whereas human intervention assumes the system must temporarily cede control due to unacceptable risk, interactive correction offers a less disruptive alternative to full human handoff. The robot remains actively in the loop, utilizing external feedback to repair a localized failure before it propagates into a broader task breakdown. Because continuous human intervention incurs prohibitively high costs and renders the operation suboptimal, maintaining partial autonomy is crucial. In long-horizon manipulation, many execution errors remain highly recoverable if addressed while their scope is narrow—whether a precondition is violated, an intermediate objective is underspecified, or a local control trajectory requires a minor adjustment. Correction methods therefore directly target the current failure locus, preserving overarching autonomy rather than rebuilding the task from scratch~\citep{celemin2022interactive}.

\paragraph{Semantic Correction and Task-plan Repair}
Interactive correction often begins at the planning layer, revising local objectives before or immediately after a step becomes invalid. Correction need not be strictly post-failure; by continuously monitoring for environmental discrepancies—such as unexpected object states or missing tools—actions can be pre-emptively revised before a physical execution error occurs~\citep{kim2024pre_emptive}. When failures do occur at the structural level—such as when the preconditions for a subsequent subtask are unmet—corrective actions can be generated to directly resolve the specific violations that halt task progress~\citep{raman2024capecorrectiveactions}. If autonomous semantic repair is insufficient, human language provides an interface to modify these high-level objectives. Language can repair a risky plan by injecting new constraints and intermediate subgoals~\citep{sharma2022correctingrobotplans}, or dynamically update subsequent task plans based on environmental changes~\citep{yang2024text2reactionenablingreactive}. To maximize the utility of such interventions, systems can distill and retrieve prior corrective knowledge, ensuring that recurring failures are not treated as independent events~\citep{zha2024distillingandretrieving}. Together, these methods preserve task flow by targeting the underlying semantic causes of a failure.

% Comment: PRED - changes in environments (i.e. object arrangement). embodied agents ignoring this change lead to failure. PRED allow agent to revise actions in response before mistakes occurs. Compare anticipated and actual environment states and maintain discrepances, when it happens, use as environmental feedback and revise the original plan. (using LLM).
% CAPE: corrective actions to resolve precondition errors during planning. in some case where condition for next subtask is not met. (for example, when next subtask is to put cereal on pantry, but the robot does not hold cereal at that point, then grabbing cereal should be put as next task). 
% Correcting Robot Plans: Human language is used to modify robot plan (constraints: stay away from yellow box, adding sub-goals: go from under the bottle of bleach) to avoid planner failure.
% Text2Reaction: LLM generate initial plan. Then, on-the-fly, observes change from environment, and LLM is trained to alter the planning (task) to add new plan and change adaptation.
% Distilling and Retrieving: human correction 정보를 저장, distill 해서 knwoeldge. 이후에 retrieval 해서 비슷한 상황에서 재사용해서 correction 필요를 최소화하기 위함. 

\paragraph{Spatial Correction and Execution-level Repair}
Beyond high-level planning, correction frequently targets the physical execution of a trajectory. Instead of treating language feedback as a semantic trigger, these signals can directly manipulate the low-dimensional latent control space, enabling real-time shared autonomy where human operators can modify the active trajectory~\citep{cui2023norightonline}. For dexterous or long-horizon tasks, streaming language corrections can not only supervise a robot policy to avoid intermediate failures but can also be used to generate additional corrective data to update policies online, actively preventing recurrent risks~\citep{liu2023interactiverobotlearning, shi2024yellatyour}. This execution-level guidance can also be mathematically grounded; for example, language feedback can be interpreted to update safety constraints online, warm-starting reachability-based controllers to maintain specified controller bounds~\citep{santos2025updatingrobotsafety}. These approaches can help adjust local trajectories on the fly without aborting the active subtask.
% Comment: No to the right: adaptivity of human-robot interaction. Language corrections given to a learned model that produces low-dim control space (i.e. 2-DoF joystick) that human can guide the robot. Ex. 'tilt down a little bit' 
% Interactive_robot_learning: 로봇이 실패할 것 같은데, 사람이 language correction을 주면, 해당 rollout 정보와 verbal correction을 통해서 LLM이 action candidate 중에서 어떤 action이 정합한지 확인하고, 이를 통해 policy 학습 추가로 진행함. -> 다시 같은 failur를 하지 않도록 막는 것에 의의.
% Yell at your robot: High-level policy (language policy), low-level language conditioned policy (action skill) framework. Intervention 하는 언어는 override high-level policy -> 바로 low-level skill을 불러온다. 하지만 추가로 signal을 보내서 high-level policy fine-tune도 진행함.
% Updating robot safety: Language guide + robot visual obs -> VLM 써서 interpret, safe set을 새로 구성 (occupancy map) -> warm-start reachability analysis -> safety filter update. 

\paragraph{Embodied Interfaces and Physical Correction}
While language interfaces excel at high-level corrections, physical guidance is often more effective for direct kinematic repair. When a rollout faces a near-miss due to a minor pose error rather than a complete task mismatch, it can be locally corrected via human intervention. Through interfaces like VR, operators can apply real-time pose nudges to avert failure without retraining the underlying policy~\citep{welte2026flowcorrect}. Furthermore, teleoperated physical interventions do more than prevent immediate failures of imitation policies. As \citet{jiang2025transic} demonstrate, these online physical overrides provide rich corrective data that can be used to train a residual policy, directly leveraging human intervention to close the sim-to-real gap.
% FlowCorrect: Flow-matching policy, VR guide human correction. Human provide occasional relative correction, it is used to train 'FlowCorrect' module that locally steers the policy's flow field. result in adapted policy without re-training the backbone. 
% Transic: Sim-to-Real gap으로 인해서 policy가 실패할 때, 사람이 개입해서 teleoperation으로 correction 진행, 그러나 이걸 data로 사용해서 추가적으로 학습 수행. Sim-to-real gap 메꾸는 방법으로 사용. 

In conclusion, all of these interactive corrections operate under a shared assumption: the overarching task structure remains viable. When a failure fundamentally invalidates the active sequence or pushes the system out of a locally recoverable state, minor fixes are no longer sufficient, compelling the system to cross the boundary into proactive replanning.

\subsubsection{Proactive and Reactive Replanning}\label{sec5:replanning}

Runtime policy steering (Sec.~\ref{sec5:steering}) already addresses an important class of pre-failure interventions: when the current task structure remains valid, the system can verify or adjust the next action to avoid an unsafe outcome. 
However, long-horizon execution often fails in a more structural way. 
When the problem is no longer that the next motion is slightly unsafe, but that the active subtask, ordering, or temporal constraint has become inconsistent, steering within the existing task structure is insufficient. Therefore, replanning is required to move from action-level redirection to task-level revision, modifying the active sequence, control code, or constraint specification so that execution can resume from the changed situation.
% Rather than tweaking a single action or accepting sparse human nudges, a replanning system fundamentally alters its strategy—synthesizing a new subtask sequence, generating alternative control codes, or mathematically relaxing execution constraints to salvage the current state. 
% The replanning system is essential in long-horizon manipulation, where failures typically manifest as plan-world mismatches (e.g., environmental drift or expired temporal requirements) rather than irreversible breakdowns. 
Operationally, this philosophy divides into two distinct regimes: proactive replanning, which preemptively re-routes the action sequence before an anticipated failure manifests, and reactive replanning, which dynamically synthesizes new manipulation actions to bypass a physical failure after it has occurred.

\paragraph{Proactive Re-routing}
Proactive replanning operates on the principle of preemptive task-level modification. Unlike runtime steering (Section~\ref{sec5:steering}), which attempts to salvage active subtasks through local adjustments, proactive re-routing operates at the semantic level by abandoning structurally unexecutable plans to trigger immediate, full-scale replanning. For instance, in long-horizon scenarios, spatial action sequences can be revised at subtask boundaries if current observations misalign with a reference scene graph corresponding to a successful demonstration~\citep{yu2025scene}. By extending this preemption to continuous execution, the system can be compelled to exit its active control loop and perform an immediate replan upon constraint violation, aiming to synthesize alternative trajectories before cumulative risk leads to failure~\citep{guo2024doremigroundinglanguage}.
% Comment: Scene Graph guided proactive replanning: Generate scene graph before execution of each subtask, compare to previous successful demo. If similar match found, do it. Otherewise, triggers replanning by reasoning over scene discrepancy. 
% Doremi: low-level execution과 high-level task planning이 일치하지 않을때. LLM grounding -> detection, recovery from misalignment. LLM이 high-level planning뿐 아니라 constraint generation, -> execution 시 misalignment 검출. 그리고 VLM은 constraint violation detect. Detect 시 replan. 

\paragraph{Reactive Action Synthesis and Dynamic Workaround}
When execution encounters an unexpected but still recoverable mismatch---for example, an occupied receptacle in a pick-and-place task---the safety problem is no longer only to reject the failing action, but to synthesize a new continuation that preserves task progress. 
Reactive workaround mechanisms address this intermediate regime by reinterpreting the current scene, identifying why the previous plan no longer applies, and revising the next action sequence before the local failure escalates into full task collapse~\citep{skreta2024replanroboticreplanning,pchelintsev2025lerareplanningwith}. 
% To address this, REPLAN~\citep{skreta2024replanroboticreplanning} and LERa~\citep{pchelintsev2025lerareplanningwith} utilize VLMs to track the workspace state, diagnose the need for intervention, and actively synthesize observation-driven workarounds when the manipulation scene changes. 
The same continuation-preserving idea can be expressed through more formal or architectural routes: temporal specifications may be minimally relaxed when unforeseen events make the original STL task infeasible, while modular VLA designs can separate global motion, local interaction, and skill recomposition so that failures trigger targeted replanning rather than wholesale task restart~\citep{buyukkocak2025resilientonlineplanning,yang2026lilo_vla}. 
% This reactive action generation can also be formally and architecturally constrained. From a formal perspective, new trajectories can be actively generated by computing the minimal temporal relaxation of STL constraints \citep{buyukkocak2025resilientonlineplanning}. Furthermore, at the architectural level, LiLo-VLA \citep{yang2026lilo_vla} separates the interaction policy from the global motion planner, enabling fast, reactive action generation through dynamic replanning and adaptive skill composition.
% Comment: 
% REPLAN: online replanning, VLM used for VQA, understand the tabletop environment, and adapt the robot's action when the initial plan fail to accomplish the goal. 
% LERa: VLM 이용, scene description 생성, error identification, 이후 failure reason 분석, 이후 corrective planning 수행 
% Resilient online planning:
% LiLo-VLA: Global planning (reaching) + local object-centric VLA (interaction with object) architecture. failure recovery 시 더 효율적으로 replan 가능하다. skill composition, long-horizon에 특히 좋다고 주장.

% Replanning mechanisms assume that task progress can safely resume once the task sequence or constraint structure is updated. Conversely, autonomous recovery assumes execution has degraded so severely that the system must physically rewind progress, clear hazards, or restore a viable state before meaningful continuation is  possible.

The evidence boundary of this layer is therefore continuation feasibility rather than autonomous recovery. 
Replanning assumes that progress can resume once the task sequence, temporal constraint, or skill composition has been updated. 
Autonomous recovery begins when this assumption breaks: the robot must first restore a viable physical state---by clearing hazards, undoing progress, or re-establishing stable conditions---before meaningful task continuation is possible.

%%%%%%%%%%%%%%%%%%%%%%%%%%%%%%%%%%%%%%%%%%%%%%%%%%%%%%%%%%%
\subsubsection{Autonomous Recovery and Task Restoration}\label{sec5:recovery}

Unlike replanning, which assumes that revising the active sequence is sufficient to salvage the current goal, the final tier of runtime adaptation addresses systemic failures where the overarching task structure has collapsed. At this stage, local corrections and sequence reroutes are no longer viable, necessitating comprehensive task restoration. Once progress fundamentally breaks down, the system must explicitly identify a recoverable state or generate a dedicated recovery policy to rescue the long-horizon task without discarding the entire episode. This distinction is critical because long-horizon failures are often path-dependent: once a precondition no longer holds or an object is severely misplaced, safe continuation depends on restoring the global task fabric, not merely swapping the next action.

\paragraph{Progress-aware Rollback and Rewind}
Long-horizon execution can fail not only through an isolated infeasible action, but through gradual drift away from a valid task progression. 
Progress-aware rollback addresses this failure mode by asking whether the current state remains a reliable basis for continuation, and if not, which previous checkpoint provides a recoverable restart point. 
This turns progress monitoring into an active recovery interface. 
Explicit milestones can mark when progress has stalled and when execution should rewind to a recoverable state~\citep{dai2026see_plan_rewind}, while temporal consistency monitoring with state respawning can restore the robot to semantically verified intermediate states after online failure detection~\citep{zheng2026rewind}. 
When the relevant checkpoint is not obvious from the current observation alone, episodic memory can recover prior task context to support rollback decisions~\citep{zeng2026helm}. 
At a lighter-weight introspective level, internal attention patterns or policy uncertainty can expose path deviation before errors fully compound~\citep{jeong2026yourvisionlanguage,hung2021introspectivevisuomotorcontrol}. 
The evidence boundary is checkpoint-relative, since rollback only supports recovery when progress is reversible, a valid return point can be identified, and moving back to that point remains physically feasible.

\paragraph{Learned Recovery Policies from Failure Data}

A second line of work treats recovery not as an improvised response at test time, but as a behavior that can be learned from explicit failure-state distributions. 
The motivation is that nominal demonstrations rarely expose the policy to the states in which recovery is actually needed; once execution drifts off the expert distribution, the policy must know not only that a failure occurred, but what corrective action can return the task to a viable trajectory. 
One route is to automatically generate paired failure states and executable recovery actions, then use these pairs to train models that provide failure reasoning and recovery guidance for VLA policies~\citep{lin2025failsafereasoningand}. 
Rich corrective language can make such failure data more actionable by describing the cause of the error, the required spatial correction, and the expected outcome, allowing a supervisor--actor policy to recover more robustly in long-horizon manipulation~\citep{dai2025racerrichlanguage}. 
When real failure data are costly or unsafe to collect, counterfactual failure synthesis offers another route: successful demonstrations can be perturbed through a generative world model to produce paired failure-correction data, with verification filters used to retain physically plausible recovery targets~\citep{li2026learningactionablemanipulation}. 
Complementing these offline data-generation strategies, on-policy distillation trains recovery-relevant behavior on the student policy's own visited states, using an expert teacher to provide dense supervision where standard offline fine-tuning would suffer from distribution shift~\citep{zhong2026vlaopd}.

\paragraph{Explanation-guided and Prompt-optimized Recovery}
Recent architectures increasingly rely on the generative reasoning of foundation models to orchestrate task restoration. REFLECT~\citep{liu2023reflect} and RoboFAC~\citep{lu2025robofac} translate failure localization and reasoning from language models directly into subsequent correction planning. This diagnostic process can be further structured via zero-shot chain prompting to decompose error handling into discrete reasoning stages~\citep{farag2025conditional}. To maximize this generative potential, the quality of post-failure guidance relies heavily on multimodal prompt optimization. Enriching inputs with explicit visual symbols (e.g., arrows for end-effector movement)~\citep{zeng2025diagnosecorrectand} or applying targeted image cropping to highlight key spatial elements~\citep{chen2025robotfailurerecovery} significantly enhances the spatial reasoning of VLMs, enabling direct motion-level correction.
Moving beyond external prompt engineering, embedding Chain-of-Thought directly into VLA architectures boosts general reasoning, implicitly equipping models to correct unexpected failures without requiring dedicated recovery modules~\citep{yin2025deepthinkvla}.

\paragraph{Hierarchical, Logical, and Structured Frameworks}
Although foundation models enable open-ended recovery reasoning, long-horizon recovery remains difficult when repair steps are generated without an explicit execution scaffold, since a model may identify a plausible workaround while losing track of preconditions, retry structure, or logical consistency across the remaining task. 
Hierarchical and symbolic recovery frameworks address this problem by anchoring open-ended reasoning inside structures that make recovery steps inspectable, reusable, and interruptible.

Behavior-tree-based approaches provide one such scaffold: recovery can be organized around precondition checks, postcondition monitoring, dynamic tree expansion, and newly generated skill templates, so that missing steps or disturbed subtasks are inserted into an executable task structure~\citep{ahmad2025unifiedframeworkreal,ahmad2025addressingfailuresrobotics,zhou2024llmbtperforming}. 
Neuro-symbolic recovery adds a semantic layer by using ontologies and logical rules to check whether generated recovery plans are consistent with the task constraints~\citep{cornelio2024recoverneurosymbolic}. 
Predicate-based safety logic extends the same principle to runtime safeguarding, where recent trajectories and anticipated risks are converted into executable safety predicates that can trigger mitigation or replanning~\citep{wang2025robosafesafeguardingembodied}. 
Taken together, structured recovery bounds reasoning within explicit task, logic, or safety interfaces so that recovery behavior remains inspectable.

Collectively, these varied response mechanisms provide the structural resilience needed to adapt a failing task. However, they leave one distinct, high-stakes execution regime largely unaddressed. When errors manifest or recovery must occur under sustained physical contact, safety depends not only on recognizing failure and repairing high-level plans, but also on the dynamic regulation of force, compliance, and low-level contact dynamics. We explore this contact-rich frontier in Section~\ref{sec5:physical}.

%%%%%%%%%%%%%%%%%%%%%%%%%%%%%%%%%%%%%%%%%%%%%%%%%%%%%%%%%%%
\subsection{Physical Interaction Safety Under Contact} \label{sec5:physical}

Ensuring execution-time safety becomes significantly more challenging when manipulation involves physical contact. At this frontier, the critical question shifts from whether the robot is following a correct symbolic plan to whether the executed motion is safely modulated against force limits, modeling uncertainties in friction or slip, and complex geometry. Many irreversible failures occur at these \textit{contact bottlenecks}, where an apparently correct action can jam an insertion, overload a grasp, or destabilize the task through a brief period of poorly regulated interaction.

Consequently, we treat contact as a distinct, specialized safety regime rather than a mere extension of generic recovery. This subsection progresses through three evolving paradigms: first, \textbf{adaptive compliance}, which softly yields to interaction forces; second, \textbf{formal constraints}, which enforce strict boundaries to prevent catastrophic overload; and finally, \textbf{hierarchical refinement}, which structurally coordinates these contact-aware mechanisms across multiple timescales. While recent surveys on contact-rich safe learning \citep{samadikhoshkho2025review, tsuji2026surveyimitationlearning, zhang2025safe} confirm this is a closely related and extensive research domain, our goal remains distinctly focused: to evaluate what current evidence actually proves about physical interaction safety specifically within long-horizon manipulation systems, and to identify where critical research gaps persist.

% Consequently, we treat contact as a distinct, specialized safety regime rather than a mere extension of generic recovery. This subsection progresses from force-aware adaptation to constrained contact regulation, and finally to phase-structured corrective execution. While recent surveys on contact-rich safe learning \citep{samadikhoshkho2025review, tsuji2026surveyimitationlearning, zhang2025safe} confirm this is a closely related and extensive research domain, our goal remains distinctly focused: to evaluate what current evidence actually proves about physical interaction safety specifically within long-horizon manipulation systems, and to identify where critical research gaps persist.
% These mechanisms support different strengths of safety claims, ranging from empirical robustness under contact to rigorous controller-side constraints on allowed physical interaction. 

%%%%%%%%%%%%%%%%%%%%%%%%%%%%%%%%%%%%%%%%%%%%%%%%%%%%%%%%%%%
\subsubsection{Adaptive Compliance: Enabling Physical Responsiveness}

Compliance and force-aware adaptation represent the least restrictive end of physical interaction safety. The objective is to render execution responsive enough that physical contact does not collapse into overload, jamming, unstable insertion, or catastrophic object damage. Long-horizon manipulation frequently fails at exactly these physical bottlenecks: a semantically flawless action becomes unsafe if the robot executes it too rigidly, reacts too slowly to force spikes, or relies exclusively on vision to infer contact states.

\paragraph{Passive Compliance and Low-level Impedance Control}
Rooted in the principles of \textit{Impedance Control}~\citep{hogan1987stable} and the \textit{Operational Space Formulation}~\citep{khatib1987unified}, this approach improves interaction robustness by regulating the robot's dynamic response to external forces rather than enforcing rigid position tracking. Modern research builds upon these anchors by automating and contextualizing these properties for complex tasks. For example, recent methods target large-force avoidance by learning state-dependent compliance profiles~\citep{hou2025adaptive} or estimating external forces directly from proprioceptive history to jointly model position and force control~\citep{zhi2025learning}. By providing a plug-and-play admittance layer that generalizes across diverse embodiments, these model-based approaches sidestep the sim-to-real gaps and safety uncertainties often inherent in RL policies~\citep{shi2026minimalist}. While much of this recent evidence remains empirical, the operational claim is physically profound: contact safety often hinges on whether the robot yields appropriately under uncertainty, rather than how rigidly it tracks a nominal geometric command.
% Comment: Adaptive Compliance Policy - diffusion control. dynamically adjust system compliance from human demo, 
% Minimalist Compliance Control - no force sensors needed, learn and estimate external wrenches, plug-and-play with any policy. Embodiement agnostic. 
% Learning a Unified Policy - 

\paragraph{Active Reactive Force Adaptation}
% A second approach makes contact adaptation explicitly force-aware and highly reactive during live execution. ForceMimic~\citep{liu2025forcemimic} and TacDiffusion~\citep{wu2025tacdiffusion} prove that injecting force or tactile cues directly to robot policy learning improves contact-rich manipulation performance, and also transfer among similar tactile tasks such as assembly. Pushing this toward real-time safety, FoAR~\citep{he2025foarforceaware} sharpens the execution-time claim by fusing future contact prediction with reactive force-aware control. FORGE~\citep{noseworthy2025forgeforceguided} strengthens this framework by estimate future robot action by predicting contact by leveraging both vision and tactile data, policy conditioned on force thresholds prevent failure under severe contact uncertainty, which is hard to be captured during simulation data. These papers establish a focused claim: contact-rich safety drastically improves when force feedback is not merely passively sensed, but aggressively utilized at the correct task phase to reshape execution before contact errors escalate.
A second approach treats contact adaptation as an explicitly force-aware and reactive process during live execution. ForceMimic~\citep{liu2025forcemimic} and TacDiffusion~\citep{wu2025tacdiffusion} demonstrate that integrating force or tactile cues directly into policy learning enhances contact-rich manipulation performance and facilitates transfer across similar tactile tasks. Real-time safety is further advanced by fusing future contact prediction with reactive force-aware control~\citep{he2025foarforceaware}. Furthermore, future robot actions can be estimated by predicting contact through multimodal vision and tactile data, and policies conditioned on force thresholds reduce failures under severe contact uncertainty~\citep{noseworthy2025forgeforceguided}. Collectively, these works support the empirical premise: the safety of contact-rich manipulation improves significantly when force feedback is actively leveraged—rather than merely sensed—to reshape execution before contact errors escalate.
% Comment: ForceMimic - adding wrench data into imitation learning demo helps improving contact-rich task. 
% TacDiffusion - Tactile info (wrench) added to observation vector, enables zero-shot transfer to other assembly tasks. 
% FoAR - estimate future robot action with future contact predictor. Vision+tactile info both used for estimation
% FORGE - sim-to-real transfer, policy conditiedn on maximum allowable force, help avoiding aggressive and unsafe action. Policy shows with safe exploration. 

\paragraph{Multi-modal Force Awareness in Foundation Models}
While standard VLA models primarily rely on visual and linguistic modalities, a significant shift is underway toward integrating force and tactile feedback to ground open-ended reasoning in physical reality. This integration generally follows three architectural strategies. First, several approaches focus on representation learning, where raw physical signals—including 6-axis wrench data~\citep{yu2025forcevla}, joint torques~\citep{zhang2025ta}, and high-dimensional tactile observations~\citep{huang2025tactile, bi2025vla, huang2026tafvlatactile}—are tokenized and fused into the VLA’s embedding space to enhance action decoding and contact-aware reasoning. Second, to circumvent the need for specialized hardware, recent work has explored force distillation, where physical interaction cues are inferred implicitly from visual and state transitions~\citep{zhao2026fdvlaforce}. Finally, recent frameworks have begun bridging foundation-model context with low-level execution, leveraging VLM-derived reasoning to modulate impedance parameters~\citep{zhang2026compliantvlaadaptorvlm} or to enable closed-loop hybrid force-position control via hierarchical, force-aware prompts~\citep{li2026forcevla2unleashinghybrid}. Despite these developments, the current state of the art remains primarily focused on improving empirical success rates in contact-rich tasks; there remains a critical gap in providing the direct, formal safety guarantees necessary for high-stakes robotic manipulation.
\subsubsection{Formal Constraints: Enforcing Contact Regulations}

Constrained contact regulation intervenes when compliance and force-aware adaptation fall short. The objective transitions from smoothing contact execution to imposing explicit, non-negotiable limits on physical interaction. This is critical because many contact failures in long-horizon manipulation are strictly irreversible: a momentary force overload, an unsafe contact transition, or a poorly executed avoidance maneuver can permanently damage the object, or irrecoverably derail the task state. 
% Consequently, the strongest claims in this subsection emerge from methods that directly bound contact forces or enforce online safety filters under rigorous modeling assumptions.

\paragraph{Analytic Force-bounding and Control-theoretic Constraints}
The clearest formal category of this literature leverages control-theoretic constraints to regulate interaction force directly. 
By constructing force-constrained CBFs atop generalized contact models, systems can enforce specified force limits during physical engagement~\citep{wang2025guarding}. This paradigm of rigid force-bounding has been extended to contact-based active search~\citep{vinter2024safe}, physical human-robot collaboration~\citep{sun2023adaptive}, and contact force bounding for soft actuators~\citep{wong2025contact}.
To prevent conservative safety filters from halting manipulation tasks, recent analytic frameworks prioritize task-consistency. This is achieved by formulating safety bounds directly within the robot's operational space~\citep{morton2025safetaskconsistent}, or by relaxing strict collision avoidance to permit safe \enquote{nudging} in cluttered environments~\citep{jin2026learning}.
While these methods provide conditional formal safety guarantees, their efficacy in contact-rich applications is often limited by the difficulty of modeling complex contact dynamics. Current implementations typically rely on meticulously designed but simplified analytical models~\citep{wong2025contact} or employ uncertainty observers to compensate for the discrepancy between the contact model and the true environment~\citep{wang2025guarding}. To overcome these modeling bottlenecks, a growing body of research seeks to bridge data-driven learning with control-theoretic safety filters.

% The resulting claim is robust yet narrow: given a sufficiently structured interaction model, contact-force limits can be guaranteed much more rigidly than in purely learned policy domains.
% Guarding Force:
% Safe contact-based robot active search:
% Adaptive admittance control:
% Contact-aware Safety Soft Robot: 

% \paragraph{Hard-to-model contact filtering and empirical safety layers}
\paragraph{Data-Driven Safety Filtering for Latent Dynamics}
A complementary research direction addresses contact-rich manipulation constraints where interaction forces and contact dynamics are analytically intractable, such as liquid handling or deformable object manipulation. Mitigating these complex failure modes requires extending reachability-style filtering into learned latent spaces~\citep{nakamura2025generalizingsafetybeyond}. Building on this foundation, recent advancements scale this paradigm to high-dimensional visual domains by leveraging pre-trained vision models, enabling the system to anticipate unsafe contact states prior to physical failure~\citep{tabbara2025designing}. However, since these latent filters risk inducing task-disruptive control discontinuities, recent methods explicitly optimize their underlying mechanisms for operational smoothness~\citep{nakamura2025how_to_train}. Parallel to these latent-state approaches, data-driven filtering for intricate, force-sensitive tasks—such as robotic dressing—can be realized directly through learned force-feedback models~\citep{sun2024force}. Extending these data-driven principles to the highest level of abstraction, recent frameworks introduce geometric inductive biases to project actions toward constraint-satisfying regions without destroying the overarching semantic task~\citep{tolle2025towardssaferobot}. Meanwhile, as a computational alternative to these purely neural representations, safety filters can handle severe environmental uncertainty by sparsely evaluating trajectories in high-fidelity simulations, effectively reducing risks associated with unknown object mass or surface friction~\citep{johansson2025safety}. Collectively, this literature establishes a practical paradigm: latent and simulation-driven safety layers significantly improve force and contact regulation for hard-to-model tasks, serving as indispensable empirical safeguards even without strict analytic guarantees.

\subsubsection{Hierarchical Refinement: Coordinating Multi-Timescale Safety}

Improving contact-aware safety often requires more than just yielding to forces (compliance) or capping them (constraints); it demands a structural solution to the inherent timescale mismatch between slow semantic reasoning and rapid physical dynamics, since contact-related low-level control requires a much faster response time than high-level task-planning reasoning. The final contact-safety paradigm frames contact-rich execution as a challenge of hierarchically scheduled refinement at physical bottlenecks. In this decoupled architecture, a nominal policy dictates coarse semantic progress in free space, while a highly reactive secondary mechanism takes over during contact to stabilize local geometry, force, or alignment drift. 
% This distinction is important since many long-horizon failures cluster exclusively within brief, decisive contact phases such as insertion or tightly constrained assembly. 
In these regimes, even flawlessly reasoned semantic plans will fail unless contact-aware physical correction operates at the exact right timescale and force-position tradeoff.

% \paragraph{Residual refinement at precision bottlenecks}
\paragraph{Timescale-decoupled Hierarchical Refinement}
Navigating precision bottlenecks is effectively achieved by decoupling the control architecture into a nominal long-horizon planner and a highly reactive residual controller.
Empirical evidence from high-rate, localized control frameworks suggests that such decoupling drastically outperforms global end-to-end planning during tight-tolerance contact~\citep{luo2024serl}. 
Building upon this principle, various mechanisms have been developed to operationalize timescale separation across distinct interaction constraints. For instance, by modeling non-Markovian task structures at a low rate (1--2~Hz) while offloading high-frequency responses to an asymmetric fast policy ($>$20~Hz), systems can effectively fuse slow visual cues with closed-loop tactile feedback without sacrificing local reactivity~\citep{xue2025reactive}. Similarly, nominal-residual architectures handle distribution shifts in assembly tasks by utilizing chunked behavior-cloning policies as coarse planners paired with learned residual policies that operate an order of magnitude faster~\citep{ankile2025fromimitationrefinement}. Further refinements in physical stability are achieved through multi-rate hierarchies that synchronize slow master guidance for task progression with high-frequency micro-correction for real-time wrench compensation~\citep{li2026master}. Alternatively, the decoupling of long-term learning from rapid adaptation is realized by maintaining a frozen nominal policy alongside a parallel residual pathway, which provides high-rate correction against dynamic disturbances without the need for retraining the base model~\citep{jayasinghe2026cerebellar}. Collectively, these architectures suggest a design pattern in which contact-rich safety relies on delegating the overarching geometric objective to a low-rate semantic layer, while reserving high-rate physical execution for a localized residual layer.
% Comment: From Imitation Refinement - assembly task : long-horizon, high-precision. BC fail due to chunk-level open-loop nature. Combine BC trajectory + closed-loop residula policy (trained by RL). Residual policy runs 10 Hz, while nominal action-chunk BC runs 1Hz. Fast residual policy handles execution-level distribution shift. 
% SERL - RL framework implementation study paper. Section 4.5에 impedance

% \paragraph{Phase-scheduled and hybrid force-position correction}
\paragraph{Adaptive Phase-scheduled Hybrid Refinement}
Multi-timescale safety is operationalized through the \textit{temporal synchronization} of high-level intent with reactive execution, by partitioning manipulation into distinct phases (e.g., approach/search/recovery/insert/done). Phase-scheduled architectures facilitate this by treating task progression as a sequence of contact-driven milestones. For instance, the synchronization of slow planning with fast correction can be explicitly routed via contact-aware phase predictors, which estimate the current phase belief to determine exactly when high-frequency residual correction should dominate the nominal plan~\citep{wang2026phaforcephasescheduled}.  
This temporal partitioning is reinforced by utilizing contact subgoals as discrete milestones to signal the transition from semantic planning to reactive stabilization~\citep{wang2025hierarchical}. Recent VLA architectures natively integrate these transitions by embedding explicit stage-aware force concepts directly into the semantic backbone~\citep{li2026forcevla2unleashinghybrid}, adaptively blending force and position objectives as the system progresses through the temporal phases of an assembly task.

Taken together, Section~\ref{sec:section5} establishes that execution-time safety is not a monolithic feature, but a highly layered runtime stack. Securing physical execution requires detecting emerging risks, shielding unsafe continuations, restoring task progress through active replanning, and strictly regulating physical interaction. Crucially, the evidence supporting these layers is fundamentally heterogeneous. While a select few methods offer narrow formal structures under strict assumptions, the vast majority of the literature provides empirical evidence of improved robustness within tested regimes. Therefore, the critical next question is not what execution-time methods exist, but what current evaluation practices and benchmarks actually prove about their structural safety.

\section{Evaluation and Benchmarks} \label{sec:section6}

\begin{sidewaystable}[p]
\centering
\scriptsize
\setlength{\tabcolsep}{4pt}
\renewcommand{\arraystretch}{1.15}

\begin{adjustbox}{max totalsize={\textheight}{0.94\textwidth},center}
\begin{tabularx}{\linewidth}{L{5.2cm} L{2.7cm} L{3.0cm} L{3.5cm} L{3.5cm}}
\toprule
\textbf{Benchmark / Framework} & \textbf{Role} & \textbf{Evidence object} & \textbf{Main metrics} & \textbf{Claim boundary} \\
\midrule
CALVIN~\citep{mees2022calvinbenchmarklanguage}           & Capability & Multi-stage outcome & Success, completion & Long-horizon outcome \\
LIBERO~\citep{feng2023liberobenchmarkingknowledge}      & Capability & Long-horizon action & Success & Long-horizon outcome \\
% SIMPLER~\citep{li2024evaluatingrealworld}         & Capability & Outcome & Success & Outcome \\
% Colosseum~\citep{pumacay2024colosseumbenchmarkevaluating}              & Capability & Outcome & Success, robustness & Outcome / Robustness \\
LoHoRavens~\citep{zhang2023lohoravenslonghorizon}             & Capability & Long-horizon plan & Success & Long-horizon reasoning \\
FurnitureBench~\citep{heo2023furniturebenchreproduciblereal}      & Capability & Multi-stage outcome & Task completion & Long-horizon outcome \\
RoboCerebra~\citep{han2025robocerebra}            & Capability & Multi-stage outcome & Success, accuracy & Long-horizon outcome \\
VLABench~\citep{zhang2025vlabench}               & Diagnostic & Multi-stage outcome & Progress score & Long-horizon outcome \\
RoboEval~\citep{wang2025roboevalwhererobotic}               & Diagnostic & Multi-stage outcome & Task progress, collision, object slip & Safety-relevant proxy \\
Term-Bench~\citep{liu2026trustworthy}       & Diagnostic & Execution quality & Smoothness, safety, efficiency & Execution \\
\midrule
Safe-control-gym~\citep{yuan2022safecontrolgym}       & Safety / control & Rollout & Constraint violation & Control-level \\
Safety-gymnasium~\citep{dai2023safetygymnasiumunified}      & Safety / RL & Rollout & Reward, cost & RL-level \\
Hasard~\citep{tomilin2025hasard}                 & Safety / navigation & Rollout & Safety cost & Domain-specific \\
HomeSafeBench~\citep{gao2025homesafebenchbenchmarkembodied}          & Safety / navigation & Plan / perception & Hazard detection & Domain-specific \\
EAsafetyBench~\citep{wang2025advancingembodiedagent}         & Safety / General embodied AI & Plan & Unsafe instruction rejection & Plan-level \\
InterMT~\citep{chen2025intermt}                & Safety / EQA & Rollout / preference & Preference alignment & Alignment \\
\midrule
SafePlan~\citep{obi2025safeplanleveragingformal}               & Safety & Plan & Fault detected rate & Plan-level \\
AgentSAFE~\citep{ying2025agentsafe}        & Safety & Plan / rollout & Rejection, grounding recall & Plan, Execution-level \\
Safety-as-Policy~\citep{ni2025dontletyour}    & Safety & Plan & Safe execution rate & Plan-level \\
EARBench~\citep{wu2025earbenchtowardsevaluating}               & Safety & Plan & Task risk rate & Plan-level \\
SAFEL~\citep{son2025subtleriskscritical}           & Safety & Plan & Refusal, recall rate & Plan-level \\
SafeMindBench~\citep{chen2025safemindbenchmarkingand}          & Safety & Plan / rollout & Constraint satisfaction & Cross-layer \\
IS-Bench~\citep{lu2026isbenchevaluating}               & Safety & Procedural & Safety recall / Goal condition check & Cross-layer \\
SafeAgentBench~\citep{yin2024safeagentbench}         & Safety & Plan / rollout & Semantic feasibility & Plan-level \\
VestaBench~\citep{sadhu2025vestabenchembodiedbenchmark}             & Safety & Plan & Feasible planning rate & Plan-level \\
SafeLIBERO~\citep{hu2025vlsa}      & Safety & Rollout & Shielding, safe execution & Execution-level \\
\bottomrule
\end{tabularx}
\end{adjustbox}

\caption{Representative capability-oriented benchmarks, diagnostic manipulation benchmarks, and safety-aware evaluation protocols organized by evidence object and claim boundary. (EQA: Embodied question answering) }
\label{tab:safety_benchmarks}
\end{sidewaystable}

The preceding sections reviewed where safety mechanisms intervene in the embodied pipeline: planning-time, policy-time, and execution-time. This section examines the nature of the evidence provided by current benchmarks and evaluation protocols. We treat evaluation as a lens for interpreting what reported safety scores actually substantiate and where their claim boundaries lie.

This distinction is important because nominal task success and safe execution are not equivalent. A robot may be evaluated as having completed a long-horizon manipulation task while passing through unstable grasps, excessive contact forces, near misses, or delayed interventions. Conversely, a robot may fail safely if it detects risk early, avoids harm, and preserves a recoverable state. The central question for safety evaluation is therefore not only whether the final task succeeded, but what safety-relevant events occurred during the process, when they became observable, which layer responded, and whether mitigation or recovery avoided fatal events.

We analyze both dedicated benchmarks—which provide reproducible tasks and metrics—and method-specific evaluation frameworks, as safety claims are often shaped by the latter even in the absence of reusable suites. This section is organized by evidence granularity: Section~\ref{sec:section6_1} reviews the gap between capability and safety in existing benchmarks; Section~\ref{sec:section6_3} categorizes safety-aware evaluation into plan-, policy-, runtime-, and contact-level evidence, and closes by clarifying why these layer-local claims do not yet amount to cross-layer procedural safety evidence.

\subsection{Outcome-level and Diagnostic Manipulation Benchmarks}
\label{sec:section6_1}

Capability-oriented manipulation benchmarks primarily standardize outcome-level evidence: whether a policy completes a task, reaches a goal state, or progresses through a long-horizon sequence. While essential for comparing functional performance, such evidence does not equate to safety assurance, as unsafe intermediate behaviors often remain invisible under the binary metric of final success. The limitations of current evaluation are twofold. First, mainstream benchmarks prioritize capability over safety. Second, emerging safety-aware benchmarks provide heterogeneous forms of evidence that are difficult to compare without an explicit definition of the evaluated object. 
% Consequently, this section moves from outcome-level and diagnostic benchmarks toward layer-local safety evidence and, finally, cross-layer procedural safety evaluation. 

The dominant paradigm in robotic manipulation benchmarking remains outcome-oriented. Suites such as CALVIN~\citep{mees2022calvinbenchmarklanguage}, LIBERO~\citep{feng2023liberobenchmarkingknowledge}, and FurnitureBench~\citep{heo2023furniturebenchreproduciblereal} evaluate multi-stage or long-horizon performance using metrics centered on task success or completion rates. While these benchmarks are vital for reproducibility, their primary evidence object is the final task outcome rather than the safety history of the rollout. 

Recent works have improved this resolution by reporting diagnostic execution metrics. For long-horizon tasks, VLABench~\citep{zhang2025vlabench} complements binary success with a progress score:$$\mathrm{PS} = \frac{N_{\mathrm{done}}}{N_{\mathrm{sub}}},$$where $N_{\mathrm{done}}$ is the number of completed subtasks and $N_{\mathrm{sub}}$ is the total number of subtasks. To assess execution quality, RoboEval~\citep{wang2025roboevalwhererobotic} reports stagewise progression, spatial proximity, and collision events. For instance, motion smoothness—often used as a proxy for safety—is approximated by the mean Cartesian jerk and mean joint jerk. Additionally, diagnostic evaluation includes event counts for environment collisions, self-collisions, and object slips. 
% \cite{liu2026trustworthy} similarly proposes fine-grained metrics for action quality, revealing failure modes that binary success may obscure. 

While smoothness, collision rates, and slip events provide safety-relevant insights, they do not define when risk first appeared or whether an intervention occurred before harm. They are best understood as a bridge between capability and safety evaluation. To address this, safety-aware frameworks have begun to evaluate safety more directly. EARBench~\citep{wu2025earbenchtowardsevaluating} assesses physical risk awareness in high-level planning, while SAFEL~\citep{son2025subtleriskscritical} decouples \textit{unsafe command refusal} from \textit{safe plan generation}. SafeMindBench~\citep{chen2025safemindbenchmarkingand} and IS-Bench~\citep{lu2026isbenchevaluating} further extend this focus to safety-specific metrics in embodied manipulation. Given that these frameworks diverge in their definitions of risk and metric families, the following section provides a detailed review of these works, summarized in Table~\ref{tab:safety_benchmarks}.

\subsection{Safety Evidence with Intervention Layers}
\label{sec:section6_3}

Safety-aware evaluation has developed across different evidence objects in the embodied pipeline. This division does not mean that each benchmark or protocol belongs exclusively to one layer. Rather, it identifies the primary object on which the safety claim is based. Plan-level evaluation assesses task specifications, generated plans, constraints, and unsafe-plan rejection before rollout. Rollout-level evaluation assesses whether policy-generated trajectories satisfy task objectives together with auxiliary safety conditions. Runtime-level evaluation assesses detection, intervention, and recovery once the robot is already acting. Contact-level evaluation measures physical interaction quantities such as force, slip, impulse, collision, or contact stability.
The following subsections therefore ask what each evidence level can support and where its claim boundary lies.

% This structure clarifies why layer-local safety scores are valuable but not interchangeable. A safe plan does not guarantee safe execution. A safe-success rollout does not guarantee calibrated intervention. A successful recovery does not guarantee that no additional harm occurred during restoration. 

\subsubsection{Plan-level Safety Evidence}

Plan-level safety evaluation asks whether unsafe or invalid task representations can be identified before physical rollout. At this stage, the evaluation object may be the task prompt, generated plan, constraint set, or formal specification. Plan-level evaluation can expose unsafe intent, invalid decomposition, constraint violation, or specification mismatch early in the pipeline. 
% However, the resulting evidence remains upstream evidence: it does not directly verify perception error, policy drift, contact instability, or delayed physical failure during embodied execution.
Early language-based planning systems did not necessarily introduce explicit safety metrics, but they addressed safety-adjacent plan-quality issues through executability, plan validity, and symbolic consistency~\citep{saycan,huang2022languagemodelsas,singh2023progpromptgeneratingsituated}. Affordance-grounded planning, action-vocabulary restriction, programmatic interfaces, and structured planning problems all reduce invalid or non-executable planning outputs. These criteria are important because they prevent some infeasible plans from reaching execution, but they do not show that the plan will remain physically safe once executed.

\paragraph{Risk Screening and Safe Replanning}

A first line of plan-level safety evaluation treats safety as the rejection of unsafe plans. In this view, the benchmark asks whether the system can classify a generated plan, task prompt, or action sequence as safe or unsafe before execution. SafePlan~\citep{obi2025safeplanleveragingformal} evaluates whether generated plans satisfy safety-relevant preconditions, postconditions, and invariants  to assess logical consistency, frequently employing binary classification metrics (e.g., accuracy, precision, recall, and F1 score) to evaluate the acceptance or rejection of unethical or hazardous tasks. 
% Simulation-based variants may additionally report executed percentage, fault-detection percentage, crash rate, or related indicators that measure whether a verifier blocked unsafe plans before they reached execution.

A related line goes beyond rejection by repairing or regenerating plans under safety constraints~\citep{ni2025dontletyour,yang2024plugsafetychip}. In these cases, the reported safety rate reflects whether the system can avoid or revise unsafe high-level plans. However, the strength of this evaluation depends heavily on the benchmark's hazard model. Many risks are not contained in the plan text alone but instead often depend on object placement, human presence, workspace layout, or other contextual factors. As a result, plan-level risk screening is often a context-dependent classification problem.

\paragraph{Constraint Satisfaction as Safety Compliance}
A second line evaluates safety as constraint satisfaction. Instead of asking whether a plan is globally safe or unsafe, these methods specify spatial, geometric, temporal, or logical constraints and then evaluate whether the planner respects them. VoxPoser~\citep{huang2023voxposercomposable3d}, ReKep~\citep{huang2024rekepspatiotemporal}, and related task-and-motion planning pipelines illustrate this direction by translating language or visual context into planner-usable constraints such as value maps, relational keypoints, object relations, or geometric feasibility conditions~\citep{garrett2020pddlstreamintegratingsymbolic,siburian2025grounded}. However, evaluating such compliance remains challenging when tasks are open-ended and extend beyond simple geometric boundaries. For instance, in the ReKep framework~\citep{huang2024rekepspatiotemporal}, assessing whether proposed keypoints and their associated constraints are appropriately formulated for complex tasks, such as laundry folding, often necessitates manual evaluation due to the problem's inherent open-endedness. GroundedPlanBench~\citep{jung2026spatiallygroundedlong} addresses this limitation by providing  a benchmark dataset to verify whether planning claims are grounded in spatially executable contexts. It facilitates the simultaneous evaluation of subtask planning and spatial action grounding, utilizing both explicit and implicit instructions within annotated embodied scenes.
This approach shifts evaluation beyond mere executability, focusing instead on the explicit restriction and validation of the admissible planning space.

% GroundedPlanBench~\citep{jung2026spatiallygroundedlong} evaluates whether planning claims are grounded in spatially executable task contexts rather than only asserted linguistically. This line moves evaluation beyond generic executability by restricting the admissible planning space. 
% However, safety is still often inferred from task success, constraint compliance, or logical soundness. Such metrics do not directly quantify contact risk, force-related hazards, or damage potential during physical interaction.

% A second line evaluates safety as constraint satisfaction. Instead of asking whether a plan is globally safe or unsafe, these methods specify spatial, geometric, temporal, or logical constraints and then evaluate whether the planner respects them. VoxPoser~\citep{huang2023voxposercomposable3d}, ReKep~\citep{huang2024rekepspatiotemporal}, and related task-and-motion planning pipelines illustrate this direction by translating language or visual context into planner-usable constraints such as value maps, relational keypoints, object relations, or geometric feasibility conditions \citep{garrett2020pddlstreamintegratingsymbolic,siburian2025grounded}. This moves evaluation beyond generic executability by restricting the admissible planning space. However, safety is still often inferred indirectly from task success, constraint compliance, or logical soundness. 
% These metrics do not directly quantify contact risk, force-related hazards, or damage potential during physical interaction.

\paragraph{Specification Correctness as Verifiable Satisfaction}
A third line strengthens plan-level evaluation through specification correctness. Here, the central question is not only whether a plan succeeds or satisfies a constraint, but whether the natural-language instruction has been translated into a formal or statistically reliable specification. ConformalNL2LTL~\citep{conformalnl2ltl2025} and related temporal logic planning methods treat the reliability of translation as an evaluation target, reporting quantities such as translation success and the frequency of requests for human intervention~\citep{wang2025conformaltemporallogic}. Verification-guided methods such as LAD-VF~\citep{yang2025advfllm} further integrate formal feedback into the planning loop, revising candidate plans upon specification violation and defining the safety score as the ratio of specifications successfully satisfied. These approaches provide stronger evidence than ordinary empirical success rates because they explicitly measure correctness with respect to a stated specification. 

Overall, plan-level evaluation can show that unsafe intent was rejected, a plan satisfied explicit constraints, or a formal specification was respected before rollout. These are essential upstream checks, but they remain plan-level claims.

\subsubsection{Policy-level Safety Evidence}

Policy-level evidence assesses the safety impact of mechanisms that shape a manipulation policy before actions are committed to the environment. Notably, many policy-time methods evaluate behavioral outcomes resulting from pre-commitment interventions—such as safe RL or constrained decoding—rather than scoring individual action proposals directly. Thus, while the intervention occurs at \enquote{policy-time}, the measurement remains \enquote{behavioral}. We categorize this evidence into three dimensions: constraint-conditioned, alignment-conditioned, and risk-stress behavior.

\paragraph{Constraint-conditioned Behavior}

This dimension evaluates whether policies shaped by explicit constraints can reduce safety violations without compromising task performance. It serves as the direct behavioral counterpart to constraint-aware policy generation. 
% By defining $y_i=1$ as task completion and $z_i=1$ as safety satisfaction for $i$-th rollout, we can represent the \textit{Safe Success Rate} (SafeSR) over $N$ rollouts as:
% \[
% \mathrm{SafeSR} = \frac{1}{N} \sum_{i=1}^{N} \mathbf{1}[y_i=1 \wedge z_i=1].
% \]
% This metric effectively decouples standard success from safety-compliant success.

Various frameworks instantiate this evidence by pairing success rates with domain-specific safety metrics. For instance, SafeVLA \citep{zhang2025safevla} and VLSA \citep{hu2025vlsa} report cumulative safety costs and obstacle avoidance rates to demonstrate the efficacy of safe RL and plug-and-play layers. Specifically, when safety is represented as a per-step cost $c(s,a)$ with state $s$ and action $a$, the aggregate evidence across $N$ episodes $\overline{\mathrm{CC}}$—where each episode $i$ has a time horizon $T_i$—is summarized as:
\[
\mathrm{CC}^{(i)} = \sum_{t=0}^{T_i} c(s_{i,t},a_{i,t}), \qquad \overline{\mathrm{CC}} = \frac{1}{N} \sum_{i=1}^{N} \mathrm{CC}^{(i)}.
\]
Similarly, SafeDec~\citep{kapoor2025constraineddecodingrobotics} utilizes STL satisfaction and robustness scores, while other foundation model approaches report maximum constraint violations \citep{tolle2025towardssaferobot}. Although these mechanisms intervene before action commitment, the resulting safety claims remain bounded by the underlying representation---whether an STL formula, cost function, or obstacle predicate.

% SafeVLA instantiates this evidence through task success together with cumulative safety cost. 
% Its evaluation reports success rate, cumulative robot cost, and cumulative object cost, showing whether safe RL alignment reduces safety violations without sacrificing task completion~\citep{zhang2025safevla}. 
% When safety is represented as a per-step cost, the reported aggregate can be summarized as
% \[
% \mathrm{CC}^{(i)}
% =
% \sum_{t=0}^{T_i}
% c(s_{i,t},a_{i,t}),
% \qquad
% \overline{\mathrm{CC}}
% =
% \frac{1}{N}
% \sum_{i=1}^{N}
% \mathrm{CC}^{(i)}.
% \]
% SafeDec provides a more formal version of the same evidence pattern: it constrains autoregressive action generation with STL specifications, but evaluates the resulting behavior using STL satisfaction rate, task success rate, and average robustness score~\citep{kapoor2025constraineddecodingrobotics}. 
% Safe robot foundation model work similarly places a safety layer after the foundation policy and reports success rate together with maximum constraint violation~\citep{tolle2025towardssaferobot}. 
% VLSA/SafeLIBERO evaluates a plug-and-play safety constraint layer using task success and obstacle avoidance rate~\citep{hu2025vlsa}. 
% Across these papers, the mechanism operates before action commitment, but the evidence is usually summarized over the behavior produced by the constrained policy. 
% The resulting safety claim is therefore bounded by the encoded cost, STL formula, obstacle predicate, or constraint model.

\paragraph{Alignment-conditioned Behavior}

The second dimension evaluates whether policy behavior aligns with human preferences, interventions, or contextual expectations. Unlike hard constraint enforcement, this evidence corresponds to objective shaping, where the primary question is whether the policy distribution shifts toward behavior that humans judge as safer, more efficient, or more desirable.

Several frameworks demonstrate this by aligning policies with trajectory-level preferences or feedback. For example, GRAPE~\citep{zhang2024grape} aligns VLA policies with such preferences, reporting improved success rates and reduced collision frequencies. Similarly, a range of preference-reward methods provide evidence that sparse or transferable human feedback can produce more sophisticated manipulation behavior~\citep{tian2024maximizing, liu2023pearlzeroshot, mattson2024representationalignment, moletta2026preferencealignedvisuomotor}. Beyond passive preferences, MEReQ \citep{chen2024mereqmaxent} utilizes active human intervention traces to infer residual rewards, enabling sample-efficient behavioral alignment.

\paragraph{Risk-stress Behavior}

The third dimension evaluates whether policy-time safety mechanisms remain effective under perturbations, semantic risks, or adversarial scenarios. This evidence is critical because standard benchmarks often mask latent vulnerabilities that only emerge when environmental risk factors or semantic contexts change.

Several studies quantify this robustness by exposing policies to diverse stress factors. For instance, SafeVLA \citep{zhang2025safevla} examines whether learned safety behaviors persist under perceptual and semantic perturbations. Shifting toward more structured risks, HazardArena~\citep{chen2026hazardarena} utilizes matched safe and unsafe \enquote{twin scenarios} to reveal how VLA policies progress toward hazards even when terminal success is low. Similarly, RedVLA~\citep{zhang2026redvla} employs physical red teaming by synthesizing risk-bearing scenes to expose unsafe behaviors that remain undetected in benign evaluations. Unlike symbolic plan-level checks, these evaluations measure how embodied policy behavior shifts under risk. However, the resulting evidence remains scenario-dependent, as the reported safety levels are inherently tied to the specific construction of hazards and the preservation of task feasibility under stress.

In summary, policy-time behavioral evidence constitutes \textit{behavioral validation for pre-commitment policy shaping}. While the intervention occurs before action commitment, current literature substantiates safety claims through downstream outcomes such as cumulative costs, preference-aligned metrics, and risk-stress behaviors. This layer provides more rigorous evidence than plan-level validation by evaluating the policy itself. However, it remains coarser than runtime or contact-level evidence, as it generally cannot pinpoint the temporal onset of risk, the potential for online intervention, or the precise severity of physical interactions.

\subsubsection{Runtime-level Safety Evidence}

Runtime-level evidence assesses a system’s operational behavior during execution. This layer focuses on whether a system can detect deviations from the nominal regime, trigger interventions, and recover task progress during a rollout. Unlike subsequent contact-level evaluation—which measures physical quantities like force or pressure—runtime-level evidence serves as a measure of operational safety. It indicates a robot's capacity to monitor and repair its own execution. We categorize this evidence into four dimensions: failure-rich datasets, failure detection and early warning, intervention and decision deferral, and post-failure recovery or replanning.

\paragraph{Failure-rich runtime datasets and observability}
The first dimension examines the observability of runtime failures. Conventional benchmarks typically report only final task success, failing to capture when or why a trajectory deviated. To bridge this gap, emerging failure-centric datasets provide annotated trajectories, anomaly labels, and diagnostic reasoning.

AHA~\citep{duan2024aha} constructs failure trajectories by perturbing successful demonstrations to train VLMs for natural-language failure reasoning. Guardian~\citep{pacaud2025guardian} extends this by synthesizing diverse planning and execution failures across simulated and real-world environments. RoboFAC~\citep{lu2025robofac} offers large-scale failure analysis with structured QA supervision for detection, localization, and correction, while ViFailback~\citep{zeng2025diagnosecorrectand} provides textual and visual-symbolic guidance through fine-grained VQA tasks. Similarly, benchmarks like LIBERO-Anomaly-10 and the RC benchmark shift evaluation toward execution-time failure traces rather than final outcomes~\citep{zhou2026rcnfrobot,skreta2024replanroboticreplanning}. While these datasets enable failure measurability, they are a prerequisite for evaluation of a robot's ability to avoid or recover from such failures.

\paragraph{Failure detection and early-warning evidence}
The second dimension evaluates a monitor’s ability to distinguish potential failures from nominal execution before irreversible task failure. Detection quality is typically assessed by treating failures (unsafe, anomalous, or OOD) as the positive class and successful executions as the negative class. At a fixed threshold $\tau$, where failure is predicted if $s \geq \tau$ for a failure score $s$, performance is reported via the True Positive Rate (TPR), True Negative Rate (TNR), and False Positive Rate (FPR):
$$\mathrm{TPR}=\frac{TP}{TP+FN}, \quad \mathrm{TNR}=\frac{TN}{TN+FP}, \quad \mathrm{FPR}=\frac{FP}{FP+TN},$$ 
where $TP, TN, FP$, and $FN$ corresponds to the counts of true positives, true negatives, false positives, and false negatives.
To evaluate threshold-independent ranking, studies frequently report the Area Under the ROC Curve (AUROC):
$$\mathrm{AUROC} = \int_{0}^{1} \mathrm{TPR}(\mathrm{FPR})\,d\mathrm{FPR}.$$
For instance, SAFE~\citep{gu2025safemultitaskfailure} utilizes AUROC to assess score discriminability and analyzes the trade-off between balanced accuracy ($\frac{\mathrm{TPR}+\mathrm{TNR}}{2}$) and detection latency. Similarly, FAIL-Detect~\citep{xu2025canwedetect} and FIPER~\citep{romer2025fiper} employ sequential OOD detection and action-chunk entropy, respectively, using calibrated thresholds to trigger alarms earlier and more accurately than baseline methods. 
I-FailSense~\citep{grislain2025failsensetowardsgeneral} extends this to VLM-based semantic failure detection.

For early-warning evaluation, the alarm time is defined as:
$$t_{\mathrm{alarm}} = \min\{t: s_t \geq \eta_t\},$$
where $\eta_t$ is a potentially time-varying threshold. Across $N_+$ failed rollouts, the average alarm time is $\bar{t}_{\mathrm{alarm}} = \frac{1}{N_+} \sum_{y=1} t_{\mathrm{alarm}}$. This evidence provides insight into a system’s situational awareness during execution
~\citep{xu2025canwedetect,romer2025fiper}. 

\paragraph{Intervention and decision-deferral evidence.}
The third dimension assesses the actionability of runtime signals. Intervention-based evaluation reports the frequency and timing of such requests. A general intervention rate ($\rho_{\mathrm{int}}$) can be defined as:
$$\rho_{\mathrm{int}} = \frac{1}{N} \sum_{i=1}^{N} \frac{1}{T} \sum_{t=0}^{T} \mathbb{I}[m_t=1],$$
where $m_t=1$ indicates an active intervention (e.g., stopping or requesting help). The corresponding success rate under intervention is expressed as $\mathrm{SR}_{\mathrm{int}} = \frac{1}{N} \sum_{i=1}^{N} \mathbb{I}[y_{i}=1]$.
A representative framework is Ask Before You Act~\citep{karli2025ask_before_you_act}, which utilizes token-level uncertainty from a VLA model to determine when a robot should solicit human corrective actions. 
This paradigm highlights the critical trade-off between maximizing task success and minimizing the operational cost of unnecessary human burden. 
Beyond immediate assistance, tracking the intervention rate alongside the resulting success rate serves as a benchmark for evaluating safe autonomy~\citep{ren2023robots} and measuring the efficacy of interactive training for continual policy learning~\citep{yuan2026act}.
% A representative framework is Ask Before You Act~\citep{karli2025ask_before_you_act}, which utilizes token-level uncertainty from a VLA model to determine when the robot should request a human corrective action. Its evaluation highlights the critical trade-off between increased success rates and the cost of unnecessary human assistance. Similarly, intervention rate and corresponding success rate are used the model's capability of minimizing human intervention for safe autonomy, and also effectiveness of intervention and continual policy learning with interactive training~\citep{ren2023robots,yuan2026act}.
While this evidence validates whether monitor outputs can be converted into deployable behavior, it remains highly sensitive to calibration and threshold selection.  Therefore, intervention metrics should be reported alongside detector specification, and calibration sets.

\paragraph{Recovery and replanning evidence}
The fourth dimension assesses post-failure behavior, examining whether a system can diagnose deviations and restore task progress without a full episode reset. In benchmarks involving induced perturbations, this capacity is primarily measured by the Recovery Success Rate ($SR_{\mathrm{rec}}$), representing the ratio of trials that reach the goal state following a corrective intervention. Frameworks such as REFLECT~\citep{liu2023reflect} generate failure explanations to condition correction planners, while RePLan~\citep{skreta2024replanroboticreplanning} and RACER~\citep{dai2025racerrichlanguage} utilize VLM feedback and recovery-aware policies for online adaptation. Diagnostic modules~\citep{lu2025robofac,pacaud2025guardian,duan2024aha,zeng2025diagnosecorrectand} further support this dimension when integrated as external critics to facilitate restoration after diagnosis. While recovery success is a critical indicator of long-horizon robustness, it remains an aggregate metric.

However, a binary success ratio is often too coarse-grained to characterize the quality or safety of the restoration process. It functions as a \enquote{black-box} outcome that obscures the efficiency and safety costs incurred during recovery. For instance, a high $SR_{\mathrm{rec}}$ may mask excessive time-to-recovery, high control effort, or—more critically—unstable intermediate behaviors that border on secondary failures. To provide a high-fidelity evaluation, recovery evidence must move beyond simple success labels toward granular metrics such as restoration efficiency (time and energy costs), secondary violation rates during recovery, and trajectory smoothness during replanning. 
Without these additional dimensions, recovery success remains an operational proxy for task completion rather than a definitive measure of safe and robust error handling.

Ultimately, runtime-level evidence constitutes an integrated execution-monitoring layer. Within this framework, failure-rich datasets establish the necessary observability of deviations, while detection metrics quantify the system’s early-warning capabilities. Intervention metrics further validate whether these warnings translate into reliable deployment behaviors, and recovery metrics assess the robot’s efficacy in autonomous task restoration.
However, they remain operational proxies: they demonstrate a system’s ability to monitor and repair its own execution, but do not inherently verify that physical harm or hazardous contact was strictly bounded or reduced.

\subsubsection{Contact-level Safety Evidence}

Contact-level evidence is the closest current evaluation comes to direct physical safety. Contact-level metrics can measure physical quantities that are directly related to harm or task-destructive interaction, such as contact force, impulse, torque, slip, or contact stability. This evidence is particularly important in long-horizon manipulation because a correct semantic action can still become unsafe when executed with excessive rigidity, unstable grasping, or poorly regulated contact.

\paragraph{Contact-aligned datasets and observability}
A primary line of research investigates whether a benchmark exposes the physical variables necessary to evaluate contact safety. While vision-only manipulation datasets can assess semantic task completion, they are generally insufficient for determining force-domain and contact-level safety. Consequently, proposed contact-aligned datasets provide time-synchronized traces—or a subset thereof—represented as:
\[
\mathcal{D}_i
=
\{(o_t, q_t, a_t, F_t, \tau_t, z_t^{\mathrm{tac}}, p_t)\}_{t=0}^{T_i},
\]
where $o_t$ denotes visual observations (external or egocentric), $q_t$ is the proprioceptive state, $a_t$ is the action, $F_t$ and $\tau_t$ represent force and torque measurements, $z_t^{\mathrm{tac}}$ is a tactile observation, and $p_t$ optionally denotes a contact phase or subtask label.

Recent force- and tactile-aware datasets shift the evaluative focus toward these modalities. For instance, ForceVLA~\citep{yu2025forcevla} integrates proprioception, vision,  and force-torque signals, while ForceVLA2~\citep{li2026forcevla2unleashinghybrid} extends this paradigm with multi-view observations, force prompts, and subtask-structured demonstrations. TaF-VLA~\citep{huang2026tafvlatactile} further advances this direction by providing datasets that explicitly align high-dimensional tactile observations with 6-axis force/torque signals and matrix force maps. Additionally, force-centric imitation and diffusion datasets contribute to this observability layer by recording wrench, tactile, and force-domain action information for contact-intensive tasks~\citep{liu2025forcemimic,wu2025tacdiffusion}.

\paragraph{Force regulation evidence}
The second dimension evaluates force magnitude and threshold compliance during execution. For studies reporting average interaction force, the representative metric per episode is typically defined as $\bar{F} = \frac{1}{T} \sum_{t=0}^{T} \|F_t\|_2$. This enables a direct comparison of force profiles, as seen in ForceMimic~\citep{liu2025forcemimic}, which evaluates whether interaction forces remain close to the expert distribution. Similarly, FORGE~\citep{noseworthy2025forgeforceguided} analyzes force-related quantities in Newtons alongside success rates to examine how force penalties reduce forceful interactions under varying controller gains. For tasks requiring sustained contact, such as wiping or scrubbing, recent works employ more specialized formulations to measure the mean contact-normal force:
$$\bar{F}_{n} = \frac{\sum_{t=0}^{T} \Delta t \cdot c_t |F_{n,t}|}{\sum_{t=0}^{T} \Delta t \cdot c_t + \epsilon},$$
where $c_t \in \{0, 1\}$ indicates active contact. As demonstrated in PhaForce~\citep{wang2026phaforcephasescheduled}, this formulation allows evaluators to distinguish effective execution from failures caused by over-pressing, contact dropout, or unstable regulation.

Other frameworks evaluate safety through a hard force-threshold criterion. For instance, CompliantVLA-adaptor~\citep{zhang2026compliantvlaadaptorvlm} employs a 30N threshold and terminates a trial as a failure if violations occur for three consecutive steps. This can be represented by the force-constrained success rate ($\mathrm{SR}_{F,k}$), averaged over $N$ trials:$$\mathrm{SR}_{F,k} = \frac{1}{N} \sum_{i=1}^{N} \mathbb{I} \left[ y=1 \wedge \neg \exists t: \prod_{\ell=0}^{k-1} \mathbb{I} \left( \|F_{t+\ell}\|_{\infty} > F_{\mathrm{safe}} \right) = 1 \right],$$where $k$ is the number of consecutive unsafe samples required to declare a violation. While providing direct contact-level evidence, the interpretation of this metric depends on the threshold selection and sensor quality.

\paragraph{Duration-sensitive quality and temporal evidence}
The next dimension examines the temporal integrity of contact, asking whether interactions remain functional and safe over a sustained duration. This is critical because contact-rich tasks often fail due to cumulative effects such as prolonged over-pressure or unnecessary post-success contact. 
Recent benchmarks address these nuances by decomposing task quality into time-based ratios. 
For instance, the over-pressure ($\rho_{\mathrm{over}}$) and under-pressure ($\rho_{\mathrm{under}}$) time ratios quantify the proportion of active contact duration spent outside safe or functional force margins:
$$\rho_{\mathrm{over/under}} = \frac{\sum_{t=0}^{T} \Delta t \cdot \mathbb{I} [c_t=1 \wedge |F_{n,t}| \gtrless F_{\mathrm{over/under}}]}{\sum_{t=0}^{T} \Delta t \cdot c_t + \epsilon}.$$
These metrics, often reported alongside task-specific quality scores (e.g., the wiping score $\bar{S}$ in PhaForce~\citep{wang2026phaforcephasescheduled}), are more informative than binary success rates because they separate effective interaction from force instability. 

Temporal evidence also manifests as termination efficiency. 
In contact-rich assembly, unnecessary delays after task completion increase contact exposure, potentially leading to jamming or hardware damage. 
To evaluate this, FORGE~\citep{noseworthy2025forgeforceguided} reports trial duration $\bar{T} = \frac{1}{N} \sum T_i$ and employs precision/recall metrics ($\frac{\mathrm{TP}}{\mathrm{TP}+\mathrm{FP}}$, $\frac{\mathrm{TP}}{\mathrm{TP}+\mathrm{FN}}$) to assess the quality of the policy's early-termination trigger. 
By penalizing both premature stopping and excessive contact, these temporal metrics help assess whether contact-related margins were maintained across the entire long-horizon trajectory.
% ensure that safety is maintained across the entire long-horizon trajectory.

Overall, contact-level evidence provides the most physically grounded safety signal, but it is not yet a unified safety benchmark. Future benchmarks should report not only whether a contact-rich task succeeded, but also various forms of contact-related information such as maximum and cumulative force, impulse, slip events, contact duration under unsafe margins, hazard exposure time, and whether recovery occurred without additional contact-induced harm.

Across these evidence levels described in Sec.~\ref{sec:section6_3}, the key limitation is that layer-local metrics are not interchangeable. 
A safe high-level plan does not guarantee safe execution, a policy with low aggregate safety cost may still produce transient hazardous contacts, and a successful recovery may still incur unsafe intermediate behavior. 
Thus, Section~\ref{sec:section6} characterizes the current evidence landscape as a set of bounded, layer-specific claims rather than a unified end-to-end safety argument. 
This motivates the future direction discussed in Section~\ref{sec:section7_procedure_benchmark}: developing procedural safety observability protocols that record how risks arise, propagate, are detected, and are mitigated across the planning--policy--execution lifecycle.

\section{Future Directions and Opportunities} \label{sec:section7}

% \begin{figure}[t]
%     \centering
%     \includegraphics[width=0.9\linewidth]{figures_tables/figure_files/open_question.pdf}
%     \caption{Sample Figure for Open Questions}
%     \label{fig:open_question}
% \end{figure}

\begin{figure}[t]
    \centering
    \includegraphics[width=1.0\linewidth]{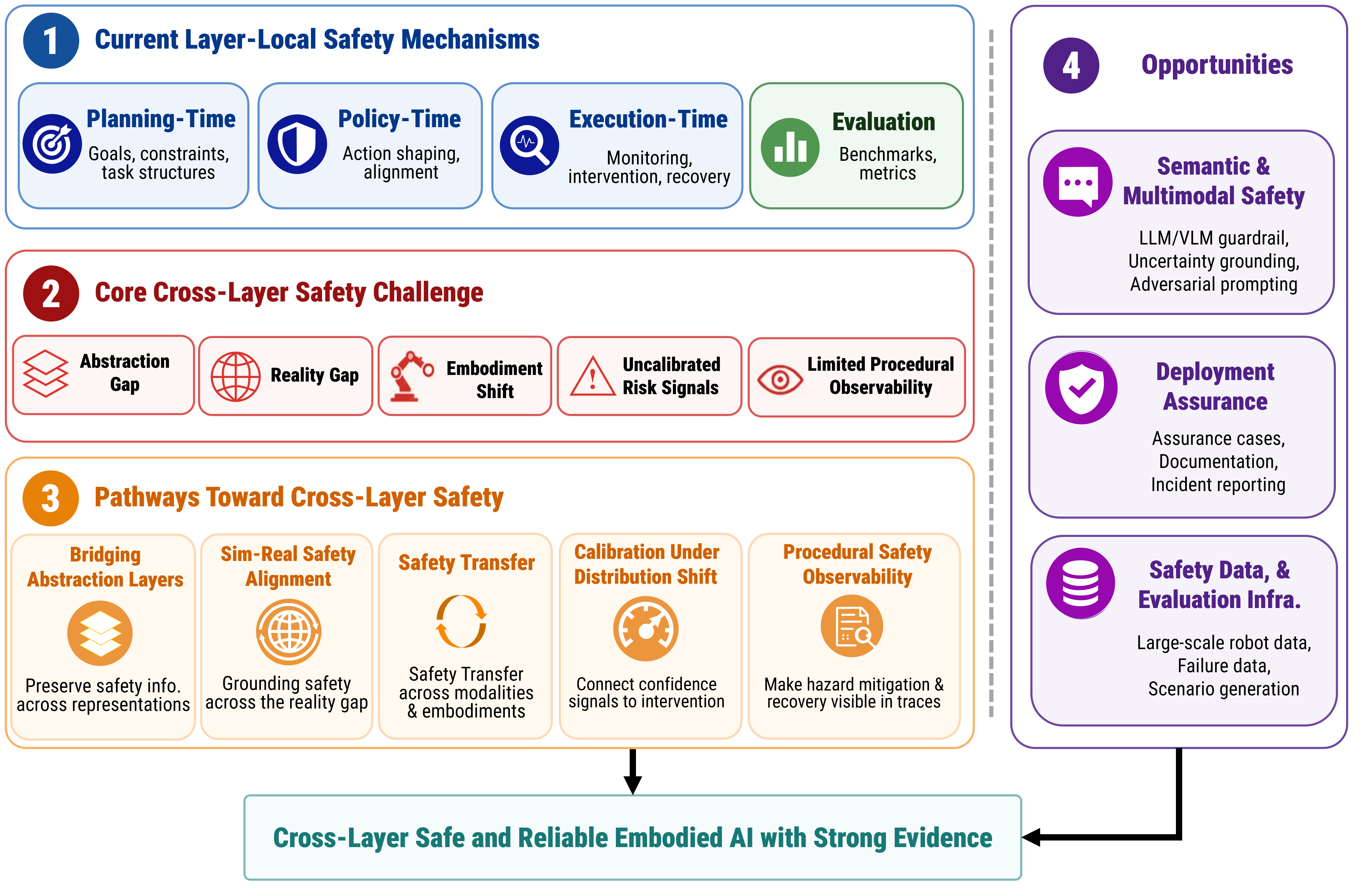}
    \caption{Illustration of roadmap for Section~\ref{sec:section7}.  
    Section~\ref{sec:pathways-cross-layer-safety} identifies five pathways toward cross-layer safety that bridge gaps in safety evidence from the prior literature. Section~\ref{sec:opportunities} highlights adjacent opportunities from semantic and multimodal safety, deployment assurance, and safety data/evaluation infrastructure. 
    % Together, these directions point toward cross-layer embodied AI safety.
    }
    \label{fig:open_question}
\end{figure}

The preceding sections have largely examined safe long-horizon manipulation through individual safety mechanisms at planning, policy, execution, and evaluation time. However, the evidence provided by these mechanisms is largely confined to their respective layers: a validated task plan may be weakened when lowered into an action abstraction, a constrained policy may fail under contact-rich execution or sim-to-real shift, and a runtime monitor may operate outside its calibration bounds.
% The preceding sections suggest that safe long-horizon manipulation is focused on  individual safety mechanisms than by the difficulty of carrying safety-relevant assumptions across the full embodied lifecycle. 
% % Planning-time methods can expose goals, constraints, and feasible task structures; policy-time methods can shape the action proposal space; execution-time methods can monitor, intervene, and recover; and evaluation methods can report partial evidence about the resulting behavior. Yet these forms of evidence remain largely local. 
% However, for instance, a validated task plan may be weakened when lowered into an action abstraction, a constrained policy may fail under contact-rich execution or under sim-to-real gap, a runtime monitor may operate outside its calibration bound. Future work therefore needs to ask not only which module can be made safer, but how safety information remains represented, grounded, and observable as it moves across abstraction, embodiment, and deployment boundaries.

This section outlines future research questions and opportunities in two complementary ways, as illustrated in Fig.~\ref{fig:open_question}. First, Section~\ref{sec:pathways-cross-layer-safety} identifies pathways toward cross-layer safety that emerge from the previous literature on robot safety. 
These pathways highlight recurring interfaces where existing layer-local methods provide partial evidence but do not yet support a coherent end-to-end safety argument. Second, Section~\ref{sec:opportunities} broadens the view to adjacent safety research and possible opportunities that long-horizon embodied tasks can benefit from progress in these neighboring fields. Semantic and multimodal safety, deployment assurance and incident learning, and data, simulation, and evaluation infrastructure may provide useful concepts and tools. Taken together, these directions point toward evidence-calibrated cross-layer safety: not a single guarantee, but a disciplined way to compose bounded claims, monitor their assumptions, and revise them when the robot leaves the conditions under which those claims were justified.

\subsection{Cross-layer Safety Directions for Long-horizon Manipulation}
\label{sec:pathways-cross-layer-safety}

The preceding sections reviewed planning-time, policy-time, and execution-time safety mechanisms. Rather than revisiting these layer-local mechanisms, this subsection highlights broader pathways that cut across them. These pathways concern how safety-relevant information is preserved across abstraction boundaries, how safety evidence survives the transition from simulation and benchmarks to real hardware, how safety claims should be revalidated under new embodiments and modalities, how uncertainty signals should be calibrated under distribution shift, and how procedural safety can be made observable over long-horizon rollouts. These directions are not intended as complete solutions, but as practical research directions for converting layer-local safeguards into stronger cross-layer safety evidence.

\subsubsection{Bridging Abstraction Layers}

A first pathway is to improve how safety-relevant information is preserved as a task moves across abstraction layers. In long-horizon manipulation, safety constraints rarely remain in a single representation. A natural-language instruction may become a grounded task description, a symbolic goal, a temporal logic constraint, a spatial or object-centric constraint, a trajectory segment, and finally a contact-rich physical interaction. At each boundary, safety-relevant information may be weakened or lost: semantic intent may be separated from physical feasibility, symbolic constraints may be separated from geometric margins, and high-level progress structure may be separated from contact limits.

Existing work already provides partial bridges across these abstractions. For example, affordance-aware planning systems~\citep{saycan} connect usefulness of language-level action with state-conditioned skill feasibility. Formal and structured planning pipelines~\citep{shirai2024visionlanguageinterpreter,yang2024plugsafetychip,obi2025safeplanleveragingformal} translate language instructions or safety constraints into planner-facing symbolic objects, temporal formulas, or precondition/postcondition checks. TAMP frameworks~\citep{garrett2020pddlstreamintegratingsymbolic} connect symbolic planning with continuous feasibility procedures, making grasp, inverse-kinematics, and collision constraints available during planning. Language-to-geometry methods~\citep{huang2023voxposercomposable3d,huang2024rekepspatiotemporal,huang2024copageneralrobotic,su2025resem3drefinable3d} further lower free-form instructions into 3D value maps, relational keypoint constraints, task-aware spatial constraints, or 3D semantic constraints.

The cross-layer issue exposed by these works is not that one representation is universally safer than another, but that each representation preserves different safety variables. Programmatic interfaces~\citep{singh2023progpromptgeneratingsituated,liang2023codeaspolicies} expose executable APIs, object lists, and assertion-like structures that can be inspected before execution. Tokenized action models~\citep{brohan2023rtvisionlanguage} expose discrete intervention surfaces for masking or constrained decoding, while trajectory and diffusion-based policies~\citep{chi2023diffusionpolicyvisuomotor} expose continuous action that can be filtered or projected geometrically. Latent or memory-augmented VLA models can support long-horizon context retention, but they may also hide safety-critical variables inside opaque representations \citep{li2024contextvla,shi2025memoryvla}. Future work should therefore ask which abstraction boundaries preserve the variables needed by downstream monitors, filters, controllers, and recovery modules.

A useful design principle would be to make abstraction boundaries safety-aware. When a constraint is introduced at the planning layer, the system should preserve enough metadata for later layers to know whether the constraint is semantic, geometric, temporal, or physical. For example, a restriction such as ``avoid the red region'' may require symbolic region naming, geometric margins, and trajectory-level exclusion. A phrase such as ``insert gently'' may require contact-phase recognition, force limits, and compliance parameters. 
One promising future research direction would be designing these safety protocols and safety-aware autonomy frameworks that preserve safety specifications and safety-relevant information across abstraction layers.

\subsubsection{Grounding Safety Across the Reality Gap}
% sim-to-real as safety evidence transfer, not only performance transfer

A second pathway is to treat sim-to-real transfer as a safety-evidence problem, not only as a performance-transfer problem. Many manipulation policies and safety mechanisms are developed in simulators, curated benchmarks, or restricted laboratory settings, where sensing, dynamics, object properties, and intervention timing are easier to control. However, a safety claim supported in simulation may be weakened or invalidated when the same policy is deployed on hardware with different sensing noise, contact physics, actuator limits, or object variation.

Existing sim-to-real research provides useful tools for narrowing this gap. Dynamics randomization trains policies across a distribution of simulated physical parameters so that the learned behavior is less tied to a single simulator configuration \citep{peng2018sim}. Adaptive simulation randomization closes the loop further by using real-world rollouts to adjust simulation distributions toward the real system \citep{chebotar2019closing}. More recent manipulation-oriented systems add other forms of transfer support. TRANSIC uses human online correction to repair unmodeled sim-to-real gaps~\citep{jiang2025transic}. MimicGen synthesizes large-scale manipulation demonstrations by adapting a small number of human demonstrations across new object and scene configurations~\citep{mandlekar2023mimicgen}, and simulation infrastructures such as ManiSkill3 and RoboCasa increase the scale and diversity of contact-rich manipulation environments~\citep{tao2025maniskill3,nasiriany2024robocasa}.

For safety, however, the relevant question is not only whether a policy remains successful after transfer. A policy that transfers task success may still transfer an invalid safety assumption. It may maintain object-placement accuracy while violating force limits, preserve collision avoidance while losing recovery margins, or maintain nominal success while triggering interventions too late on real hardware. This is particularly crucial for contact-rich manipulation, where small discrepancies in friction, tactile sensing, or force control can turn a semantically correct action into a physical hazard. Future work should therefore make the reality gap part of the safety argument. 
% For example, simulation should specify which assumptions were randomized, identified, or left fixed. Also, real-world deployment should test which assumptions remain valid.

% This framing also show how simulation should be used in regard of safety. Simulation is valuable not merely as a scalable training substrate, but as a way to stress-test safety assumptions before hardware deployment. It can expose policies to rare hazards, counterfactual failures, perturbations, and near-miss states that would be costly or unsafe to collect directly. 
% Regarding this simulation work as 
% Unless the simulator's contact, sensing, actuation, and intervention assumptions are explicitly related to the target hardware, simulation-based safety evidence should be treated as a hypothesis to be validated rather than as a deployment guarantee.

\subsubsection{Safety Transfer Across Diverse Embodiments and Policy Models}
% capability transfer is not safety transfer; revalidate under new robot, sensor, action space, model version

A third pathway is to separate capability transfer from safety transfer. Large-scale robot datasets and generalist policies increasingly support transfer across tasks, sensors, action spaces, and robot embodiments. 
% \textcolor{red}{Open X-Embodiment, DROID, BridgeData V2, RH20T, and related datasets broaden the diversity of robot experience available for training~\citep{collaboration2023openembodimentrobotic,khazatsky2024droid,walke2023bridgedata,fang2024rh20t}, while Octo and OpenVLA demonstrate how large-scale robot data can support generalist policies and adaptation across platforms~\citep{team2024octo,kim2024openvla}.} 
Recent large-scale robotic datasets and foundational generalist policies have significantly broadened the diversity of experience available for training and enabled seamless adaptation across diverse platforms~\citep{collaboration2023openembodimentrobotic,khazatsky2024droid,team2024octo,kim2024openvla}.
This trend is essential for scalable manipulation, but it creates a new safety question: when a policy, representation, reward model, or safety prior is reused on a new platform, which parts of the original safety claim remain valid?

The answer is likely to differ across abstraction levels. Semantic priors, such as avoiding humans or treating fragile objects cautiously, may transfer more readily than physical constraints. Spatial constraints must be re-grounded in the target robot's geometry, sensor calibration, kinematic feasibility, payload, and tool configuration. Contact-related claims are even more embodiment-specific, depending on conditions such as gripper compliance, tactile sensing, and actuator limits. Recent force- and tactile-aware VLA work highlights how much physical interaction information is missing from ordinary VLA representations~\citep{yu2025forcevla,zhang2025ta,huang2025tactile,zhao2026fdvlaforce,li2026forcevla2unleashinghybrid}.

Model updates create a similar revalidation problem. A VLA policy may be fine-tuned, distilled, aligned, or post-trained on new data. For instance, a reward model may be updated with new preferences or a failure detector may be recalibrated on new rollouts. Each update can improve capability while changing the distribution of actions, confidence scores, or intervention triggers. Future work should therefore make revalidation boundaries explicit whenever the robot, sensor stack, action space, dataset, or policy model changes. Furthermore, mitigating the revalidation burden across diverse embodiments represents another critical research direction for scaling generalist robot models.

% A conservative view is to treat safety transfer as conditional reuse. A transferred policy or prior should expose what is assumed to be invariant, what is conditioned on the source embodiment, and what must be locally recalibrated before the safety claim can be reissued on the target platform. This avoids a common overclaim: that cross-embodiment task success implies cross-embodiment safety. In reality, negative transfer may create false safety confidence if a behavior that was cautious on one robot becomes unsafe on another because of different gripper geometry, force sensing, compliance, controller rate, field of view, or contact dynamics.

\subsubsection{Calibrated Risk Interpretation for Intervention Selection}
% risk/confidence signals across planning, policy, monitoring, contact, and intervention need calibrated semantics

A fourth pathway is to calibrate safety-relevant signals across layers and deployment conditions. Long-horizon manipulation systems increasingly rely on heterogeneous indicators of risk, including language ambiguity, visual grounding uncertainty, action confidence, anomaly scores, failure detector outputs, force thresholds, and human intervention traces. These signals are useful, but they do not support the same safety claim. A formally checked constraint violation, a statistically calibrated failure detector, and a human correction signal should therefore not be treated as interchangeable evidence.

Recent work exposes different parts of this problem. For instance, work on confidence calibration for VLA models demonstrates that high task success does not necessarily imply a reliable self-estimation of success likelihood~\citep{zollo2025confidence_calibraion}. SAFE~\citep{gu2025safemultitaskfailure} moves closer to runtime safety by using internal VLA features and conformal prediction to produce failure alarms with an explicit accuracy--timeliness tradeoff. Other detection methods identify distinct failure categories, such as observation-space OOD, robot-object state OOD, task-level failure, action distribution uncertainty, or temporal inconsistency~\citep{xu2025canwedetect,zhou2026rcnfrobot,romer2025fiper,agia2024unpacking}. These studies suggest that the central challenge is not only increasing detector accuracy, but also determining the specific semantic implications of each calibrated signal.

The missing link is a systematic mapping from calibrated evidence to intervention semantics. A weak semantic uncertainty signal may justify a request for clarification, but not necessarily immediate physical shielding. Similarly, a calibrated failure alarm may justify stopping, backtracking, or a handoff, yet it does not specify whether the task requires high-level replanning or if local policy adjustment suffices. 
% Conversely, a physical margin violation may necessitate immediate shielding even if the high-level plan remains semantically valid. 
% Intervention-oriented systems such as \textit{Ask Before You Act}~\citep{karli2025ask_before_you_act} illustrate one concrete mapping, where token-level uncertainty triggers a halt and a one-step human correction. 
Broader long-horizon manipulation lacks a principled account of which signals should trigger specific assistance modes and at what operational scope.

Finally, the fragility of this signal-to-intervention mapping is exacerbated by distribution shift. A detector calibrated on a specific task, embodiment, and sensor configuration may lose its validity when environment or policy parameters change. Future systems must therefore develop signal-to-intervention mappings that remain robust and capable of recalibration across generalized settings.

\subsubsection{Procedural Safety Observability}
\label{sec:section7_procedure_benchmark}
% trace-level instrumentation, not just final success or layer-local metric

A final pathway is to make the safety history of a rollout visible, not only its final outcome. In long-horizon manipulation, a task may be completed while the robot passes through near misses, unstable grasps, excessive contact, delayed intervention, or recovery states that are unsafe to continue from. Conversely, a failed rollout may still be safety-relevant if the robot detects the problem early, avoids damage, and preserves a recoverable state. Consequently, the central question becomes: what specific information must an execution trace record to effectively distinguish between \textit{safe success, safe failure, unsafe success, and unsafe failure}?

Recent benchmark work provides early steps in this direction, but the observations remain fragmented. RoboEval~\citep{wang2025roboevalwhererobotic} highlights that binary success can hide execution-quality failures such as slipping during grasping. IS-Bench~\citep{lu2026isbenchevaluating} evaluates whether embodied agents notice emerging risks and perform mitigation actions in the correct procedural order. EARBench~\citep{wu2025earbenchtowardsevaluating} focuses on physical risk awareness before deployment.
Taken together, these benchmarks suggest that safety evaluation is moving beyond final success, but they do not yet provide a common trace-level account of how risk appears, evolves, and is mitigated across layers.

The missing link is a comprehensive, cross-layer safety record. Such a record should report not only task completion but also the onset of hazards, the efficacy of mitigation before safety margins were breached, the maintenance of contact bounds, and the timeliness of interventions. The objective is not merely to append more metrics to existing benchmarks, but to make the safety-relevant history of a rollout sufficiently inspectable to support a safety claim. Procedural safety observability requires that evaluation preserves the temporal and causal structure of safety evidence: identifying which risks were anticipated, which emerged, which layer responded, and whether that response led to a non-violating continuation. Establishing datasets and benchmarks based on these holistic embodied AI lifecycles remains a critical direction for future research.

\subsection{Opportunities}
\label{sec:opportunities}
Beyond manipulation-specific safety mechanism studies, several adjacent safety research areas offer useful opportunities for safer long-horizon embodied AI. These fields provide established concepts, tools, datasets, and evaluative practices that can strengthen various components of the safety stack. For instance, semantic and multimodal safety research—focused on safeguarding LLMs and VLMs—can contribute \textit{upstream} mechanisms for identifying unsafe instructions, latent task constraints, and grounding uncertainties. Furthermore, deployment assurance and incident-learning practices can clarify how safety claims should be documented, bounded, monitored, and revised after deployment. Additionally, large-scale robotic data infrastructures, simulation platforms, and scenario-generation tools can make safety-relevant evidence both more scalable and more comparable across platforms. 
% The fundamental challenge lies in translating these cross-domain concepts into grounded embodied safety evidence that accounts for robot state, sensing uncertainty, contact dynamics, and the specific context of deployment.

\subsubsection{Semantic and Multimodal Safety}

A first source of opportunity comes from safety research on general AI agents, language models, and multimodal foundation models. Within language-only systems, alignment and safety research has developed mechanisms for shaping model behavior through human feedback, explicit principles, model-generated critiques, and input-output safeguards~\citep{dong2025safeguarding}. RLHF-style instruction tuning shows how human preference feedback can align model outputs with user intent~\citep{ouyang2022training}, while Constitutional AI~\citep{bai2022constitutional} illustrates how rule-like principles can be used to critique and revise model behavior, reducing the reliance on direct human labeling. Guardrail models such as Llama Guard~\citep{inan2023llamaguardllm} further demonstrate that safety policies can be operationalized as risk taxonomies and input-output classifiers, and multimodal safety benchmarks such as MM-SafetyBench~\citep{liu2024mm} highlight that VLMs  remain vulnerable to unsafe image-text combinations even when their language backbones are safety-aligned.
% \textcolor{red}{Uncertainty quantification ~\citep{han2026flow}}

For long-horizon manipulation, these methods are most effective when reinterpreted as upstream \textit{semantic safety mechanisms}. A guardrail-like model can assist in detecting hazardous or underspecified instructions, exposing latent constraints omitted from natural-language commands, or determining when a task requires human clarification prior to execution. Multimodal safety models may similarly help identify visually grounded hazards, such as unsafe object-context relations, human proximity, fragile objects, or ambiguous affordances~\citep{chi2024llamaguardvision}. Research into object hallucination in Large VLMs is particularly pertinent: POPE~\citep{li2023evaluating} and LURE~\citep{zhou2023analyzing} show that VLMs can assert the presence of objects not supported by visual input. In an embodied context, such hallucinations become safety-critical when they are used to select manipulation targets, infer affordances, or define task preconditions. Multimodal personalization methods that ground vision-language models in user-specific visual and semantic contexts ~\citep{repic, covip, omni-persona} offer a complementary direction: extending these capabilities to VLA settings can help narrow plausible user intentions through personal priors, thereby supporting safer operation.

This convergence of opportunities defines a distinct and compelling research frontier: \textit{embodied semantic safety}. Several promising avenues emerge for further exploration within this domain.
First, safety foundation models can expand from static text-image filtering toward \textit{dynamic, long-horizon risk taxonomies} that capture physical hazards, object fragility, and spatial-temporal anomalies during execution.
Second, rather than acting as passive veto filters, semantic monitors could focus on developing \textit{safety-driven active elicitation} to quantify their own contextual uncertainty and proactively query users when high-level instructions lack explicit safety specifications. By exploring these robot-centric challenges, multimodal guardrails have the potential to evolve from digital filters into a proactive foundation for physical autonomy.

% However, the utility of semantic and multimodal safety evidence remains inherently bounded. A model that classifies a high-level instruction as \textit{safe} does not prove that the resulting grasp, trajectory, contact force, or recovery state will be physically safe. Similarly, while a hallucination detector can indicate that the perceptual basis of a plan is unreliable, but it cannot certify downstream execution. The opportunity, therefore, lies in utilizing semantic and multimodal safety methods as sources of safety priors, grounding uncertainty signals, clarification triggers, and adversarial test cases—all while explicitly linking them to embodiment-specific constraints, runtime monitors, contact-aware controllers, and trace-level validation.

\subsubsection{Deployment Assurance and Incident Learning}
\label{sec:deployment-assurance-incident-learning}

A second opportunity arises from the fields of systems safety, AI governance, and assurance engineering. In safety-critical domains, a system is deployable only when its claims, assumptions, evidence, risks, and operating boundaries are explicitly documented and maintained. Assurance engineering practices, such as Goal Structuring Notation~\citep{GSNstandard}, autonomous-system standards such as UL~4600~\citep{UL4600_2022}, and machine learning assurance processes such as AMLAS~\citep{hawkins2021guidance}, all emphasize that safety evidence must be structured around explicit claims rather than reported only as benchmark performance. For long-horizon embodied AI tasks, this suggests an opportunity to move from local performance claims toward auditable deployment arguments.

This perspective expands the scope of documentation required for embodied systems. A comprehensive safety claim characterizes the operational context in which the system is intended to function—encompassing the robot embodiment, gripper, sensors, action representation, workspace assumptions, and object categories. It further accounts for human-proximity conditions, contact limits, intervention protocols, and recovery procedures. Documentation practices from responsible machine learning, such as model cards~\citep{mitchell2019model} and datasheets for datasets~\citep{gebru2021datasheets}, offer useful templates for making these boundaries visible. In robotic manipulation, analogous policy cards or robot-data datasheets could record which embodiments, sensors, tasks, hazards, contact regimes, failure cases, and recovery scenarios were covered during training and evaluation. Such documentation would help prevent overclaiming by clarifying where evidence is strong, where it is missing, and when deployment requires revalidation.

A complementary opportunity is incident learning. Current AI incident-reporting initiatives aim to collect and classify real-world harms and \textit{near-harms} to identify recurring risks post-deployment~\citep{mcgregor2021preventing,OECD2023Towards}. For long-horizon manipulation, a comparable practice would log not only catastrophic failures but also near misses, unsafe successes, unnecessary human interventions, delayed handoffs, contact anomalies, and instances where the robot violated implicit constraints despite completing the task. The value of deployment assurance is to treat safety not as a static benchmark result, but as a living deployment argument that is continuously updated through documentation, incident learning, and embodiment-specific revalidation.

\subsubsection{Large-Scale Data, Simulation, and Evaluation Infrastructure}

A third opportunity lies in the infrastructure used to collect, curate, simulate, and evaluate safety-relevant robot experiences. Recent large-scale data initiatives indicate a shift in manipulation research from isolated task demonstrations toward shared, multi-environment, and multi-embodiment data ecosystems. Datasets such as RoboNet, BridgeData V2, Open X-Embodiment, DROID, RH20T, and AgiBot World~\citep{dasari2020robonet,walke2023bridgedata,collaboration2023openembodimentrobotic,khazatsky2024droid,fang2024rh20t,bu2025agibot_iros} significantly expand the scale and diversity of robot experience, while generalist policies such as Octo and OpenVLA~\citep{team2024octo,kim2024openvla} demonstrate how such data can support broad pretraining and adaptation across tasks, sensors, and embodiments. This trend toward data scaling offers a significant safety opportunity, as broader data coverage may reduce brittle behaviors often triggered by ordinary distribution shifts.

However, broader capability data does not necessarily equate to safety data. If a dataset primarily consists of successful expert demonstrations, the resulting policy can be trained without learning the capacity to stop, seek clarification, recover from failure, or avoid latent physical risks. This distinction is consistent with recent work on safety pretraining for LLMs: safe behavior does not automatically emerge with model scale; rather, it requires targeted efforts such as safety-conscious data curation, structured refusal of hazardous content, and explicit harm annotations~\citep{maini2025safetypretraining}. The robotics analogue is therefore not simply to collect more successful demonstrations, but to build data infrastructure that captures safety-relevant diversity: hazards, near-misses, failed attempts, contact measurements, and unsafe successes. Failure-centric datasets and reasoning frameworks such as REFLECT~\citep{liu2023reflect}, AHA~\citep{duan2024aha}, RoboFAC~\citep{lu2025robofac}, and Guardian~ \citep{pacaud2025guardian} point in this direction by treating erroneous trajectories, failure explanations, and correction signals as supervision. Many papers incrementally generate failure data by perturbing successful trajectories, but this practice remains limited; pursuing this direction requires more scalable failure-dataset generation. 

Finally, simulation and scenario-generation infrastructure can complement real robot data because safety-relevant failures are rare, costly, or unsafe to collect directly on hardware. Scenario-based evaluation in autonomous driving, including platforms such as CARLA~\citep{dosovitskiy2017carla} and probabilistic scenario languages such as Scenic~\citep{fremont2019scenic}, illustrates how rare \enquote{long-tail} events and constrained configurations can be stress-tested prior to deployment. Robot-specific tools such as MimicGen~\citep{mandlekar2023mimicgen} and ManiSkill~\citep{tao2025maniskill3} similarly suggest that synthetic demonstrations and scalable simulation can broaden coverage for manipulation. These developments suggest an opportunity to integrate data, simulation, and evaluation into a unified safety substrate: successful demonstrations provide behavioral priors, failure data helps characterize risk boundaries, simulation generates rare hazards, and execution traces test whether safety-relevant assumptions remain valid throughout a rollout.

\section{Conclusion} \label{sec:section8}

This survey has reviewed safe long-horizon robotic manipulation as an anchor domain for embodied AI and argued that its safety challenges are inherently cross-layer. In this setting, safety depends not only on whether a system can plan, act, or recover, but on how safety-relevant assumptions, constraints, and failure signals propagate across specification, planning, policy formation, execution, and evaluation. Motivated by this view, we organized the literature not only by method family, but also by intervention locus, failure mode, and evidence type.

The review reveals a clear imbalance between progress and assurance. Planning-time methods can improve grounding, specification, decomposition, verification, and motion-feasible support, but they rarely guarantee safe rollout in open-world manipulation. Policy-time methods can constrain or align action generation, yet much of their evidence remains empirical and task-specific. Execution-time methods are therefore indispensable, because long-horizon manipulation remains vulnerable to compounding drift, contact uncertainty, state mismatch, and recovery failure even when upstream components appear strong. At the same time, current evaluation practice remains limited: final task success alone is too coarse to capture procedural safety, intermediate violations, recovery quality, and residual risk.

More broadly, the literature remains rich in empirical heuristics but relatively thin in formal or statistically grounded safety claims for manipulation-centered embodied systems. This gap does not diminish the practical value of current methods, but it does require greater precision in how safety claims are formulated and compared. A central message of this survey is therefore that it is essential to distinguish what a method improves from what it actually guarantees, and to separate planning-time, policy-time, and execution-time safety rather than treating them as interchangeable notions of safe behavior.

Looking ahead, several directions appear especially important: more faithful safety specification under natural-language ambiguity; more expressive and monitorable action representations; clearer decision rules for when a robot should act, ask, halt, or hand off; stronger notions of task restorability after failure; and evaluation protocols that treat safety as a procedural property rather than a binary outcome. More generally, progress will likely depend less on any single algorithmic advance than on clearer cross-layer safety architectures, stronger evidence discipline, and more principled deployment arguments. In this sense, safe long-horizon robotic manipulation is not only an important application area, but also a useful lens through which to study the broader alignment problem between semantic intent, physical interaction, and system-level oversight in embodied AI.

\backmatter

\bmhead{Acknowledgements}

This work was supported by the InnoCORE program of the Ministry of Science and ICT (1.260007.01).

\bibliography{sn-bibliography} % common bib file

% \newpage

% \input{section/paper_notes}

\end{document}